\documentclass[sigconf]{acmart}

\copyrightyear{2025}
\acmYear{2025}
\acmConference[KDD '25]{Proceedings of the 31st ACM SIGKDD Conference on Knowledge Discovery and Data Mining V.2}{August 3--7, 2025}{Toronto, ON, Canada}
\acmBooktitle{Proceedings of the 31st ACM SIGKDD Conference on Knowledge Discovery and Data Mining V.2 (KDD '25), August 3--7, 2025, Toronto, ON, Canada}
\acmDOI{10.1145/3711896.3736910}
\acmISBN{979-8-4007-1454-2/2025/08}

\AtBeginDocument{%
  }

\usepackage{caption}
\usepackage{subcaption}
\usepackage{algorithm}
\usepackage{algorithmic}
\usepackage{amsmath}

\usepackage{amssymb}
\usepackage{longtable}
\usepackage{graphicx}
\usepackage{hyperref}
\usepackage{bm}
\usepackage{enumitem}

\renewcommand\footnotetext[1]{}
\thanks{Preprint.}

\begin{document}

\title[Diffusion-Scheduled Denoising Autoencoders for Anomaly Detection in Tabular Data]
    {Diffusion-Scheduled Denoising Autoencoders \\ for Anomaly Detection in Tabular Data}


\author{Timur Sattarov}
\authornote{Research conducted in part as an external PhD candidate at the School of Computer Science, University of St.Gallen (HSG), St.Gallen, Switzerland.}
\affiliation{%
    \institution{Deutsche Bundesbank}
    \department{Research Data and Service Centre}
    \city{Frankfurt am Main}
    \country{Germany}
}
\email{timur.sattarov@bundesbank.de}

\author{Marco Schreyer}
\authornote{Research conducted in part while the author was with the International Computer Science Institute (ICSI), Berkeley, CA, USA.}
\affiliation{%
    \institution{Swiss Federal Audit Office (SFAO)}
    \department{AI \& DA Group}
    \city{Bern}
    \country{Switzerland}
}
\email{marco.schreyer@efk.admin.ch}

\author{Damian Borth}
\affiliation{%
    \institution{University of St.Gallen (HSG)}
    \department{School of Computer Science}
    \city{St.Gallen}
    \country{Switzerland}
}
\email{damian.borth@unisg.ch}

\renewcommand{\shortauthors}{Timur Sattarov, Marco Schreyer, \& Damian Borth}
\renewcommand{\figureautorefname}{Fig.}
\renewcommand{\tableautorefname}{Tab.}
\renewcommand{\equationautorefname}{Eq.}
\renewcommand{\sectionautorefname}{Sec.}

\begin{abstract}

Anomaly detection in tabular data remains challenging due to complex feature interactions and the scarcity of anomalous examples. Denoising autoencoders rely on fixed-magnitude noise, limiting adaptability to diverse data distributions. Diffusion models introduce scheduled noise and iterative denoising, but lack explicit reconstruction mappings. We propose the \textit{Diffusion-Scheduled Denoising Autoencoder (DDAE)}, a framework that integrates diffusion-based noise scheduling and contrastive learning into the encoding process to improve anomaly detection. We evaluated DDAE on 57 datasets from ADBench. Our method outperforms in semi-supervised settings and achieves competitive results in unsupervised settings, improving PR-AUC by up to 65\% (9\%) and ROC-AUC by 16\% (6\%) over state-of-the-art autoencoder (diffusion) model baselines. We observed that higher noise levels benefit unsupervised training, while lower noise with linear scheduling is optimal in semi-supervised settings. These findings underscore the importance of principled noise strategies in tabular anomaly detection.

\end{abstract}



\maketitle

\section{Introduction}

\begin{figure}[t!]
    \centering
    \includegraphics[width=1\linewidth]{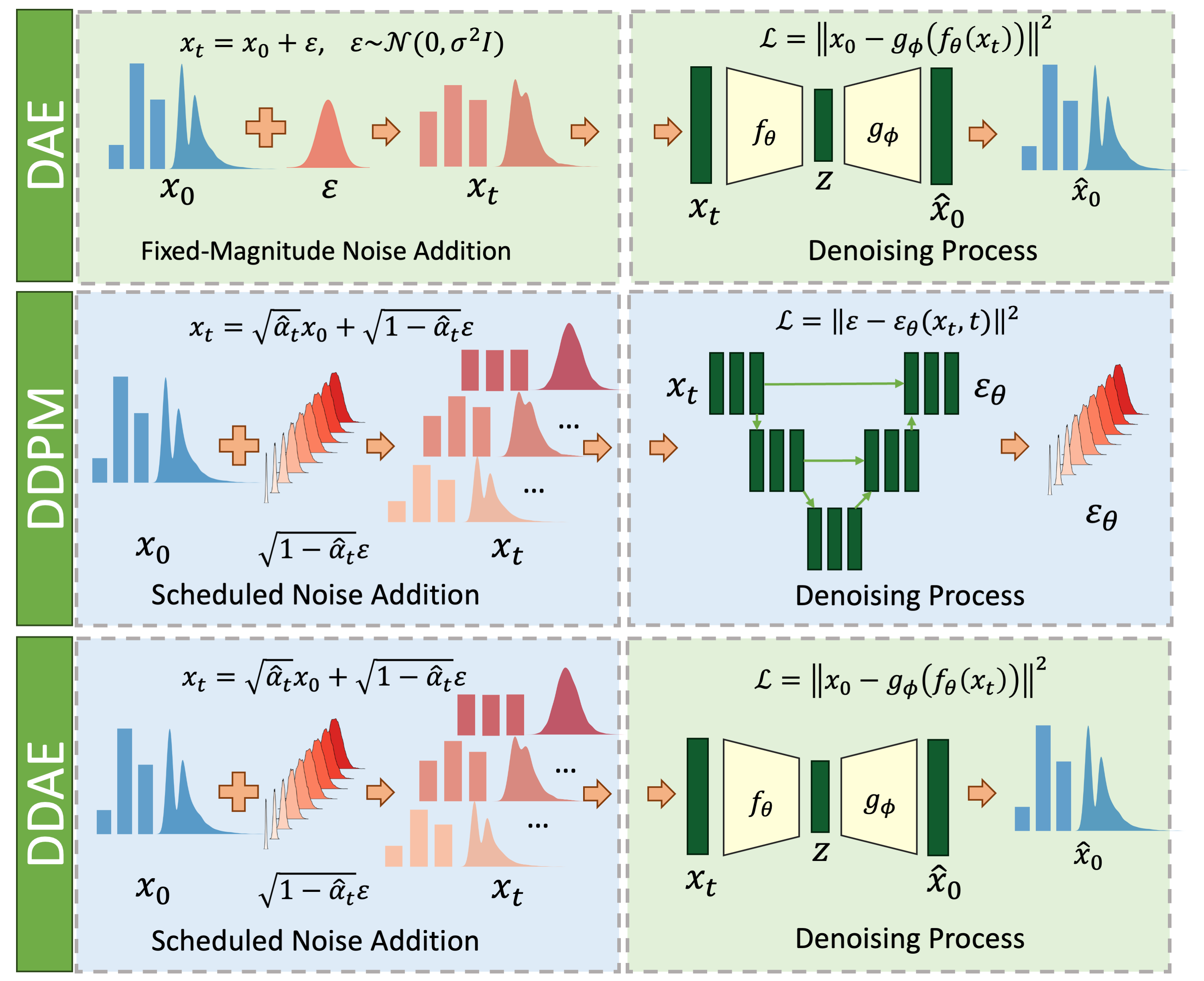}
    \caption{Comparison of noise addition and denoising mechanisms in DAE~\citep{vincent2008extracting}, DDPM~\citep{ho2020denoising}, and DDAE. DAE applies fixed-magnitude noise, training an autoencoder to reconstruct the original data. DDPM progressively adds and removes noise across multiple steps. DDAE merges both approaches, incorporating DDPM's scheduled noise addition while retaining DAE's reconstruction objective.}
    \label{fig:fig1}
\end{figure}

\begin{figure*}[t!]
  \centering
  \begin{subfigure}[b]{0.49\linewidth}
    \centering
    \includegraphics[width=\linewidth]{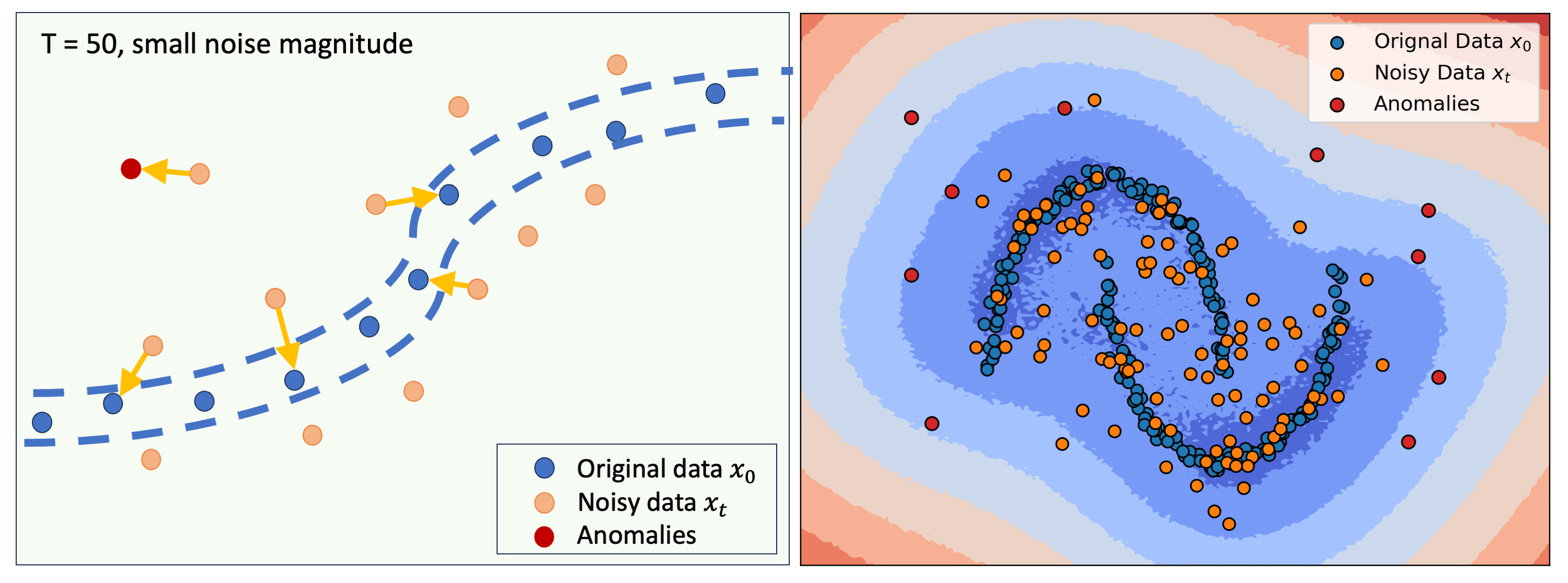}
    \caption{T = 50}
    \label{fig:decision_boundary1}
  \end{subfigure}
   \hfill
  \begin{subfigure}[b]{0.49\linewidth}
    \centering
    \includegraphics[width=\linewidth]{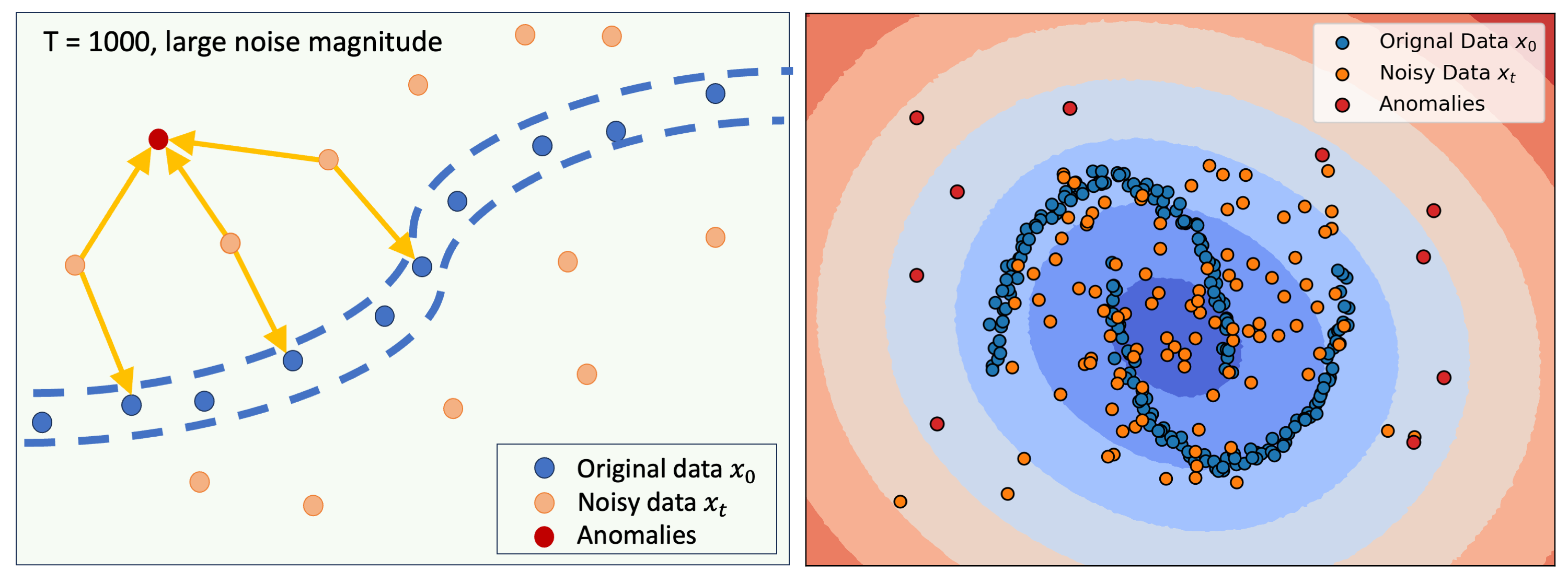}
    \caption{T = 1000}
    \label{fig:decision_boundary2}
  \end{subfigure}
 
  \vspace{-0.2cm}
\caption{Decision boundaries of anomaly scores on a toy dataset learned by DDAE with small (T = 50) and large (T = 1000) noise magnitudes. Each subfigure shows an explanatory schematic (left) and decision boundary (right). The dashed blue line represents the original data boundary, while arrows indicate reconstruction direction. With small noise (a), the model preserves the data manifold, enhancing generalization. High noise levels (b) introduce “confusion samples”, causing the distortion of decision boundary, misaligning the model with the true data distribution.}
  \label{fig:decision_boundary}
  \vspace{-0.1cm}
\end{figure*}

The detection of anomalies is a fundamental challenge in data science with critical applications in cybersecurity~\citep{yaseen2023role, rajbahadur2018survey},  manufacturing~\citep{kamat2020anomaly, liu2024deep}, finance~\citep{ahmed2016survey, hilal2022financial}, and healthcare~\citep{fernando2021deep}. Identifying rare or abnormal instances is essential for risk management, fraud detection, quality assurance, and threat mitigation. Although significant advances have been made in structured data (\textit{e.g.,} time series) and unstructured data (\textit{e.g.,} images, text), tabular data remains particularly challenging due to its mixed data types, high dimensionality, and complex feature dependencies~\citep{chalapathy2019deep}. The scarcity of labeled anomalies, class imbalance, and the need for interpretability further complicate the problem, necessitating specialized approaches.

\vspace{0.1cm}

Deep learning has introduced promising solutions to anomaly detection in tabular data~\citep{pang2021deep, borisov2022deep, neloy2024comprehensive}, with \textit{autoencoders} (AEs), \textit{denoising autoencoders} (DAEs), and \textit{variational autoencoders} (VAEs) widely adopted for their ability to learn compact representations of normal patterns~\citep{berahmand2024autoencoders, sattarov2022explaining, kascenas2022denoising, eduardo2020robust, schreyer2017detection}. More recently, diffusion models~\citep{ho2020denoising}, originally developed for generative tasks, have demonstrated strong anomaly detection performance, particularly in computer vision~\citep{wolleb2022diffusion, wyatt2022anoddpm, zhang2023diffusionad}. These models leverage a progressive noising process, leading to successful applications in anomaly detection for images and, more recently, tabular data~\citep{livernoche2023diffusion}.

\vspace{0.1cm}

Adding noise denotes a key factor in deep learning-based anomaly detection~\citep{kascenas2023role, barker2023robust}. The type of noise introduced, its magnitude, and spatial distribution, directly affect a model’s ability to discriminate between regular observations and anomalies. Traditional DAEs apply fixed-magnitude additive noise, whereas diffusion models apply scheduled noise injection, gradually increasing noise over multiple steps. \autoref{fig:fig1} illustrates the integration of diffusion-based noise scheduling into the autoencoder framework. Despite its success in medical imaging~\citep{barker2023robust}, diffusion-based noise scheduling has not been systematically explored for tabular data.

\vspace{0.1cm}

The effect of noise scheduling on anomaly detection is illustrated in Fig.~\ref{fig:decision_boundary}. Using a small timestep noise (T = 50), the model preserves the data manifold and improves generalization. In contrast, a large timestep noise (T = 1000) introduces `confusion samples' that distort the model's alignment with the original data distribution. The observation highlights the impact of integrating scheduled denoising into the autoencoder framework.

\vspace{0.1cm}

In this work, we propose \textit{diffusion-scheduled denoising autoencoder} (DDAE), which integrates diffusion-based noise scheduling into the denoising autoencoder framework to enhance anomaly detection in tabular data. Unlike conventional DAEs that rely on static noise addition, DDAE adaptively modulates noise levels using a diffusion scheduler, improving anomaly discrimination. Drawing inspiration from recent studies~\citep{xiang2023denoising, shenkar2022anomaly, schreyer2021multi}, we introduce an extension that incorporates contrastive learning in the latent space, enhancing the separation of normal and anomalous instances.

\vspace{0.1cm}

In summary, we provide the following contributions.

\begin{enumerate}[left=0pt, itemsep=2pt]
    \item \textbf{Methodology:} Introduction of the \textit{diffusion-scheduled denoising autoencoder} (DDAE) that integrates diffusion noise scheduling into the denoising autoencoder framework.
    
    \item \textbf{Representation:} Extension of DDAE using a contrastive learning objective (DDAE-C) that improves tabular data representations and anomaly detection performance.
    
    \item \textbf{Evaluation:} Comprehensive evaluation of DDAE and DDAE-C, demonstrating their effectiveness in semi-supervised and unsupervised anomaly detection settings.
    
    \item \textbf{Analysis:} In-depth analysis of the scheduled noise addition process, examining its impact on anomaly detection performance across multiple noise protocols.
\end{enumerate}

\vspace{0.1cm}

\noindent Our evaluation is based on the protocol used by Livernoche et al.~\citep{livernoche2023diffusion}, who benchmarked various anomaly detection methods, including diffusion-based models. We adopt the same evaluation pipeline, including the \textit{ADBench}~\citep{han2022adbench} benchmark, dataset preprocessing, and baselines. This ensures comparability across unsupervised and semi-supervised anomaly detection settings.\footnote{ Reference implementation available at \href{https://github.com/sattarov/AnoDDAE}{https://github.com/sattarov/AnoDDAE}}

\section{Related Work}
\label{sec:related_work}

In recent years, diffusion-enabled anomaly detection in tabular data has gained significant interest. The following literature survey encompasses (i) conventional methods, (ii) deep learning, and (iii) diffusion-enabled methods.

\subsection{Conventional Methods}

Conventional anomaly detection methods often rely on statistical models, distance metrics, and tree-based ensembles~\citep{chandola2009anomaly}. For a complete review of these methods, we refer to existing surveys~\citep{ahmed2016survey, hodge2004survey, domingues2018comparative}. Notable techniques include \textit{principal component analysis}, \textit{k-nearest neighbors}, and \textit{isolation forest}, which identify anomalies based on deviations from expected distributions or variations in local density. Although effective in low-dimensional contexts, these methods exhibit limited scalability and fail to capture complex, high-dimensional dependencies. Ensemble-based models, including \textit{ random forests} and \textit{gradient-boosted trees}, offer improved robustness but remain constrained by their reliance on feature engineering, limiting their generalizability to heterogeneous tabular domains~\citep{Aggarwal2017, goldstein2016comparative}. 

\subsection{Deep Learning Methods}

Deep learning has advanced anomaly detection by modeling complex non-linear relationships in high-dimensional spaces~\citep{pang2021deep}. \textit{Autoencoders}, \textit{variational autoencoders}~\citep{kingma2013auto}, and \textit{generative adversarial networks}~\citep{goodfellow2020generative} have been widely used to learn compact representations of normal data, detecting anomalies as deviations in the latent space~\citep{schreyer2019detection}. Recently, self-supervised learning has further improved the anomaly detection performance by mitigating the challenge of limited labeled anomalies~\citep{hojjati2022self, chalapathy2019deep}. These methods often struggle with robust density estimation and the capture of feature dependencies, particularly in tabular data. To address this, Dai et al.~\citep{dai2024unsupervised} proposed generating noise-augmented instances and identifying anomalies based on the evaluation of their noise magnitude. 

\subsection{Diffusion Methods}

Diffusion-based generative models have shown strong performance in anomaly detection~\citep{wolleb2022diffusion, wyatt2022anoddpm, zhang2023diffusionad}. Wolleb et al.~\citep{wolleb2022diffusion} introduced a diffusion-based method to localize diseased regions in medical images. Wyatt et al.~\citep{wyatt2022anoddpm} proposed \textit{AnoDDPM}, which transitions from Gaussian to simplex noise using reconstruction-based anomaly detection. Zhang et al.~\citep{zhang2023diffusionad} proposed \textit{diffusionAD}, integrating norm-guided procedures and one-step denoising to enhance inference efficiency. Mousakhan et al.~\citep{mousakhan2023anomaly} proposed a conditioning mechanism that uses the noise extracted from the target image to guide detection. Shin et al.~\citep{shin2023anomaly} introduced score-based diffusion models that directly estimate anomaly scores. Recently, Zamberg et al.~\citep{zamberg2023tabadm} introduced \textit{TabADM}, an unsupervised anomaly detection framework that evaluates the discrepancy between predicted and injected noise across timesteps. Similarly, Livernoche et al.~\citep{livernoche2023diffusion} proposed diffusion time estimation, which performs distributional alignment across noise schedules to inform detection.

\vspace{0.1cm}

Although prior work has explored various noise injection strategies for anomaly detection, to the best of our knowledge, this is the first study to systematically examine their role in the context of tabular data. 

\begin{figure*}[ht!]
  \centering
  \includegraphics[width=\linewidth, keepaspectratio]{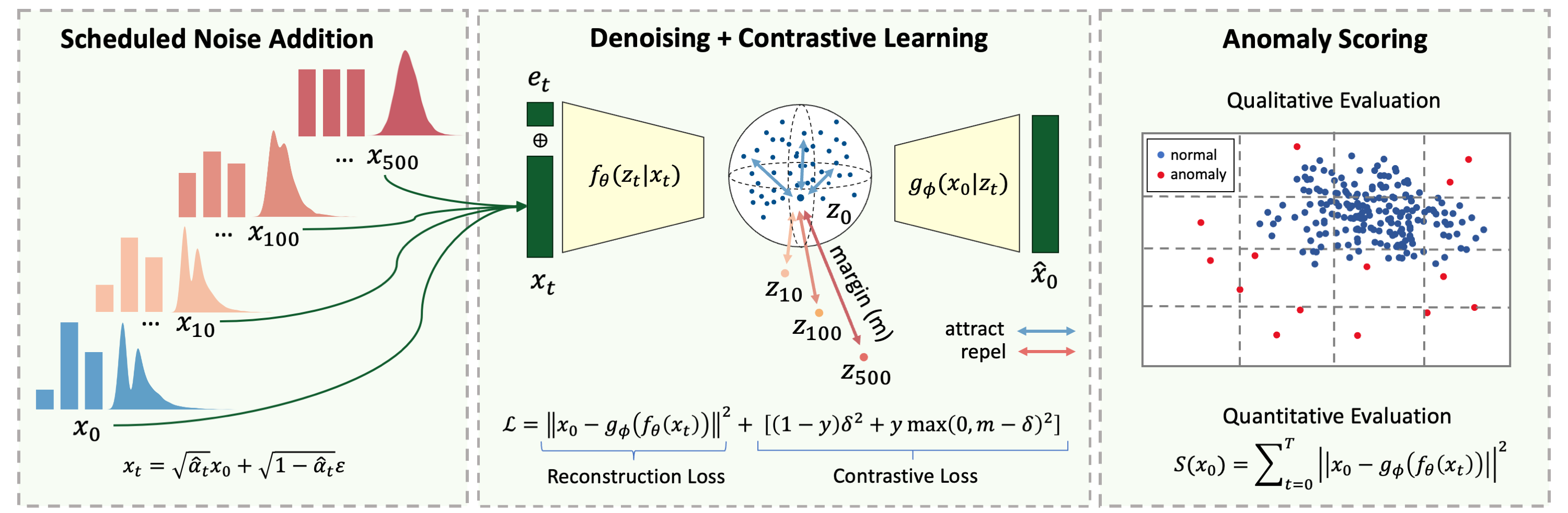}
  \vspace{-0.5cm}
  \caption{Schematic diagram of the proposed model. The process begins by producing noisy data samples $\mathbf{x}_t$ by adding Gaussian noise as per the forward diffusion process. In the denoising process, the goal of the model is to reconstruct the original $\mathbf{x}_0$ sample given its noisy counterpart $\mathbf{x}_t$. At the same time, the contrastive loss repels the representation of the noisy $\mathbf{z}_t$ from the original data point $\mathbf{x}_0$. The repelling magnitude is proportional to the amount of injected noise (e.g. timestep $t$).}
  \vspace{-0.2cm}
  \label{fig:ddim_ssl_sketch}
\end{figure*}

\section{Methodology}
\label{sec:methodology}

Next, we introduce the \textit{diffusion-scheduled denoising autoencoder} and its contrastive learning extension, extending the \textit{denoising autoencoder} by incorporating diffusion-based noise scheduling to improve anomaly detection.

\subsection{Denoising Autoencoder}

The \textit{denoising autoencoder} (DAE)~\citep{vincent2008extracting} learns to recover clean inputs from their corrupted versions. It comprises an encoder–decoder architecture, where the encoder \( f_{\theta} \) maps an input \( \mathbf{x} \in \mathbb{R}^d \) to a lower-dimensional latent representation \( \mathbf{z} \in \mathbb{R}^k \), with \( k < d \), and the decoder \( g_{\phi} \) reconstructs the input as \( \hat{\mathbf{x}} = g_{\phi}(\mathbf{z}) \). During training, Gaussian noise \( \boldsymbol{\epsilon} \sim \mathcal{N}(\mathbf{0}, \sigma^2 \mathbf{I}) \) is added to inputs $\mathbf{x}$:

\[
\mathbf{x}_{\text{noisy}} = \mathbf{x} + \boldsymbol{\epsilon},
\]

\noindent and the model is optimized to minimize the mean squared reconstruction loss, as given by:

\[
\mathcal{L}_{\text{rec}} = \mathbb{E}_{\mathbf{x}, \boldsymbol{\epsilon}} \left\| \mathbf{x} - g_{\phi}(f_{\theta}(\mathbf{x}_{\text{noisy}})) \right\|^2,
\]

\noindent where $\phi$ and $\theta$ denote the encoder and decoder network parameters. This objective encourages the network to learn robust representations by denoising corrupted inputs while preserving the underlying structure of the data.

\subsection{Denoising Diffusion Probabilistic Models}

The \textit{diffusion probabilistic denoising model} (DDPM)~\citep{sohl2015deep, ho2020denoising} is a latent variable model that utilizes a forward process to perturb the data $\mathbf{x}_0 \in \mathbb{R}^d$  step by step with Gaussian noise $\bm{\epsilon}$, and then restores the data in the reverse process. The forward process is started at $\mathbf{x_0}$ and latent variables $\mathbf{x}_1 \ldots \mathbf{x}_T$ are generated with a Markov Chain by gradually perturbing it into a pure Gaussian noise $\mathbf{x}_T \sim \mathcal{N}(\mathbf{0},\mathbf{I})$. Sampling $\mathbf{x}_t$ from $\mathbf{x}_0$ for an arbitrary $t$ can be achieved in a closed form, as defined by:

\vspace{-0.3cm}

\begin{equation}
    q(x_t|x_0)=\mathcal{N}(x_t;\sqrt{1-\hat{\beta_t}}x_0, \hat{\beta_t}, \textbf{I}) 
    \label{eq:q_step_fast}
\end{equation}

\noindent where~\scalebox{0.95}{$\hat{\beta_t}=1-\prod_{i=0}^{t} (1-\beta_i)$} and ~\scalebox{0.95}{$\beta_t$} is the noise level added at timestep $t$. The parameters  ~\scalebox{0.95}{$\beta_t$}  are linearly spaced hyperparameters controlling the noise schedule. With the help of such a noise scheduler, any arbitrary noisy version $\mathbf{x}_t$ can be directly computed:

\begin{equation}
    \mathbf{x}_t = \sqrt{\bar{\alpha}_t} \mathbf{x}_0 + \sqrt{1 - \bar{\alpha}_t} \boldsymbol{\epsilon},    
    \label{eq:ddpl_noise_add}
\end{equation}

\vspace{0.1cm}

\noindent where~\scalebox{0.95}{$\alpha_t := 1-\beta_t$} and~\scalebox{0.95}{$\hat{\alpha_t} := \prod_{i=0}^{t} \alpha_i$}. In the reverse process, the model denoises $\mathbf{x}_t$ to recover $\mathbf{x}_0$. A neural network is trained to approximate each step as~\scalebox{0.95}{$p_\theta(x_{t-1}|x_t)=\mathcal{N}(x_{t-1}; \mu_\theta(x_t, t), \Sigma_\theta(x_t,t))$}, where $\mu_\theta$ and~\scalebox{0.95}{$\Sigma_\theta$} are the estimated mean and covariance. According to Ho et al.~\cite{ho2020denoising}, with~\scalebox{0.95}{$\Sigma_\theta$} being diagonal, $\mu_\theta$ is calculated as:


\begin{equation}
    \mu_\theta(x_t, t)=\frac{1}{\sqrt{\alpha_t}}(x_t - \frac{\beta_t}{\sqrt{1-\hat{\alpha_t}}} \epsilon_\theta(x_t, t)),
\end{equation}

\vspace{0.1cm}

\noindent where~\scalebox{0.95}{$\bm{\epsilon_\theta(x_t, t)}$} denotes the predicted noise component. Empirical evidence suggests using reconstruction loss that yields better results compared to the variational lower bound~\scalebox{0.95}{$\log p_{\theta}(x_0)$}, defined as:


\begin{equation}
    \mathcal{L}_{rec}=\mathbb{E}_{x_0,\epsilon,t}||\epsilon-\epsilon_\theta(x_t,t)||^2.
\end{equation}

\vspace{0.1cm}

\subsection{Diffusion-Scheduled Denoising Autoencoder}

Traditional reconstruction-based models, such as autoencoders, have demonstrated strong performance in anomaly detection by learning to reconstruct normal data while assigning higher reconstruction errors to anomalies. In contrast, diffusion models have shown remarkable success in modeling complex data distributions through progressive noise injection. The \textit{diffusion-scheduled denoising autoencoder} (DDAE) takes advantage of both approaches by integrating diffusion-based noise scheduling into an autoencoder framework. The DDAE consists of the following three components as illustrated in \autoref{fig:ddim_ssl_sketch} and described in the following:

\vspace{0.1cm}

\begin{enumerate}[left=0pt, itemsep=2pt]

    \item \textbf{Scheduled Noise Addition.} DDAE replaces static noise addition with a progressive forward diffusion process. At each timestep \( t \), noise is added incrementally to the input, generating a perturbed sample \( \mathbf{x}_t \) as defined in~\autoref{eq:ddpl_noise_add}. This scheduling produces a smooth degradation from clean data to pure noise in \( T \) steps, allowing the model to learn reconstruction over varying noise intensities.
    
    \vspace{0.1cm}

    \item \textbf{Denoising with Timestep Conditioning.} DDAE employs a feedforward encoder–decoder architecture to reconstruct the original input \( \mathbf{x}_0 \) from \( \mathbf{x}_t \). Timestep information is encoded through sinusoidal position embeddings \( \mathbf{e}_t \), following~\citet{nichol2021improved}, and concatenated with \( \mathbf{x}_t \) prior to encoding. The encoder computes a latent representation\( \mathbf{z} = f_\theta(\mathbf{x}_t \oplus \mathbf{e}_t) \), which the decoder maps back to a denoised estimate \( \hat{\mathbf{x}}_0 = g_\phi(\mathbf{z}) \). The model is trained to minimize the mean squared reconstruction loss, as defined by:
    
    \begin{equation}
    \mathcal{L}_{\text{rec}} = \mathbb{E}_{x_0, \epsilon, t} \lVert x_0 - g_{\phi}(f_\theta(x_t, e_t)) \rVert^2.
    \end{equation}
    
    \vspace{0.01cm}
    
    \item \textbf{Anomaly Scoring.} Samples that deviate from the learned distribution result in higher reconstruction errors, particularly under high noise conditions. DDAE computes anomaly scores as the cumulative mean squared reconstruction error across all diffusion steps $T$, as defined by: 
    
    \begin{equation}
    S(x_0) = \textstyle\sum_{t=1}^{T} \lVert x_0 - g_\phi(f_\theta(x_t, e_t)) \rVert_2^2. 
    \label{eq:ddae_anomaly_score}
    \end{equation}

    \vspace{0.01cm}
    
\end{enumerate}

\noindent By combining the noise injection of DDPMs with the denoising of DAEs, the DDAE framework provides robust anomaly detection, benefiting from gradual corruption and recovery.

\begin{figure*}[t]
  \centering
  \includegraphics[width=\linewidth]{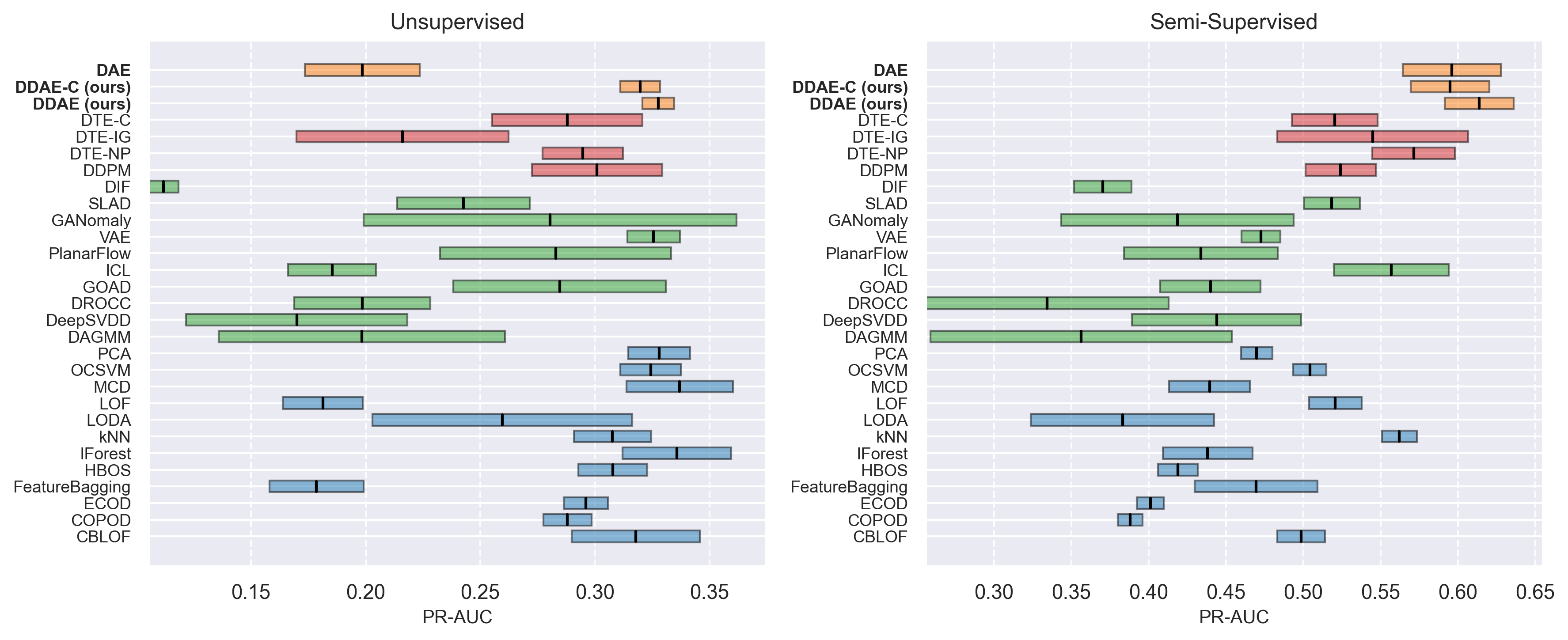}
  \vspace{-0.4cm}
  \caption{Mean PR-AUC values and their standard deviations across 57 datasets from ADBench with five distinct seeds in both unsupervised and semi-supervised contexts. Color coding: blue for conventional methods, green for deep learning approaches, red for diffusion-based techniques, and orange for our method. Results for models other than ours are sourced from \citep{livernoche2023diffusion}.}
  \label{fig:baseline_comparison_grouped}
  \vspace{-0.1cm}
\end{figure*}

\subsection{Contrastive DDAE} While DDAE effectively reconstructs normal patterns under varying noise levels, it does not explicitly enforce the separation in the latent space between normal and anomalous data. To address this limitation, we propose DDAE-C, an extension of DDAE that integrates contrastive learning into its latent space. 

\vspace{0.1cm}

Contrastive learning encourages the model to attract semantically similar inputs (positive pairs) and repel dissimilar ones (negative pairs), thereby enhancing feature-level discrimination. This mechanism is particularly beneficial for detecting anomalies, where deviations from normal behavior can be subtle. DDAE-C promotes more robust and discriminative latent representations by combining denoising and contrastive learning objectives.

\vspace{0.1cm}

\noindent\textbf{Contrastive Training Framework.} In DDAE-C, we enhance latent representation quality by integrating contrastive learning. We define positive and negative pairs as follows:

\begin{itemize}
    \item \textbf{Positive pairs:} $\mathbf{x}^+ = (\mathbf{x}_0^{(i)}, \mathbf{x}_0^{(j)})$, where both $\mathbf{x}_0^{(i)}$ and $\mathbf{x}_0^{(j)}$ are independently drawn from the training set. Given the typically low proportion of anomalies in real-world datasets, the majority of such pairs are expected to consist of normal samples and can be treated as approximate positive pairs. While some pairs may include anomalous instances, their impact on the overall representation learning is anticipated to be limited due to their sparsity.
    
    \item \textbf{Negative pairs:} $\mathbf{x}^- = (\mathbf{x}_0, \mathbf{x}_t)$, where $\mathbf{x}_t$ denotes a noisy variant of $\mathbf{x}_0$ obtained via the forward diffusion process. This formulation encourages the model to structure the latent space such that clean inputs $\mathbf{x}_0$ are clustered more closely, while their corrupted counterparts $\mathbf{x}_t$ are pushed away. As a result, samples that differ substantially from the dominant data distribution tend to occupy peripheral regions in the learned representation space.
\end{itemize}

\noindent Let $\mathbf{z}_0$ and $\mathbf{z}_t$ denote the latent representations of $\mathbf{x}_0$ and its noised version $\mathbf{x}_t$, respectively. The contrastive loss is defined as:


\begin{equation}
\mathcal{L}_{\text{cont}} = \mathbb{E}_{\delta} \left[ (1 - y)\cdot \delta^2 + y \cdot \max(0, m - \delta)^2 \right],
\end{equation}

\vspace{0.1cm}

\noindent where $\delta = \lVert z_0 - z_t \rVert_2$ is the Euclidean distance in latent space, $y = 0$ is a label for positive pairs and $y = 1$ for negative pairs, while $m = 1+t/T$ is a vector of margin terms, ensuring stronger separation at higher noise levels. The final loss function combines reconstruction loss and contrastive loss, ensuring both accurate denoising and improved feature separation:


\begin{equation}
\mathcal{L}_{DDAE-C} = (1-\alpha)\mathcal{L}_{\text{rec}} + \alpha \mathcal{L}_{\text{cont}},
\end{equation}

\vspace{0.1cm}

\noindent where $\alpha$ controls the trade-off between reconstruction accuracy and contrastive feature separation.

\vspace{0.1cm}

\noindent\textbf{Impact on Latent Representation.} DDAE-C promotes feature-level separation in the latent space, enhancing the model’s ability to distinguish subtle deviations from regular patterns. The diffusion variance $\beta$ controls the magnitude of noise applied during training, enabling the model to gradually increase the distance between perturbed inputs and their original counterparts. This process encourages noisy samples to occupy regions further from the core distribution of normal data. As illustrated in Fig.~\ref{fig:ddim_ssl_sketch}, normal samples tend to form a compact cluster, while noisy or structurally abnormal inputs are displaced further away. The resulting contrastive regularization sharpens decision boundaries and leads to a more structured and discriminative latent representation.

\begin{algorithm}[!]
\caption{Training of DDAE and DDAE-C Models}
\label{alg:merged_ddae}
\begin{algorithmic}[1]
\REQUIRE Input data $\mathbf{X} \in \mathbb{R}^{N \times D}$, Diffusion steps $T$, Model parameters $\theta$, Noise schedule $\beta_t$
\ENSURE Reconstructed data $\hat{\mathbf{X}}$, Anomaly scores $S(\mathbf{X})$
\STATE Initialize model parameters $\theta$
\FOR{each epoch}
    \FOR{each batch $\mathbf{x}_0 \subset \mathbf{X}$}
        \STATE Sample timesteps $t \sim \text{Uniform}(1, T)$
        \STATE Sample noise $\boldsymbol{\epsilon} \sim \mathcal{N}(0, \mathbf{I})$
        \STATE Compute noisy data: $\mathbf{x}_t = \sqrt{\bar{\alpha}_t} \mathbf{x}_0 + \sqrt{1 - \bar{\alpha}_t} \boldsymbol{\epsilon}$
        \STATE Predict $\hat{\mathbf{x}}_0$ using $f_\theta(\mathbf{x}_t, t)$
        \STATE Compute reconstruction loss  $\mathcal{L}_{\text{rec}}$
        \IF{contrastive learning is enabled (DDAE-C)}
            \STATE Generate positive ($\mathbf{x^+}$) and negative ($\mathbf{x^-}$) pairs 
            \STATE Define pairwise labels $\mathbf{y}$
            \STATE Compute margins: $m=1+\frac{t}{T}$
            \STATE Compute embeddings $\mathbf{z_0}$ using $f_\theta(\mathbf{x^+}, t)$
            \STATE Compute embeddings $\mathbf{z_t}$ using $f_\theta(\mathbf{x^-}, t)$
            \STATE Compute contrastive loss $\mathcal{L}_{\text{cont}}$
            \STATE Compute total loss: $\mathcal{L} = \mathcal{L}_{\text{rec}} + \mathcal{L}_{\text{cont}}$
        \ELSE
            \STATE Compute total loss: $\mathcal{L} = \mathcal{L}_{\text{rec}}$
        \ENDIF
        \STATE Update $\theta$ using gradient descent
    \ENDFOR
\ENDFOR
\RETURN $\hat{\mathbf{X}}$
\end{algorithmic}
\end{algorithm}

\section{Experimental Setup}
\label{sec:experimental_setup}

This section describes the datasets, training setup, baselines, and evaluation metrics used in our experiments. Our setup follows the evaluation framework of Livernoche et al.~\citep{livernoche2023diffusion}. All models were implemented using PyTorch v2.2.0~\citep{paszke2019pytorch}. 

\vspace{0.1cm}

\noindent\textbf{Datasets and Data Preparation.} We evaluate our approach on the ADBench benchmark~\citep{han2022adbench}, which comprises 47 tabular datasets along with five datasets from images and five from natural language tasks. Prior to training, all datasets are standardized to zero mean and unit variance.

\vspace{0.1cm}

\noindent\textbf{Training Setup.} We evaluate DDAE in both unsupervised and semi-supervised settings. 
\begin{itemize}
    \item \textbf{Unsupervised:} The entire dataset is used for training and evaluation, and anomaly scores are computed for all instances. The model is trained using bootstrapping across the entire dataset. 
    
    \item \textbf{Semi-supervised:} The model is trained on 50\% of the normal samples (excluding anomalies). The remaining 50\% of normal samples and all available anomalies are used for evaluation to measure detection performance.
\end{itemize}
The architectural choices for DDAE and DDAE-C were determined through extensive hyperparameter tuning. A detailed overview of grid search experiments for architecture and training hyperparameters is provided in Appendix~\ref{sec:ablation_study}.

\vspace{0.1cm}

\noindent\textbf{Baselines.} We evaluate our method against all methods in ADBench~\citep{han2022adbench} and those introduced by Livernoche et al.~\citep{livernoche2023diffusion}. These include 12 conventional,\footnote{CBLOF \cite{He2003DiscoveringCU}, COPOD \cite{Li2020COPOD}, ECOD \cite{Li2022ECOD}, HBOS \cite{Goldstein2012HistogramBasedOD}, IForest \cite{Liu2008IsolationF}, kNN \cite{Ramaswamy2000EfficientAF}, LODA \cite{Pevny2016LodaLE}, LOF \cite{Breunig2000LOF}, MCD \cite{Rousseeuw1999FastMR}, OCSVM \cite{Scholkopf2001EstimatingTS}, PCA \cite{Shyu2003NovelAF}, and Feature Bagging \cite{Lazarevic2005FeatureBF}} 10 deep learning,\footnote{DAGMM \cite{Zong2018DeepAG}, DeepSVDD \cite{Ruff2018DeepSV}, DROCC \cite{Goyal2020DROCC}, GOAD \cite{Bergman2020GOAD}, ICL \cite{Tack2020CSI}, PlanarFlow \cite{Rezende2015VariationalIW}, VAE \cite{kingma2013auto}, GANomaly \cite{Akcay2018GANomaly}, SLAD \cite{Sehwag2021SSD}, and DIF \cite{Zhao2022DIF}} and 4 diffusion\footnote{DTE-C~\citep{livernoche2023diffusion}, DTE-IG~\citep{livernoche2023diffusion}, DTE-NP~\citep{livernoche2023diffusion}, and DDPM~\citep{livernoche2023diffusion}} approaches, covering a diverse range of anomaly detection methods. This setup comprehensively compares (i) conventional, (ii) deep learning, and (iii) generative modeling approaches.

\vspace{0.1cm}

\noindent\textbf{Evaluation Metrics.} We evaluate the performance using PR-AUC (Precision-Recall AUC) and ROC-AUC (Receiver Operating Characteristic AUC). PR-AUC is favored over ROC-AUC due to its robustness in imbalanced anomaly detection tasks. The anomaly scores $S(x_0)$ are computed as the cumulative reconstruction errors across all diffusion steps $T$, as defined in \autoref{eq:ddae_anomaly_score}.
Each experiment is evaluated using five random initialization seeds, and we report mean ± standard deviation for all results.

\begin{table}[t]
  \centering
   \caption{Comparison against DTE models from~\citep{livernoche2023diffusion} and DAE. PR-AUC and ROC-AUC scores are averaged over 57 datasets from ADBench with five random seeds. DDAE consistently achieves the highest performance across settings.}
   \vspace{-0.1cm}
  \label{tab:performance_comparison}
  \resizebox{0.5\textwidth}{!}{%
  \begin{tabular}{lcc|cc} 
    \toprule 
    & \multicolumn{2}{c|}{\textbf{Unsupervised}} & \multicolumn{2}{c}{\textbf{Semi-Supervised}} \\ \cline{2-5} 
    \raisebox{0.9ex}{\textbf{Model}} & \textbf{PR-AUC} $\uparrow$ & \textbf{ROC-AUC} $\uparrow$ & \textbf{PR-AUC} $\uparrow$ & \textbf{ROC-AUC} $\uparrow$ \\ \midrule 
    DAE & 19.85 ± 2.51 & 63.73 ± 3.44 & 59.60 ± 3.18 & 81.02 ± 1.73 \\
    DDPM & 30.09 ± 2.84 & 70.01 ± 2.29 & 52.41 ± 2.26 & 77.92 ± 1.37 \\ 
    DTE-NP & 29.47 ± 1.75 & 72.15 ± 1.19 & 57.13 ± 2.67 & 81.49 ± 1.29 \\ 
    DTE-IG & 21.62 ± 4.63 & 65.95 ± 7.53 & 54.49 ± 6.17 & 77.85 ± 4.65 \\ 
    DTE-C & 28.79 ± 3.28 & 72.76 ± 3.15 & 52.02 ± 2.77 & 80.37 ± 1.43 \\ \hline 
    \textbf{DDAE-C} & 31.97 ± 0.87 & 73.72 ± 0.79 & 59.48 ± 2.55 & 81.54 ± 1.11 \\
    \textbf{DDAE} & \textbf{32.77} ± \textbf{0.69} & \textbf{74.08} ± \textbf{0.51} & \textbf{61.36} ± \textbf{2.23} & \textbf{83.17} ± \textbf{1.0} \\
     \bottomrule         
  \end{tabular}%
  }
  \vspace{-0.2cm}
\end{table}

\section{Experimental Results}
\label{sec:experimental_results}

We evaluate DDAE and DDAE-C in three key aspects: (1) anomaly detection performance compared to state-of-the-art baselines, (2) the effect of diffusion noise scheduling, and (3) the impact of contrastive learning on latent space representations.


\subsection{Comparison Against Baselines}

We compare DDAE and DDAE-C to conventional, deep learning, and diffusion-based anomaly detection models using the ADBench benchmark~\citep{han2022adbench}. Baseline results are sourced from Livernoche et al.~\citep{livernoche2023diffusion}. ~\autoref{fig:baseline_comparison_grouped} provides a visual comparison of PR-AUC scores, while detailed numerical results against diffusion-based models are in Table~\ref{tab:performance_comparison}. 
The architectural configurations for DDAE and DDAE-C were optimized through extensive grid search, ensuring fair comparison against baselines. These experiments, including variations in encoder-decoder depth, latent dimensions, and timestep embeding size are detailed in~\autoref{sec:ablation_study}.

\begin{figure}[t]
    \centering
    \includegraphics[width=1\linewidth]{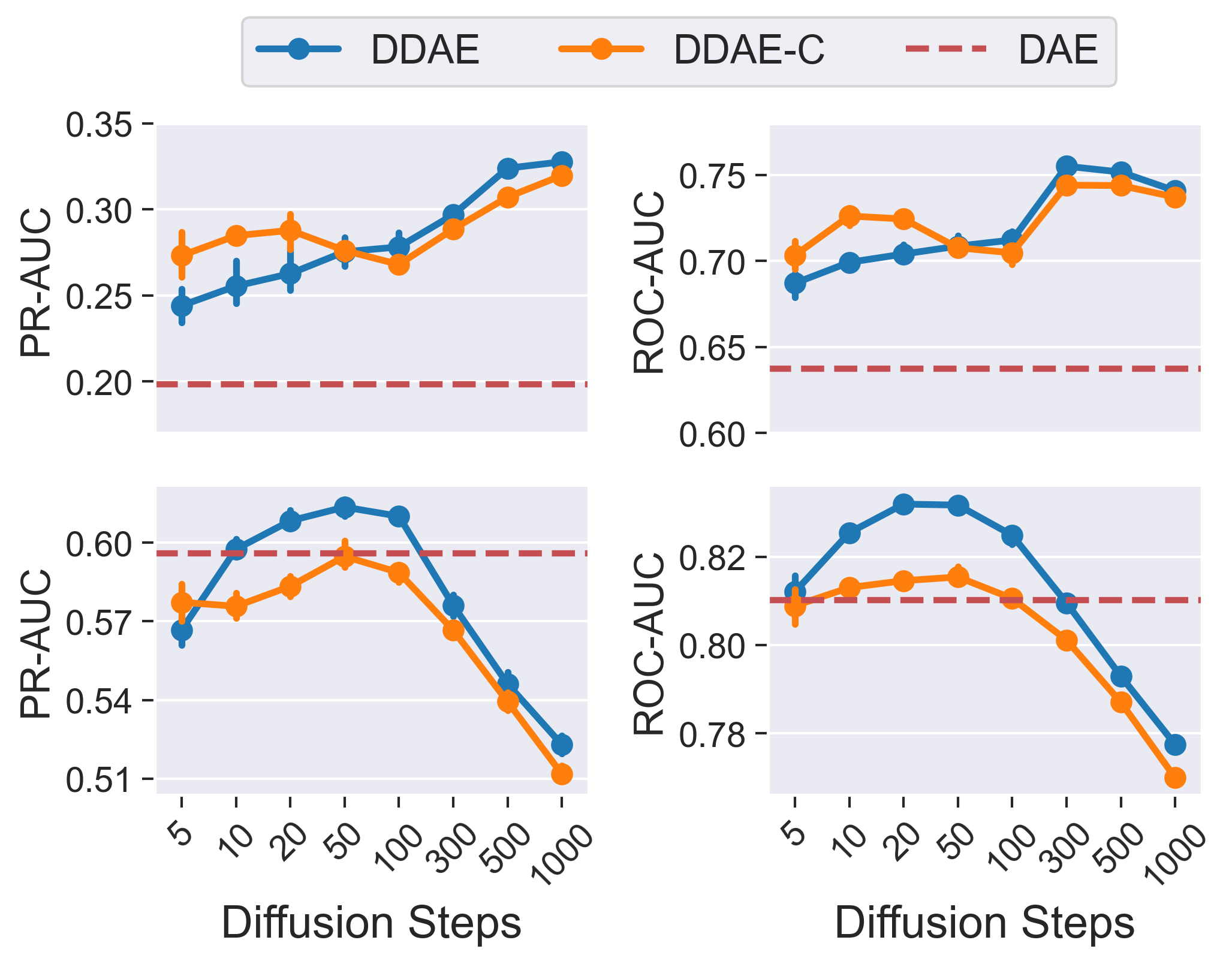}
    \vspace{-0.5cm}
  \caption{Effect of diffusion steps on anomaly detection, shown as PR-AUC and ROC-AUC scores for DDAE, DDAE-C, and the baseline DAE (dashed line) in unsupervised (top row) and semi-supervised (bottom row) settings. In unsupervised learning, performance improves with more steps, while in semi-supervised learning, it peaks around 50-100 steps before degrading due to excessive noise.}
    \label{fig:grid_diffusion_steps}
    \vspace{-0.3cm}
\end{figure}

\vspace{0.1cm}

\noindent \textbf{Comparison to DAE.} To assess the impact of diffusion-based noise scheduling, we compare DDAE to its baseline DAE. In the unsupervised setting, DDAE improves PR-AUC by 65.1\% and ROC-AUC by 16.2\%, while in the semi-supervised setting, it shows a 3\% improvement in PR-AUC and 2.7\% in ROC-AUC. These gains indicate that the incorporation of diffusion scheduling enhances anomaly detection performance, particularly in fully unsupervised scenarios, where progressive noise injection helps refine the model’s ability to distinguish anomalies from normal instances.

\vspace{0.1cm}

\noindent \textbf{Comparison to Diffusion Models.} We further compare DDAE and DDAE-C to prior diffusion-based anomaly detection models (\autoref{tab:performance_comparison}). In the unsupervised setting, DDAE outperforms DDPM, the strongest diffusion-based model in this setting, by 8.9\% in PR-AUC and 5.8\% in ROC-AUC, while DDAE-C achieves 6.3\% and 3.9\% improvements, respectively. In the semi-supervised setting, DDAE surpasses DTE-NP, the best-performing diffusion-based model, by 7.4\% in PR-AUC and 2.6\% in ROC-AUC. These results indicate that incorporating an autoencoder-based denoising process offers a stronger anomaly detection framework compared to diffusion models relying solely on iterative noise removal. The ability to reconstruct original data rather than predict noise likely contributes to better generalization, particularly in tabular anomaly detection tasks.

\vspace{0.1cm}

\noindent \textbf{Comparison to State-of-the-Art Models.} A broader comparison with all anomaly detection methods (\autoref{fig:baseline_comparison_grouped}
) reveals that DDAE and DDAE-C outperform all competing models in the semi-supervised setting and achieve competitive performance in unsupervised settings. Specifically, in unsupervised learning, DDAE ranks 4th in PR-AUC and 2nd in ROC-AUC, outperforming all deep learning-based approaches. This highlights the effectiveness of diffusion-scheduled denoising in feature separation, particularly when labeled anomalies are available.

\vspace{0.15cm}

These results validate the effectiveness of diffusion-based noise scheduling in anomaly detection. The autoencoder-based reconstruction process in DDAE provides clear advantages over existing diffusion-based approaches, while DDAE and DDAE-C set a new benchmark in semi-supervised anomaly detection.

\begin{figure}[t]
  \centering
  \includegraphics[width=\linewidth]{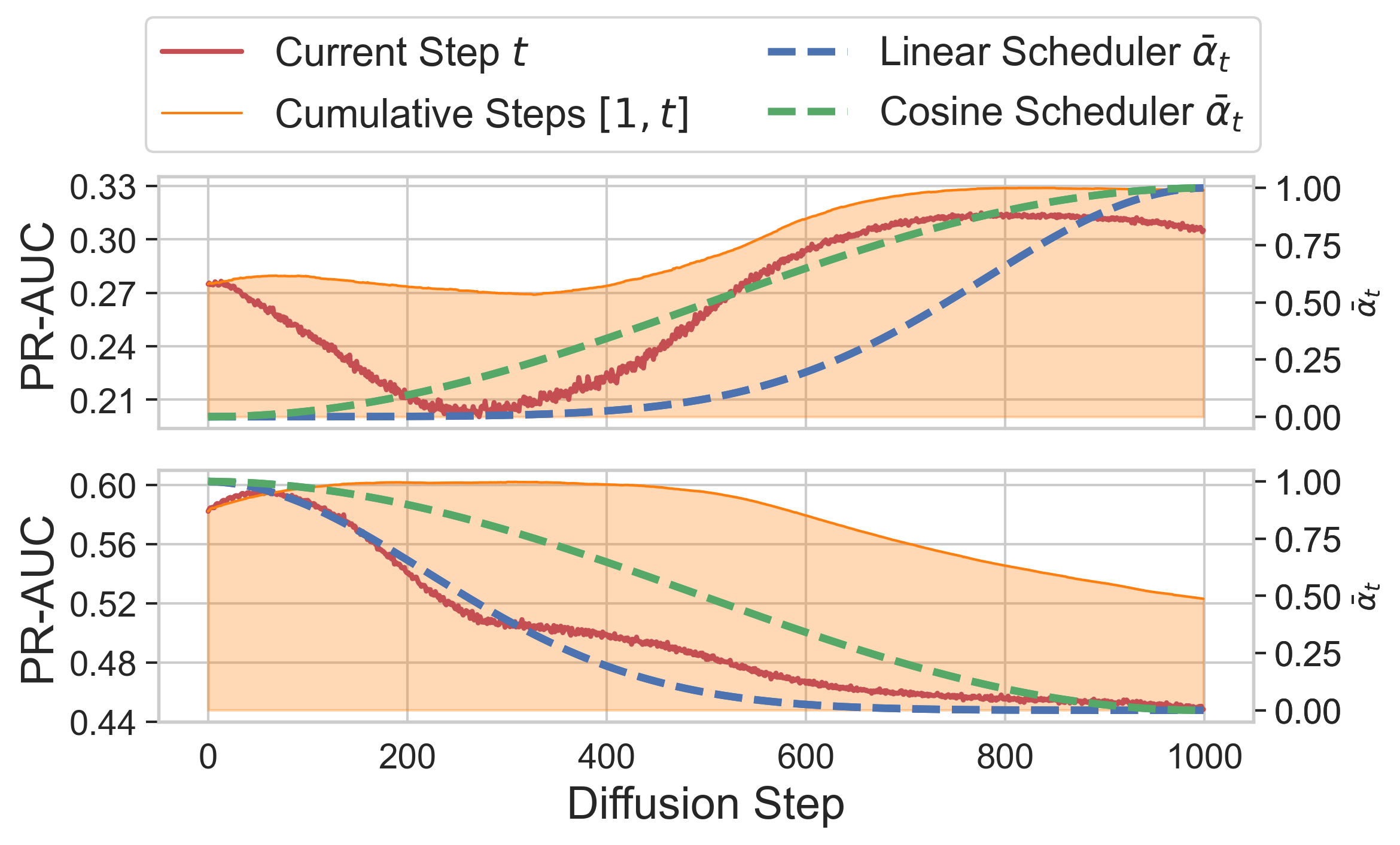}
  \vspace{-0.5cm}
  \caption{Anomaly scoring across diffusion steps. PR-AUC scores in unsupervised (top) and semi-supervised (bottom) settings, averaged over all datasets at each step $T \in \{ 1,...,1000 \}$. The red line shows per-step performance, the orange line tracks cumulative performance, and dashed lines represent the signal strength ($\hat{\alpha}_t$) for linear and cosine schedulers. In unsupervised settings, later steps $T \in \{500, \ldots, 1000 \}$ act as a regularizer, while in semi-supervised settings, earlier steps $T \in \{5, \ldots ,500 \}$ are more influential, peaking at $T \in \{ 50, \ldots, 100 \}$.}
  \vspace{-0.3cm}
  \label{fig:noise_analysis}
\end{figure}

\vspace{-0.3cm}

\subsection{Effect of Noise Scheduling}

\begin{figure}[t]
    \centering
    \includegraphics[width=1\linewidth]{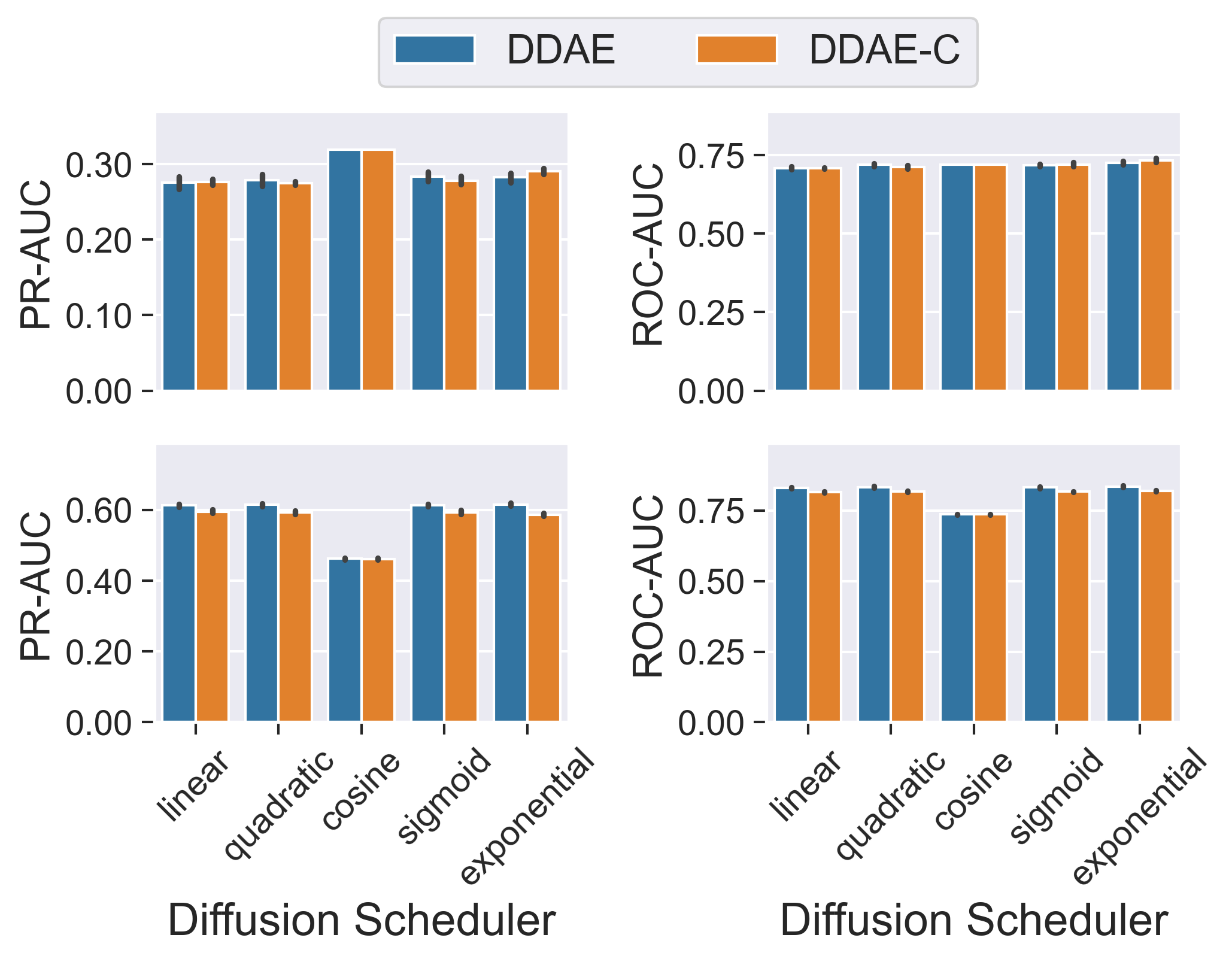}
    \vspace{-0.5cm}
    \caption{Effect of diffusion schedulers on anomaly detection, shown as PR-AUC and ROC-AUC scores for DDAE and DDAE-C in unsupervised (top row) and semi-supervised (bottom row) settings. The cosine scheduler performs best in the unsupervised setting but shows the worst performance in the semi-supervised setting, highlighting its suitability for fully unsupervised anomaly detection.}
    \label{fig:grid_diffusion_schedulers}
    \vspace{-0.3cm}
\end{figure}

To investigate the role of noise scheduling in anomaly detection, we conduct a series of experiments evaluating different diffusion step configurations and noise schedulers. First, we analyze the impact of varying number of diffusion steps on detection performance (\autoref{fig:grid_diffusion_steps}). Next, we examine the correlation between diffusion schedulers and model performance across training settings (\autoref{fig:noise_analysis}). Finally, we compare five noise scheduling strategies to determine the most effective approach for each learning paradigm (\autoref{fig:grid_diffusion_schedulers}). These experiments provide insights into how noise addition affects feature separation and model generalization in both unsupervised and semi-supervised anomaly detection.

\vspace{0.1cm}

\noindent \textbf{Diffusion Steps Analysis.} To assess the impact of noise scheduling, we evaluate performance across different diffusion steps $T \in \{5, 10, 20, 50, 100, 300, 500, 1000\}$ in unsupervised and semi-supervised training. As shown in \autoref{fig:grid_diffusion_steps}, in the unsupervised setting, performance steadily improves with increasing diffusion steps, suggesting that higher noise magnitudes act as a regularizer, enhancing model robustness against overfitting. In contrast, in semi-supervised case, performance peaks at $T \in \{50, \ldots, 100 \}$, indicating that excessive noise disrupts useful feature representations.

\vspace{0.1cm}

This trend aligns with the decision boundary visualizations in \autoref{fig:decision_boundary}, where lower noise magnitudes (a) help preserve the data manifold, improving generalization in semi-supervised settings. Conversely, larger noise magnitudes (b) act as an implicit regularizer. They prevent the model from overfitting to anomalies and instead strengthen its ability to learn normal patterns. This effect is particularly beneficial in unsupervised training.

\vspace{0.1cm}

\noindent \textbf{Anomaly Scoring Analysis.} Since the anomaly score function (\autoref{eq:ddae_anomaly_score}) aggregates reconstruction errors over all diffusion timesteps, detection performance varies per timestep. To analyze this, we train a DDAE model with $T$ diffusion timesteps and collect anomaly scores at individual timesteps $t \in \{1,\ldots,T\}$. \autoref{fig:noise_analysis} illustrates per-step anomaly scores (red line), cumulative anomaly score (orange line), and the signal strength $\hat{\alpha}_t$ for linear and cosine schedulers. 

The results reveal distinct patterns across training paradigms. In the unsupervised training, later diffusion steps $T \in \{500,\ldots,1000\}$ contribute most to anomaly detection, suggesting that extremely noisy inputs act as a regularizer, preventing the model from overfitting anomalies. In contrast, semi-supervised training relies more on early diffusion steps $T \in \{5,\ldots,500\}$, with peak performance around $T \in \{50, \ldots, 100\}$ before degradation occurs due to excessive noise interference. Notably, performance saturates beyond $T=500$, where the signal-to-noise ratio $\hat{\alpha}_t$ vanishes, leaving only Gaussian noise. This explains why unsupervised models benefit from high noise levels, while in semi-supervised case, smaller noise magnitudes allow the model to better capture the data manifold.

\vspace{0.1cm}

\noindent \textbf{Diffusion Schedulers Analysis.} To evaluate the impact of different noise scheduling strategies on anomaly detection, we compare five schedulers: linear, cosine, quadratic, sigmoid, and exponential. These schedulers control the rate at which noise is injected during training, influencing the model’s ability to separate normal and anomalous instances. We evaluate their effectiveness in both training settings using PR-AUC and ROC-AUC across all datasets.

\vspace{0.1cm}

\autoref{fig:grid_diffusion_schedulers} shows that the cosine scheduler performs best in unsupervised learning but worst in semi-supervised settings. In contrast, the linear scheduler provides stable performance across both settings, slightly excelling in semi-supervised learning. This aligns with \autoref{fig:noise_analysis}, where in unsupervised training, cosine scheduling exhibits a stronger correlation with anomaly detection performance, likely due to its smooth, non-linear decay preserving structural information. In semi-supervised training, the linear scheduler aligns better with model performance, suggesting that a gradual and uniform noise reduction aids in learning well-defined decision boundaries. These findings emphasize that progressive noise decay benefits unsupervised anomaly detection, while controlled, linear noise reduction enhances semi-supervised learning.

\begin{figure}[t]
    \centering
    \includegraphics[width=1\linewidth]{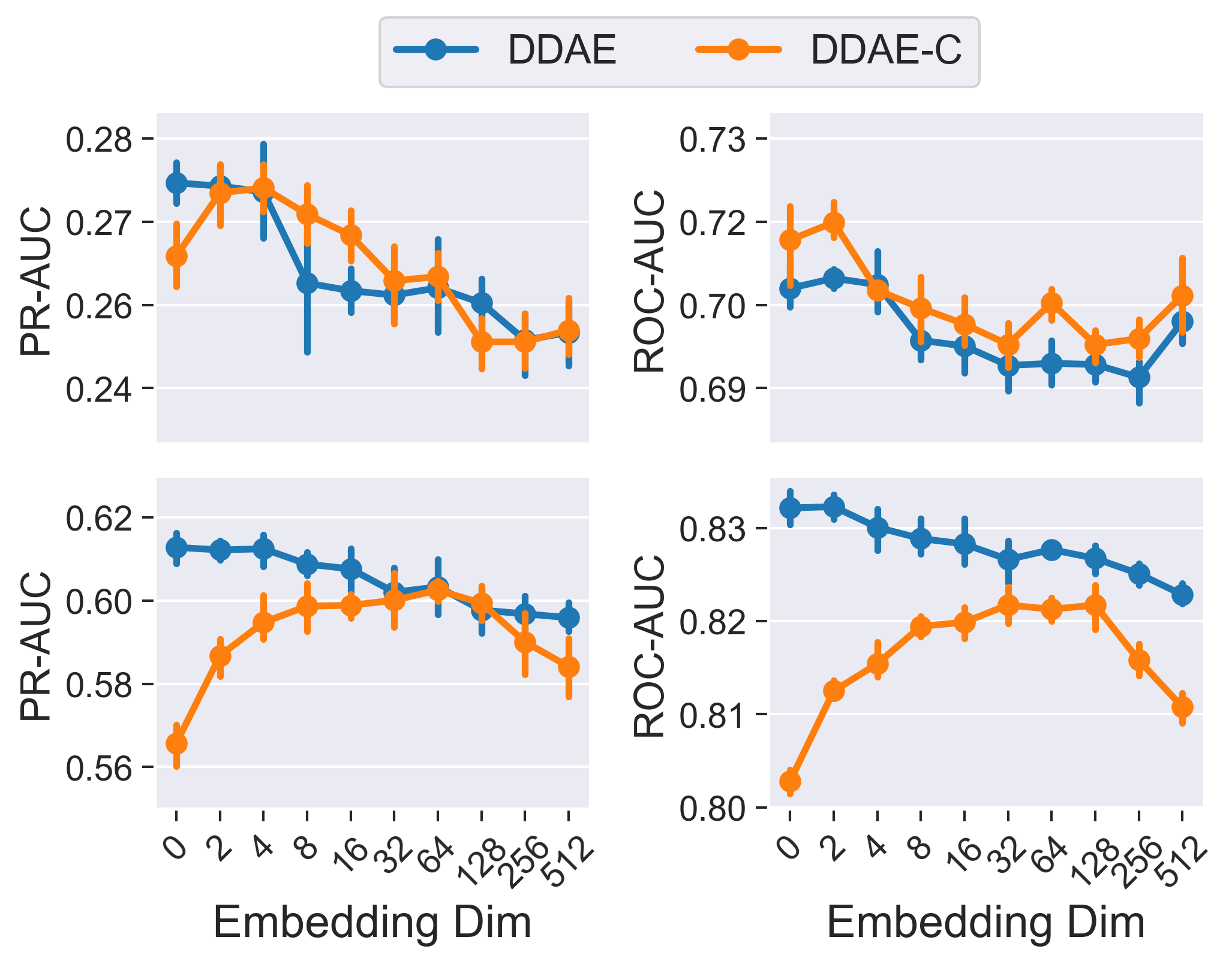}
    \vspace{-0.5cm}
    \caption{Effect of diffusion timestep embedding dimension on anomaly detection, shown as PR-AUC and ROC-AUC scores for DDAE and DDAE-C in unsupervised (top row) and semi-supervised (bottom row) settings. In the unsupervised case, larger embeddings degrade performance, while in semi-supervised settings, DDAE-C benefits from moderate sizes (4–32) but declines at higher/smaller values, emphasizing the need for careful embedding selection.}
    \label{fig:grid_diffusion_timestep_dim}
    \vspace{-0.3cm}
\end{figure}

\begin{figure*}[t!]
  \centering
  \begin{subfigure}[b]{0.24\linewidth}
    \centering
    \includegraphics[width=\linewidth]{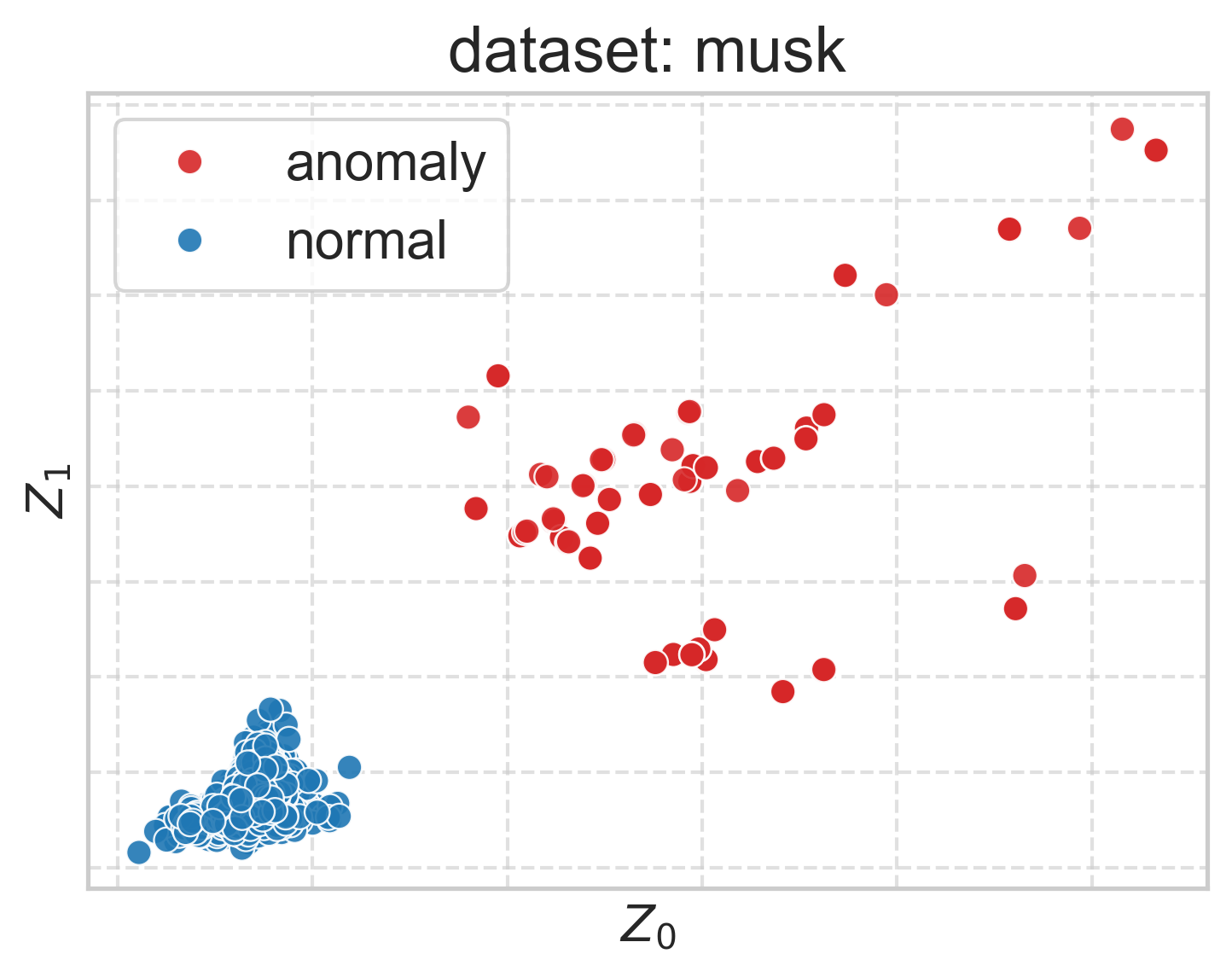}
    \label{fig:latent_data1}
  \end{subfigure}
  \begin{subfigure}[b]{0.24\linewidth}
    \centering
    \includegraphics[width=\linewidth]{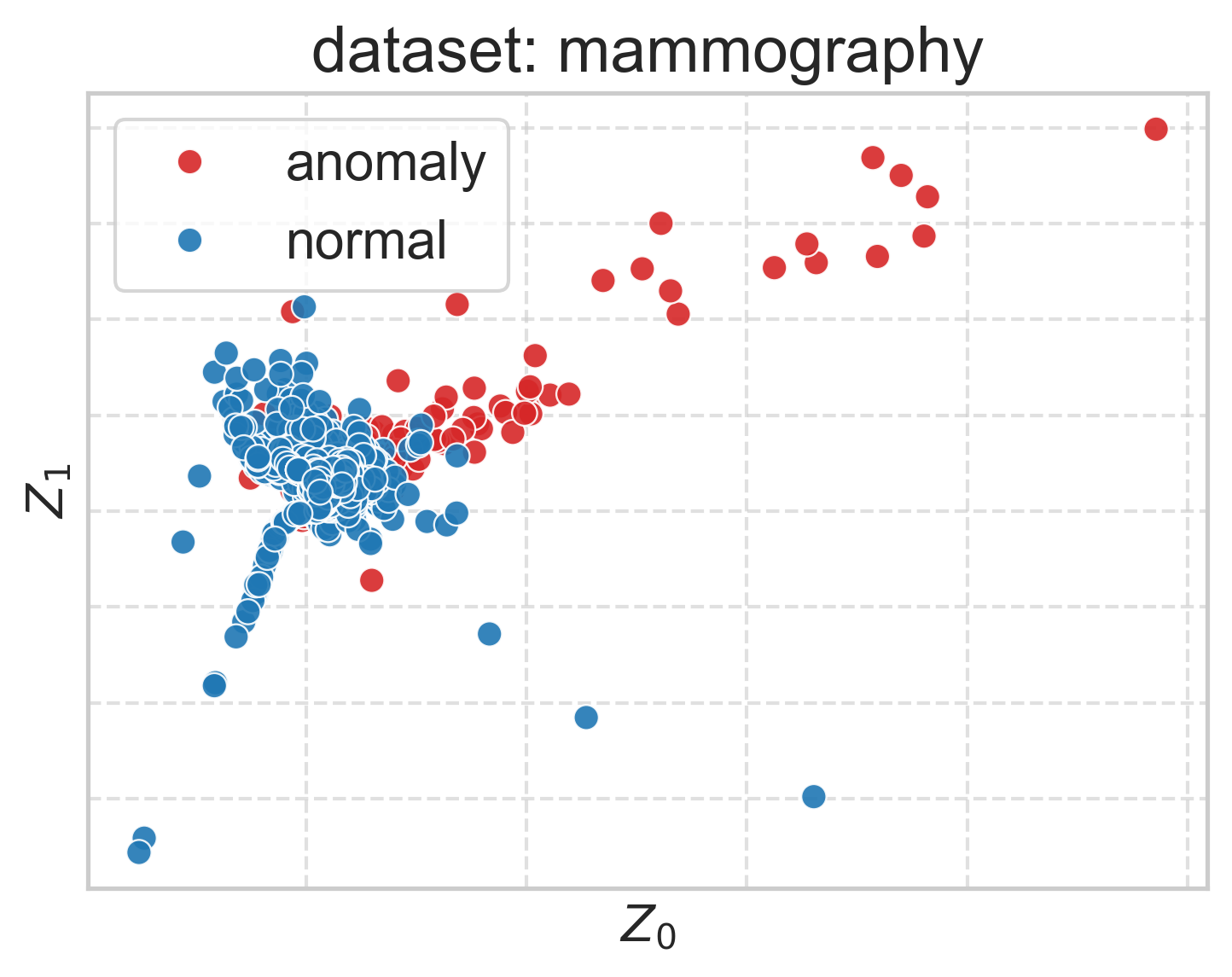}
    \label{fig:latent_data2}
  \end{subfigure}
  \begin{subfigure}[b]{0.24\linewidth}
    \centering
    \includegraphics[width=\linewidth]{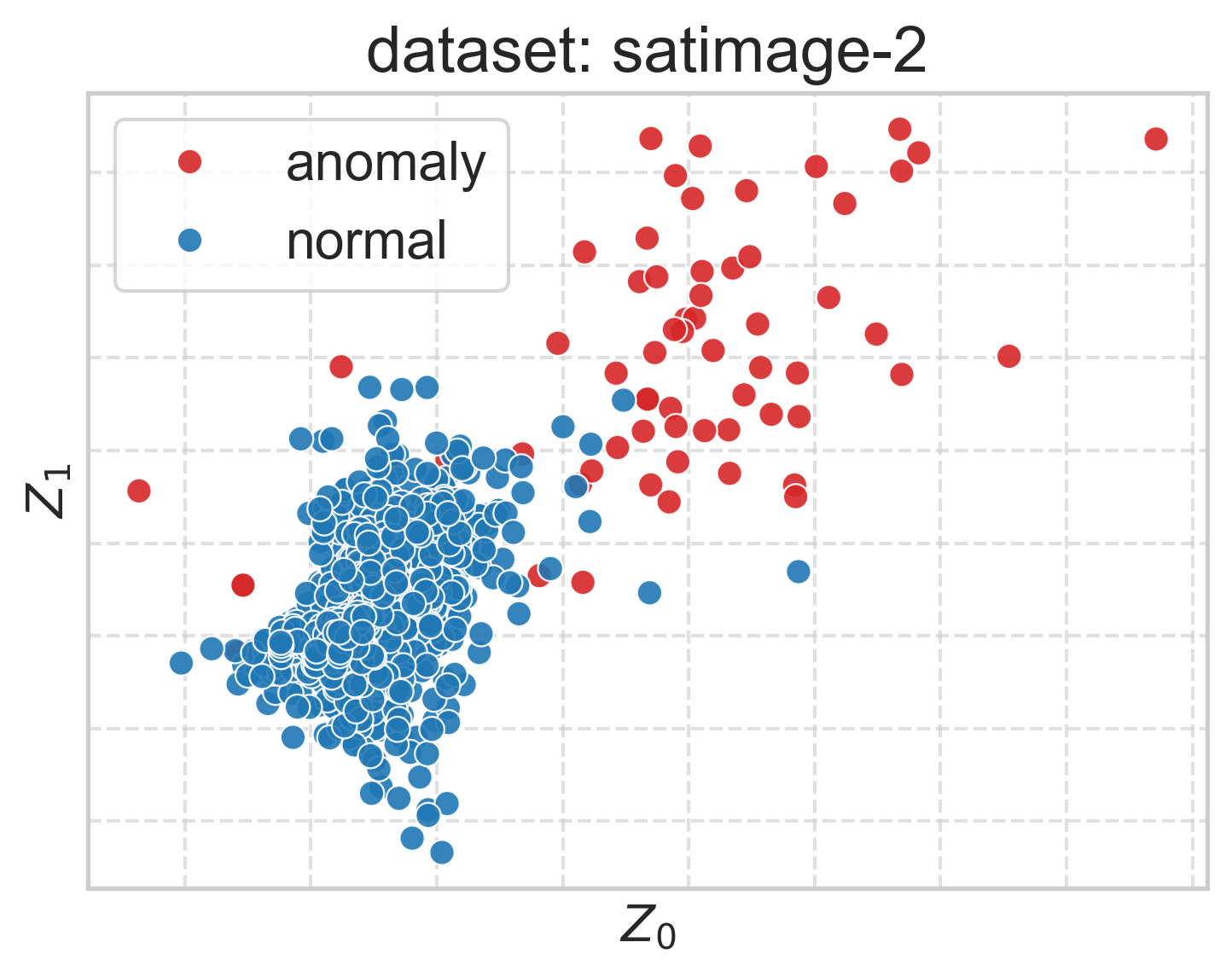}
    \label{fig:latent_data3}
  \end{subfigure}
  \begin{subfigure}[b]{0.24\linewidth}
    \centering
    \includegraphics[width=\linewidth]{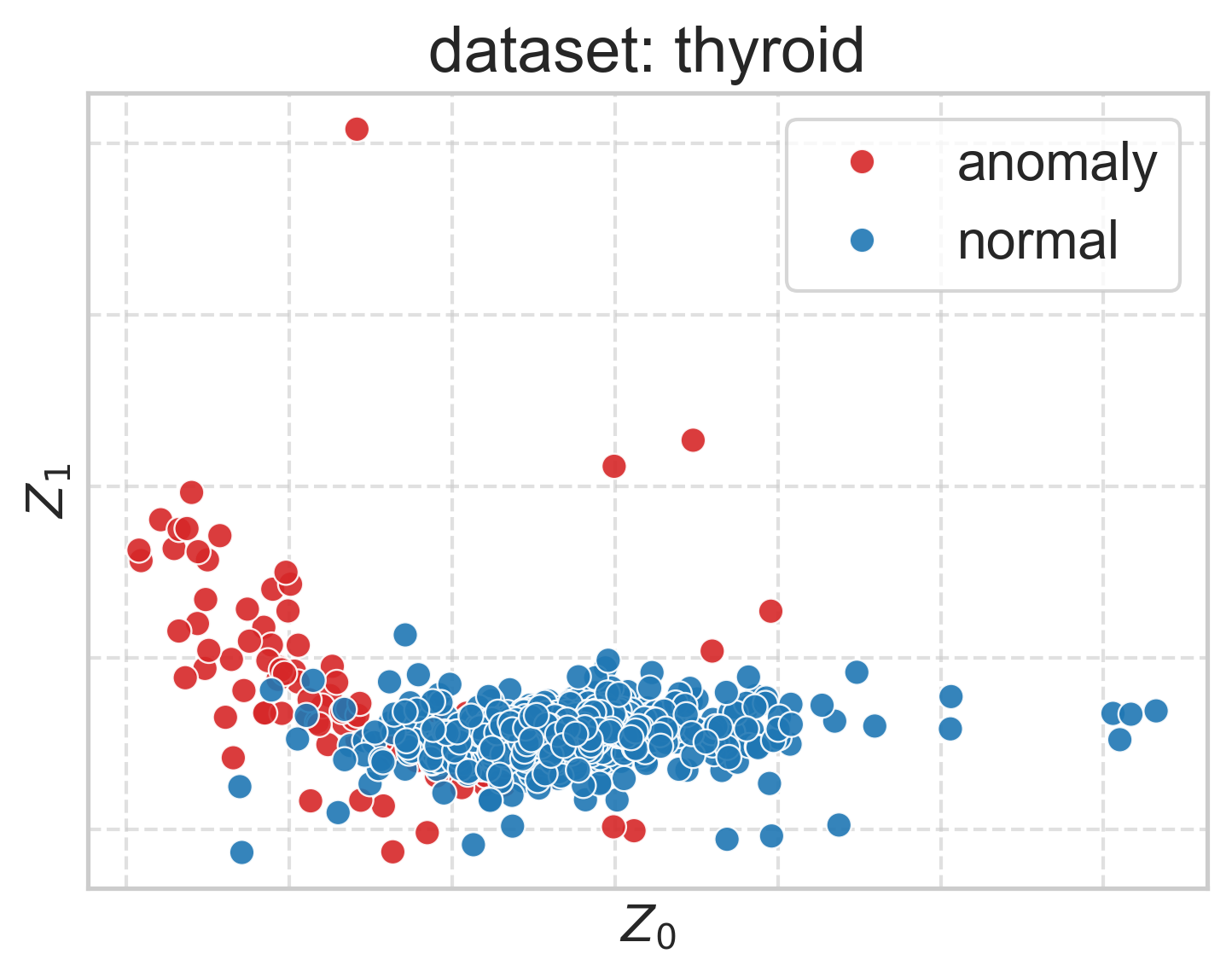}
    \label{fig:latent_data4}
  \end{subfigure}  

  \vspace{-0.2cm}
  
    \begin{subfigure}[b]{0.24\linewidth}
    \centering
    \includegraphics[width=\linewidth]{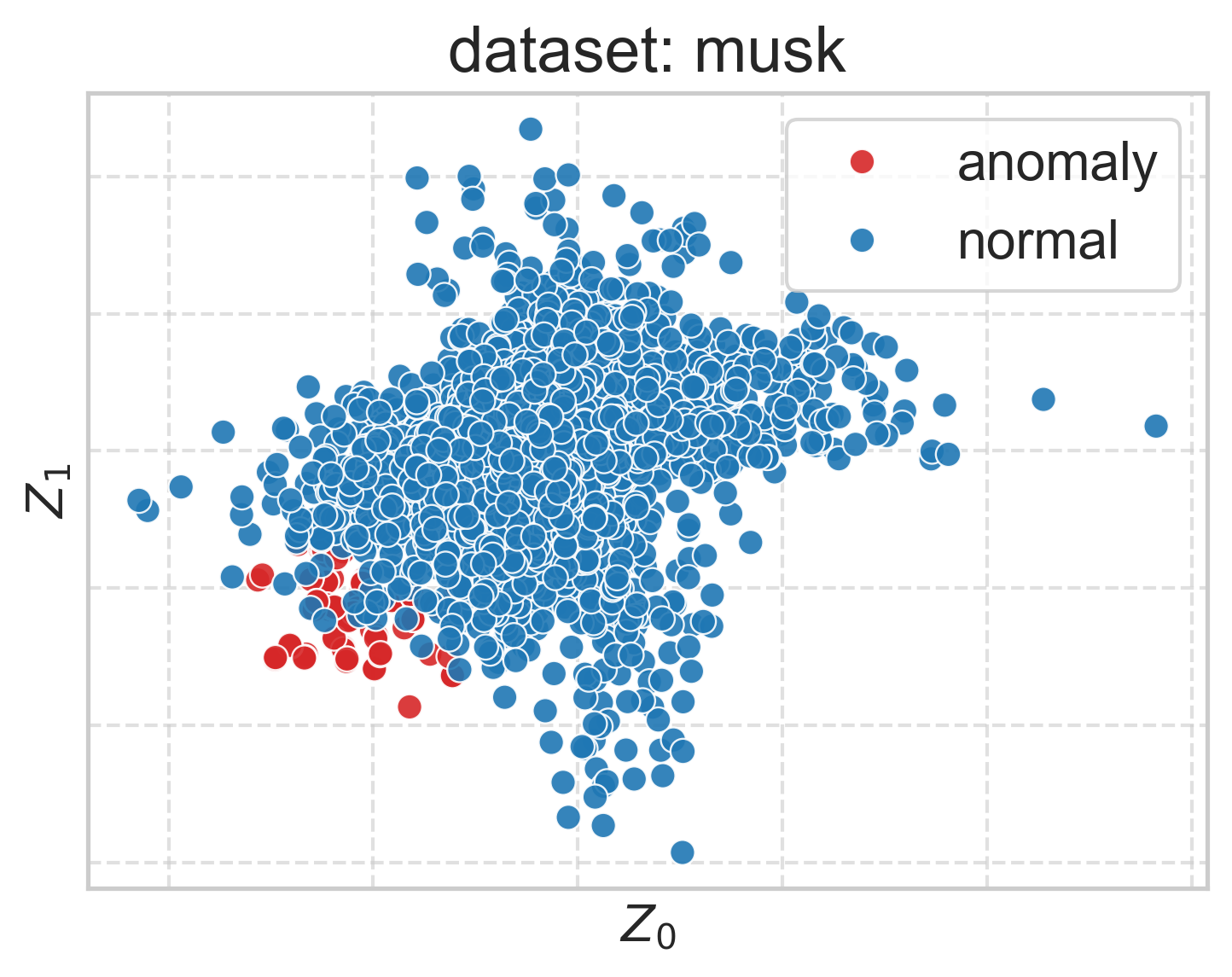}
    \label{fig:latent_data5}
  \end{subfigure}
  \begin{subfigure}[b]{0.24\linewidth}
    \centering
    \includegraphics[width=\linewidth]{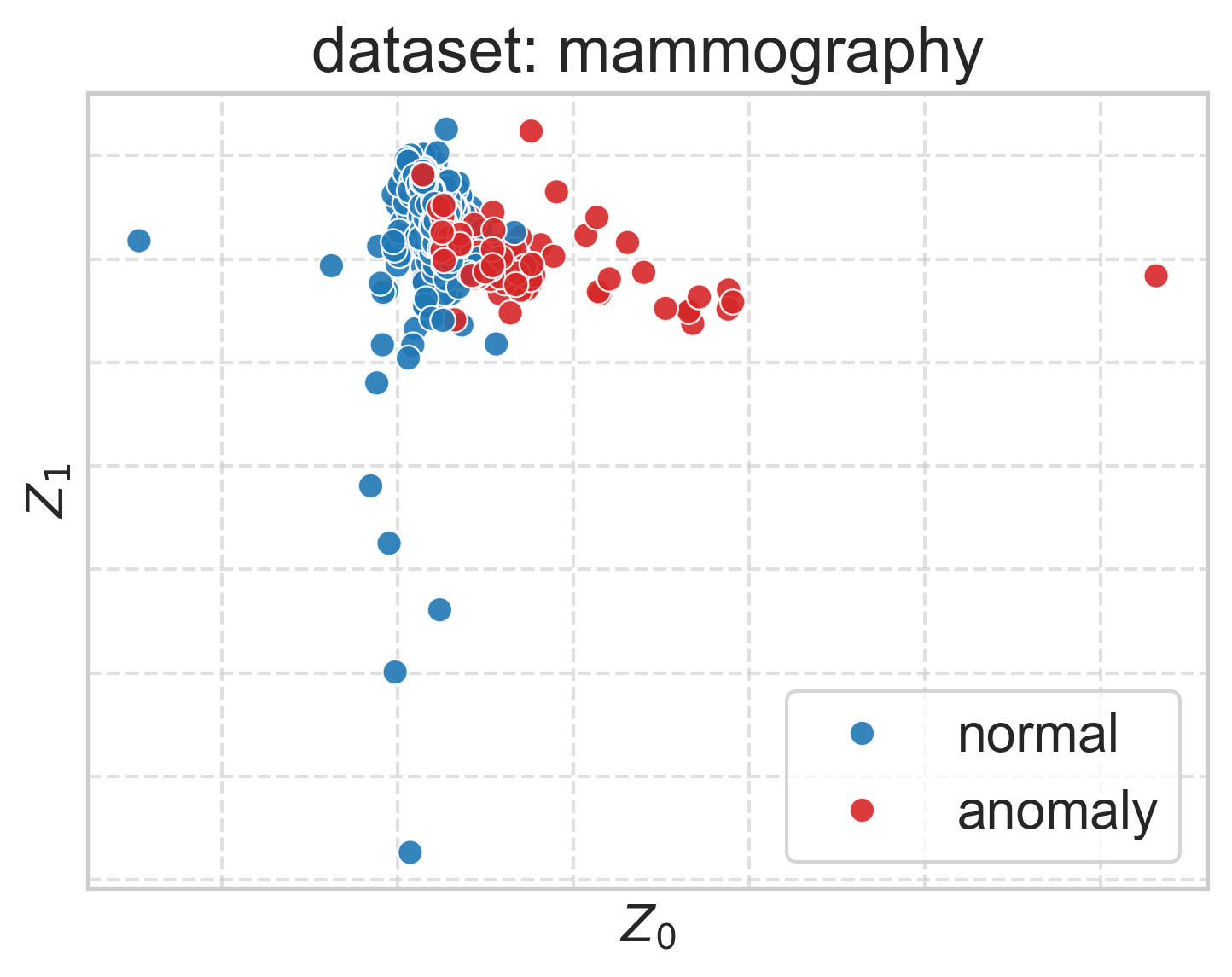}
    \label{fig:latent_data6}
  \end{subfigure}
  \begin{subfigure}[b]{0.24\linewidth}
    \centering
    \includegraphics[width=\linewidth]{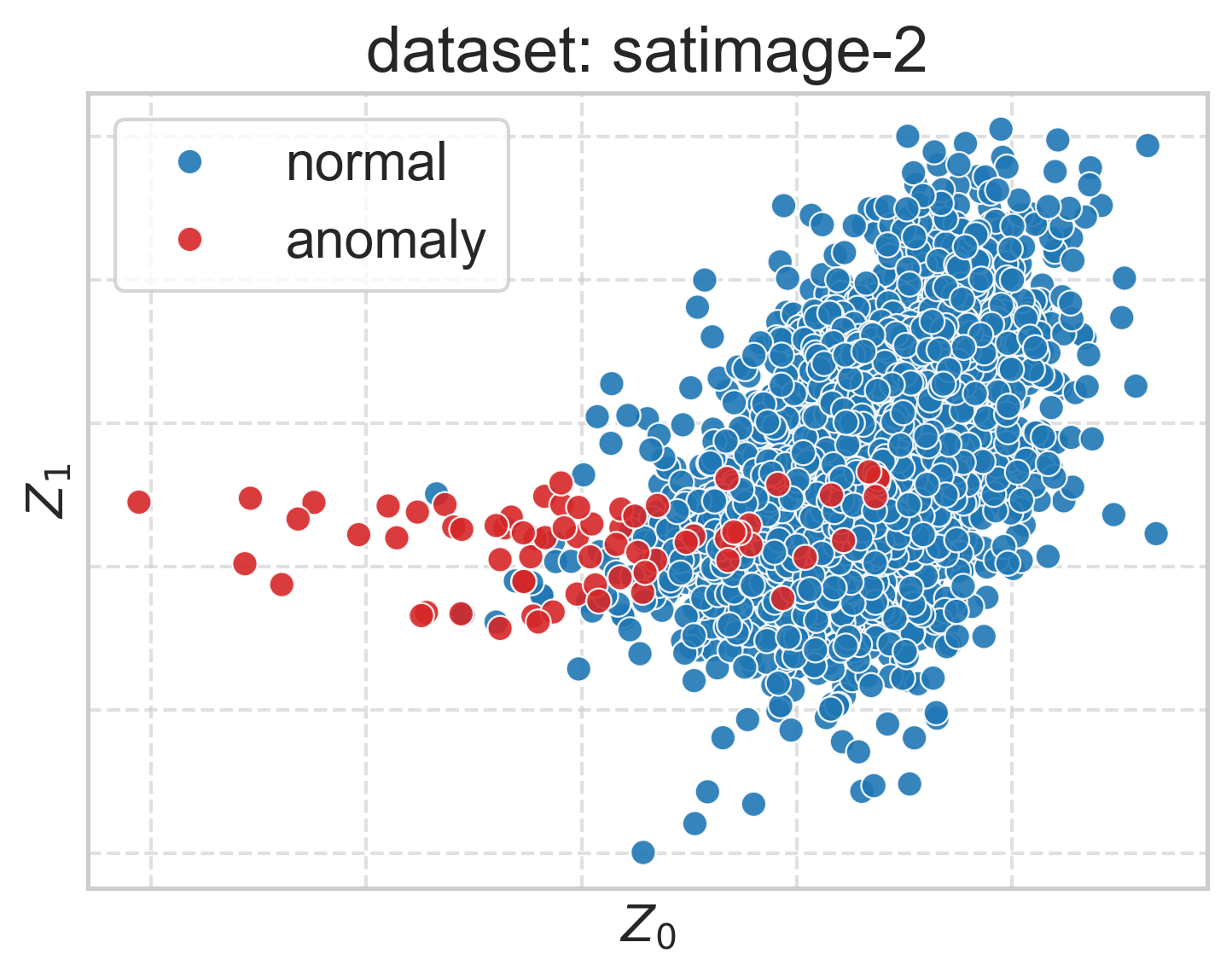}
    \label{fig:latent_data7}
  \end{subfigure}
  \begin{subfigure}[b]{0.24\linewidth}
    \centering
    \includegraphics[width=\linewidth]{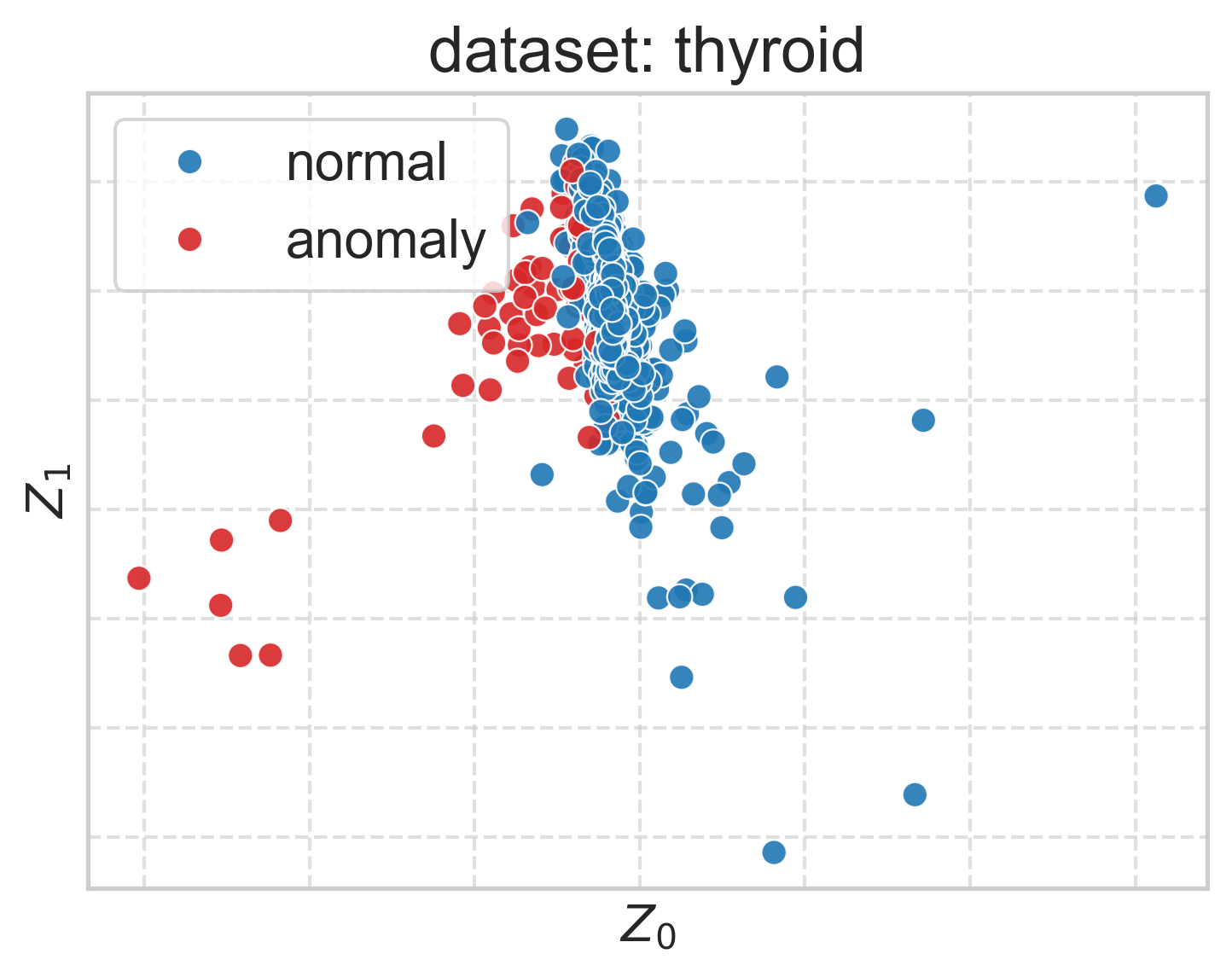}
    \label{fig:latent_data8}
  \end{subfigure}  
  
  \vspace{-0.5cm}
 \caption{Latent space projections from DDAE-C trained with a 2D compressed latent dimension $\mathbf{z} \in \mathbb{R}^2$, shown for unsupervised (top) and semi-supervised (bottom) settings. Normal samples (blue) and anomalies (red) are visualized across four datasets: \textit{musk}, \textit{mammography}, \textit{satimage-2}, and \textit{thyroid}. Normal samples tend to cluster, while anomalies are pushed away, demonstrating the effect of contrastive learning. The degree of separation varies across datasets, reflecting differences in data structure and anomaly characteristics.}
  \vspace{-0.3cm}
  \label{fig:latent_repr_data}
\end{figure*}

\vspace{0.1cm}

\noindent\textbf{Diffusion Timestep Embedding Analysis.} Timestep embedding serves as a control of the contribution of temporal information to the diffusion scheduler. We evaluate how its dimensionality affects anomaly detection performance in unsupervised and semi-supervised settings. We vary the embedding dimension from 0 (no embedding) to 512, and evaluate models using PR-AUC and ROC-AUC. A value of 0 completely disables the timestep conditioning.

\vspace{0.1cm}

Across both settings (\autoref{fig:grid_diffusion_timestep_dim}), we observe that low to moderate embedding dimensions (e.g., 4–32) yield the best performance. In the unsupervised case, increasing the dimension consistently degrades both DDAE and DDAE-C, suggesting that high-dimensional timestep encodings may introduce noise and hinder generalization. In the semi-supervised case, DDAE peaks at intermediate dimensions, while DDAE-C benefits from dimensions around 8–32 before declining at the extremes. This pattern highlights the importance of expressive but not overly complex temporal encodings, particularly when contrastive learning is used. We hypothesize that a trade-off between temporal conditioning capacity and representation noise influences the observed trends. Small embeddings may under-condition the model, leading to poor discrimination, while overly large embeddings may introduce irrelevant variance that disrupts latent structure. This effect is particularly pronounced in the semi-supervised setting, where contrastive learning depends on stable feature geometry. Understanding how embedding dimensionality shapes anomaly decision boundaries remains an interesting direction for future work.

\vspace{0.15cm}

The experiments highlight the critical role of noise scheduling in anomaly detection. Unsupervised models benefit from higher noise magnitudes, leveraging excessive noise as an implicit regularizer, while semi-supervised models perform best with moderate noise levels that preserve feature structure. The choice of scheduler also impacts performance—cosine scheduling is optimal for unsupervised learning, whereas linear scheduling is better suited for semi-supervised settings. These findings underline the importance of tailoring noise scheduling strategies based on the training paradigm to maximize anomaly detection performance.

\subsection{Latent Representation Analysis}

To understand how the contrastive DDAE-C model structures data in the latent space, we visualize the learned representations of four datasets (\autoref{fig:latent_repr_data}). 
Each scatter plot illustrates the separation between normal and anomalous instances in a 2D latent space. The results show distinct clustering behavior. In the \textit{musk} dataset, anomalies are well-separated from normal instances, indicating clear decision boundaries. The \textit{mammography} and \textit{thyroid} datasets present a more interconnected structure, suggesting that some anomalies share similarities with normal data. In the \textit{satimage-2} dataset, the model achieves a good separation, but a few normal points are scattered among anomalies, potentially affecting detection accuracy.

\vspace{0.1cm}

These findings suggest that DDAE-C effectively learns meaningful latent representations, varying separability based on dataset characteristics. While contrastive learning enhances anomaly discrimination, its effectiveness depends on the inherent structure of the dataset.

\subsection{Summary of Key Findings}

Our analysis of DDAE and DDAE-C for tabular anomaly detection reveals key insights across learning paradigms:

\vspace{0.1cm}

\noindent\textbf{Unsupervised Anomaly Detection:}

\begin{itemize}
    \item \textbf{Leading Performance}: DDAE achieves top scores among diffusion methods and outperforms most deep learning and conventional models on ADBench.
    \item \textbf{Higher Diffusion Steps}: Increasing diffusion steps \\ $T \in \{500, \dots, 1000\}$ improve performance acting as a regularizer, reducing overfitting to anomalies.
    \item \textbf{Cosine Scheduling}: Cosine scheduling improves learning by maintaining long-range dependencies through non-linear decay, proving optimal for unsupervised settings.
    \item \textbf{Feature Separation}: DDAE-C improves feature separation in the latent space, effectively distinguishing anomalies from normal samples via contrastive learning.
\end{itemize}

\vspace{0.3cm}

\noindent\textbf{Semi-Supervised Anomaly Detection:}
\begin{itemize}
    \item \textbf{New State-of-the-Art}: DDAE and DDAE-C achieve state-of-the-art results, setting a new benchmark in semi-supervised anomaly detection.
    \item \textbf{Moderate Noise Levels}: Moderate noise levels \\ $T \in \{50, \dots, 100\}$ yield the best performance, as excessive noise disrupts useful feature representations.
    \item \textbf{Linear Scheduling}: Linear noise scheduling performs best by providing gradual noise decay, which aligns well with semi-supervised learning requirements.
    \item \textbf{Latent Representation}: DDAE-C leverages contrastive learning to push anomalies away from normal samples, thereby enhancing feature separation.
\end{itemize}

\section{Conclusion and Future work}
\label{sec:conclusion}

We propose DDAE and its contrastive variant DDAE-C, which combine denoising autoencoders with diffusion-scheduled noise addition for anomaly detection. Our approach leverages controlled noise injection to enhance feature separation by gradually perturbing input samples during training. This leads to improved detection performance in unsupervised and semi-supervised settings.

Extensive experiments on 57 tabular datasets demonstrate that DDAE surpasses denoising autoencoders and diffusion-based models, particularly excelling in semi-supervised anomaly detection. DDAE-C further improves representation learning, enforcing stronger anomaly separability in the latent space. Analysis of noise scheduling reveals that higher diffusion steps benefit unsupervised training by acting as a regularizer, while moderate noise magnitudes and linear scheduling enhance semi-supervised learning by preserving decision boundaries.

These findings underscore the effectiveness of diffusion-scheduled reconstruction for anomaly detection. DDAE-C shows strong promise, and refining its contrastive learning through more principled pair construction and adaptive loss functions may yield consistent improvements. Future work could also explore adaptive noise scheduling to better fit dataset-specific characteristics, and extend DDAE to handle categorical attributes as a separate modality for richer representations in heterogeneous tabular data.

\begin{acks}
The authors thank the members of the Deutsche Bundesbank for
their valuable review and comments. Marco's research was partly funded by a fellowship (No. 57515245) from the German Academic Exchange Service (DAAD). The opinions expressed in this work are those of the authors and do not necessarily reflect the views of the Deutsche Bundesbank or the Swiss Federal Audit Office. 
\end{acks}

\bibliographystyle{ACM-Reference-Format}
\bibliography{bibliography}


\appendix

\section{Ablation Study}
\label{sec:ablation_study}

\begin{figure}[h]
    \centering
    \includegraphics[width=1\linewidth]{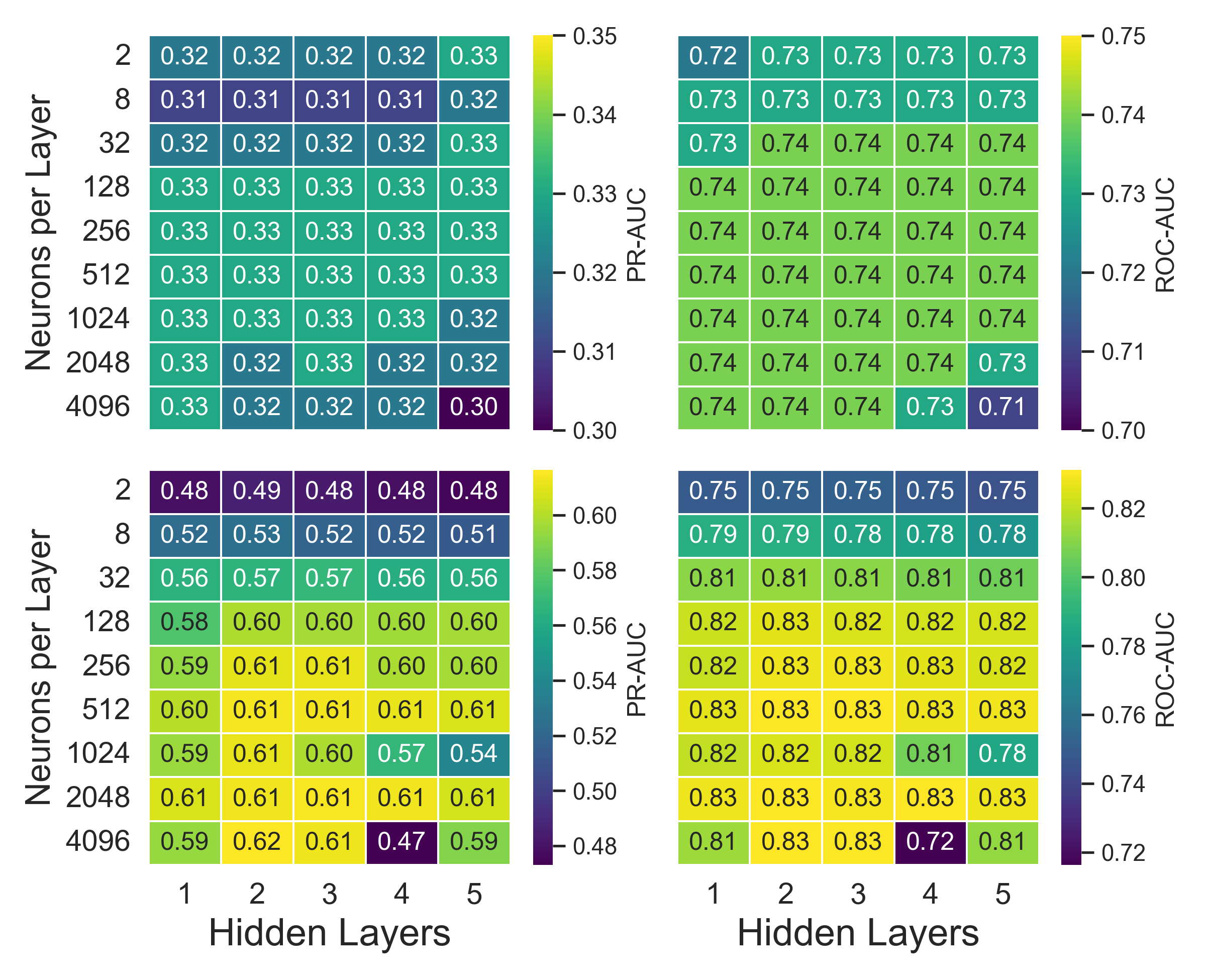}
  \caption{Network architecture search for DDAE in unsupervised (top row) and semi-supervised (bottom row) settings by varying the number of hidden layers and neurons per layer.}
    \label{fig:grid_architecture_size}
\end{figure}

\noindent \textbf{Network Architecture.} \autoref{fig:grid_architecture_size} shows PR-AUC and ROC-AUC heatmaps for DDAE architectures, varying hidden layers (1–5) and neurons per layer (2–4096). Results are averaged over 57 datasets and five random seeds. Performance improves with wider networks, with optimal results at $\geq$512 neurons per layer. Depth beyond two layers yields marginal gains, indicating that width contributes more to model effectiveness than depth. This trend holds across both metrics, suggesting that a 2–3 layer architecture with wide hidden layers offers a strong trade-off between accuracy and efficiency.

\begin{figure}[t]
    \centering
    \includegraphics[width=1\linewidth]{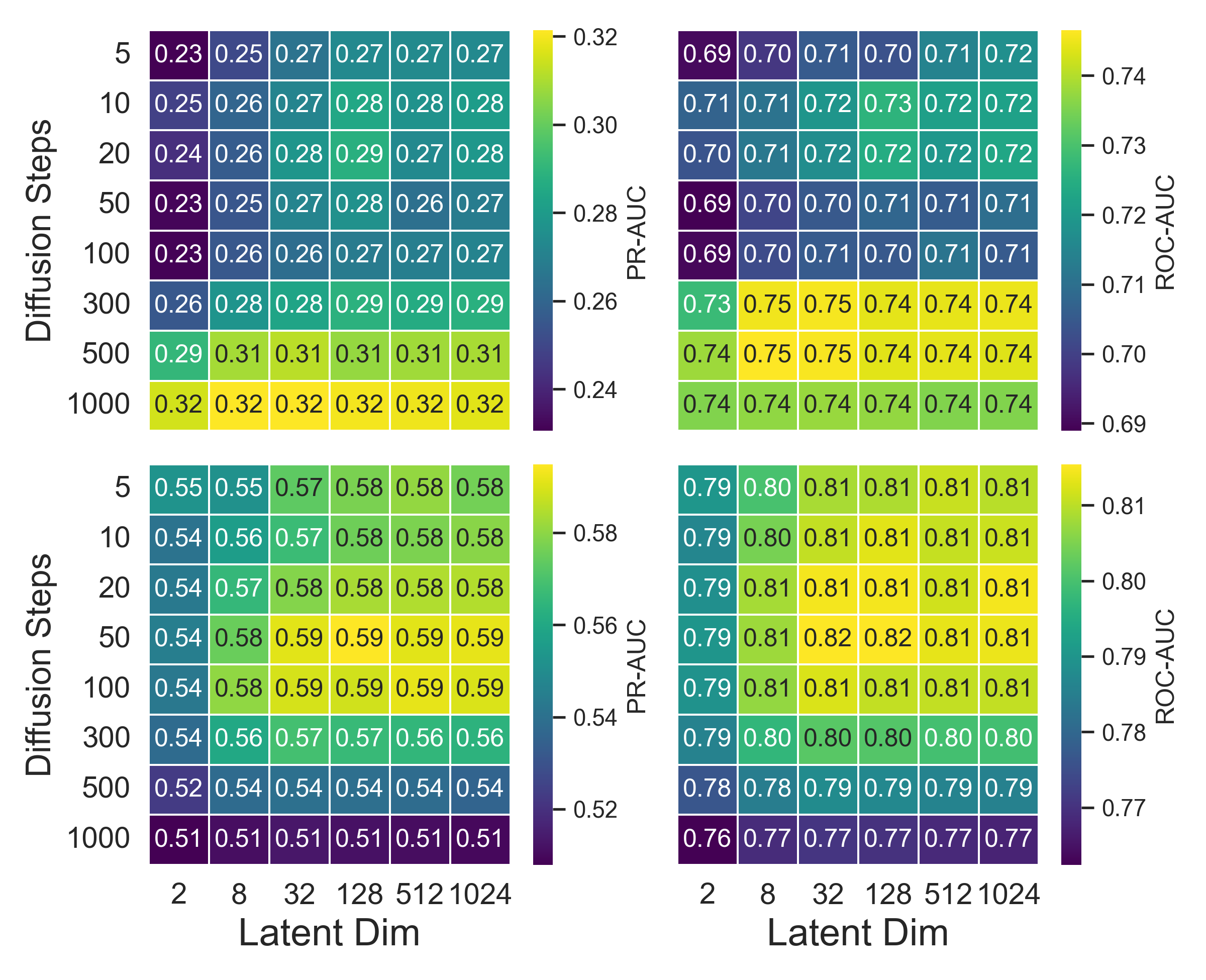}
    \caption{Heatmaps showing PR-AUC and ROC-AUC performance across varying diffusion steps and latent dimensionalities in the contrastive model for unsupervised (top row) and semi-supervised (bottom row) settings. Higher performance is observed with larger latent dimensions ($\mathbf{z\geq32}$).}
    \label{fig:grid_latent}
\end{figure}

\noindent \textbf{Latent Dimensionality.}
\autoref{fig:grid_latent} shows how diffusion steps and latent dimensionality affect DDAE-C performance, measured via PR-AUC and ROC-AUC. In the semi-supervised setting, performance improves with larger latent dimensions ($z\geq32$) and earlier diffusion steps, indicating that minor noise levels and richer latent spaces help capture complex feature structure and improve anomaly discrimination. Similar trends appear in the unsupervised setting, though at larger diffusion steps reaching highest scores.

\section{Experimental Setup Details}
\label{sec:experimental_setup_details}

\noindent \textbf{Model Architecture and Hyperparameters.} 
DDAE and DDAE-C models are implemented as feedforward encoder-decoder networks, LeakyReLu activation function, with the number of hidden layers, neurons per layer, and latent dimensions selected via grid search (\autoref{fig:grid_architecture_size}). All models are trained for up to 100 epochs using the Adam optimizer \citep{kingma2014adam} with $\beta_1{=}0.9$, $\beta_2{=}0.999$, and Glorot initialization \citep{glorot2010understanding}. To ensure scalability, batch size is selected dynamically from $\{2^3, \dots, 2^{13}\}$ to approximate $|X|/10$ for each dataset, where $|X|$ is the dataset size. Sinusoidal positional encodings are used for diffusion timestep embedding, with dimensions explored as shown in \autoref{fig:grid_diffusion_timestep_dim}.  \autoref{tab:grid_search} summarizes the full hyperparameter configuration.

\begin{table}[h]
    \centering
    \caption{Hyperparameter Grid Search Configuration}
    \label{tab:grid_search}    
    \begin{tabular}{l p{4.2cm}}
        \toprule
        \textbf{Hyperparameter} & \textbf{Search Space} \\
        \midrule
        Learning Rate ($\eta$) & $\{1e^{-3}, 1e^{-4}, 1e^{-5}\}$ \\
        Number of Layers & $\{1, 2, 3, 4, 5\}$ \\
        Hidden Units per Layer & $\{2, 8, 32, 128, 256, 512, 1024, 2048, 4096\}$ \\
        Latent Dimension ($z$) & $\{2, 8, 32, 128, 512, 1024\}$ \\
        Diffusion Scheduler & \text{linear, quadratic, cosine, sigmoid, exponential} \\
        Diffusion Steps ($T$) & $\{5, 10, 20, 50, 100, 300, 500, 1000, 1500, 2000\}$ \\
        Time Embedding Dim & $\{0, 2, 4, 8, 16, 32, 64, 128, 256, 512\}$ \\        
        \bottomrule
    \end{tabular}        
\end{table}

\subsection{Datasets}

We present the results derived from our methods and baseline comparisons across a variety of datasets provided by ADBench~\citep{han2022adbench}, as detailed in \autoref{tab:dataset_overview}. This encompasses 47 tabular datasets spanning various applications. Additionally, there are five datasets comprised of image representations extracted post the final average pooling layer of a ResNet-18~\citep{he2016deep} model, which is pre-trained on ImageNet~\citep{fei2009imagenet}. Furthermore, there are five datasets consisting of embeddings extracted from NLP tasks utilizing BERT~\citep{kenton2019bert}.

\begin{table*}[h!]
\centering
\caption{ADBench Dataset Overview: Summary Statistics and Categories}
\label{tab:dataset_overview}
\resizebox{0.7\textwidth}{!}{%
\begin{tabular}{lccccccc}
\toprule
\textbf{No.} & \textbf{Dataset} & \textbf{\# Samples} & \textbf{\# Features} & \textbf{\# Anomalies} & \textbf{\% Anomalies} & \textbf{Category} \\ 
\midrule
1	&	ALOI	&	49534	&	27	&	1508	&	3.04	&	Image	 \\	
2	&	annthyroid	&	7200	&	6	&	534	&	7.42	&	Healthcare	 \\	
3	&	backdoor	&	95329	&	196	&	2329	&	2.44	&	Network	 \\	
4	&	breastw	&	683	&	9	&	239	&	34.99	&	Healthcare	 \\	
5	&	campaign	&	41188	&	62	&	4640	&	11.27	&	Finance	 \\	
6	&	cardio	&	1831	&	21	&	176	&	9.61	&	Healthcare	 \\	
7	&	Cardiotocography	&	2114	&	21	&	466	&	22.04	&	Healthcare	 \\	
8	&	celeba	&	202599	&	39	&	4547	&	2.24	&	Image	 \\	
9	&	census	&	299285	&	500	&	18568	&	6.20	&	Sociology	 \\	
10	&	cover	&	286048	&	10	&	2747	&	0.96	&	Botany	 \\	
11	&	donors	&	619326	&	10	&	36710	&	5.93	&	Sociology	 \\	
12	&	fault	&	1941	&	27	&	673	&	34.67	&	Physical	 \\	
13	&	fraud	&	284807	&	29	&	492	&	0.17	&	Finance	 \\	
14	&	glass	&	214	&	7	&	9	&	4.21	&	Forensic	 \\	
15	&	Hepatitis	&	80	&	19	&	13	&	16.25	&	Healthcare	 \\	
16	&	http	&	567498	&	3	&	2211	&	0.39	&	Web	 \\	
17	&	InternetAds	&	1966	&	1555	&	368	&	18.72	&	Image	 \\	
18	&	Ionosphere	&	351	&	32	&	126	&	35.90	&	Oryctognosy	 \\	
19	&	landsat	&	6435	&	36	&	1333	&	20.71	&	Astronautics	 \\	
20	&	letter	&	1600	&	32	&	100	&	6.25	&	Image	 \\	
21	&	Lymphography	&	148	&	18	&	6	&	4.05	&	Healthcare	 \\	
22	&	magic.gamma	&	19020	&	10	&	6688	&	35.16	&	Physical	 \\	
23	&	mammography	&	11183	&	6	&	260	&	2.32	&	Healthcare	 \\	
24	&	mnist	&	7603	&	100	&	700	&	9.21	&	Image	 \\	
25	&	musk	&	3062	&	166	&	97	&	3.17	&	Chemistry	 \\	
26	&	optdigits	&	5216	&	64	&	150	&	2.88	&	Image	 \\	
27	&	PageBlocks	&	5393	&	10	&	510	&	9.46	&	Document	 \\	
28	&	pendigits	&	6870	&	16	&	156	&	2.27	&	Image	 \\	
29	&	Pima	&	768	&	8	&	268	&	34.90	&	Healthcare	 \\	
30	&	satellite	&	6435	&	36	&	2036	&	31.64	&	Astronautics	 \\	
31	&	satimage-2	&	5803	&	36	&	71	&	1.22	&	Astronautics	 \\	
32	&	shuttle	&	49097	&	9	&	3511	&	7.15	&	Astronautics	 \\	
33	&	skin	&	245057	&	3	&	50859	&	20.75	&	Image	 \\	
34	&	smtp	&	95156	&	3	&	30	&	0.03	&	Web	 \\	
35	&	SpamBase	&	4207	&	57	&	1679	&	39.91	&	Document	 \\	
36	&	speech	&	3686	&	400	&	61	&	1.65	&	Linguistics	 \\	
37	&	Stamps	&	340	&	9	&	31	&	9.12	&	Document	 \\	
38	&	thyroid	&	3772	&	6	&	93	&	2.47	&	Healthcare	 \\	
39	&	vertebral	&	240	&	6	&	30	&	12.50	&	Biology	 \\	
40	&	vowels	&	1456	&	12	&	50	&	3.43	&	Linguistics	 \\	
41	&	Waveform	&	3443	&	21	&	100	&	2.90	&	Physics	 \\	
42	&	WBC	&	223	&	9	&	10	&	4.48	&	Healthcare	 \\	
43	&	WDBC	&	367	&	30	&	10	&	2.72	&	Healthcare	 \\	
44	&	Wilt	&	4819	&	5	&	257	&	5.33	&	Botany	 \\	
45	&	wine	&	129	&	13	&	10	&	7.75	&	Chemistry	 \\	
46	&	WPBC	&	198	&	33	&	47	&	23.74	&	Healthcare	 \\	
47	&	yeast	&	1484	&	8	&	507	&	34.16	&	Biology	 \\	
48	&	CIFAR10	&	5263	&	512	&	263	&	5.00	&	Image	 \\	
49	&	FashionMNIST	&	6315	&	512	&	315	&	5.00	&	Image	 \\	
50	&	MNIST-C	&	10000	&	512	&	500	&	5.00	&	Image	 \\	
51	&	MVTec-AD&	5354    &   512  &  1258    &   23.50	&	Image	 \\	
52	&	SVHN	&	5208	&	512	&	260	&	5.00	&	Image	 \\	
53	&	Agnews	&	10000	&	768	&	500	&	5.00	&	NLP	 \\	
54	&	Amazon	&	10000	&	768	&	500	&	5.00	&	NLP	 \\	
55	&	Imdb	&	10000	&	768	&	500	&	5.00	&	NLP	 \\	
56	&	Yelp	&	10000	&	768	&	500	&	5.00	&	NLP	 \\	
57	&	20newsgroups	&	11905   &	768	& 591 &   4.96    &	NLP \\
\bottomrule
\end{tabular}
}
\end{table*}

\section{Latent Representations}
\label{sec:latent_space}

\begin{figure*}[h]
  \centering

  \begin{subfigure}[b]{0.24\linewidth}
    \centering
    \includegraphics[width=\linewidth]{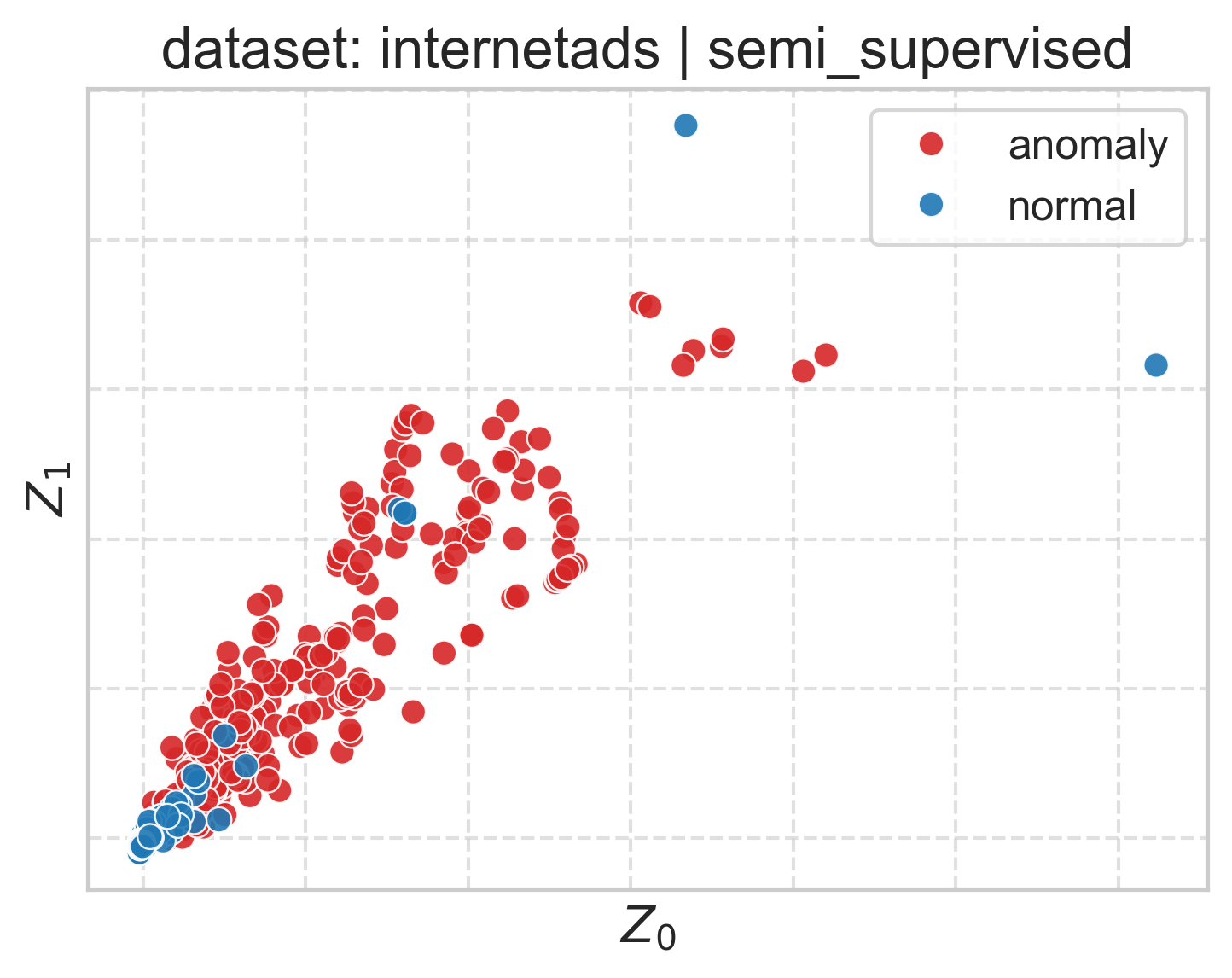}
  \end{subfigure}
   \hfill
  \begin{subfigure}[b]{0.24\linewidth}
    \centering
    \includegraphics[width=\linewidth]{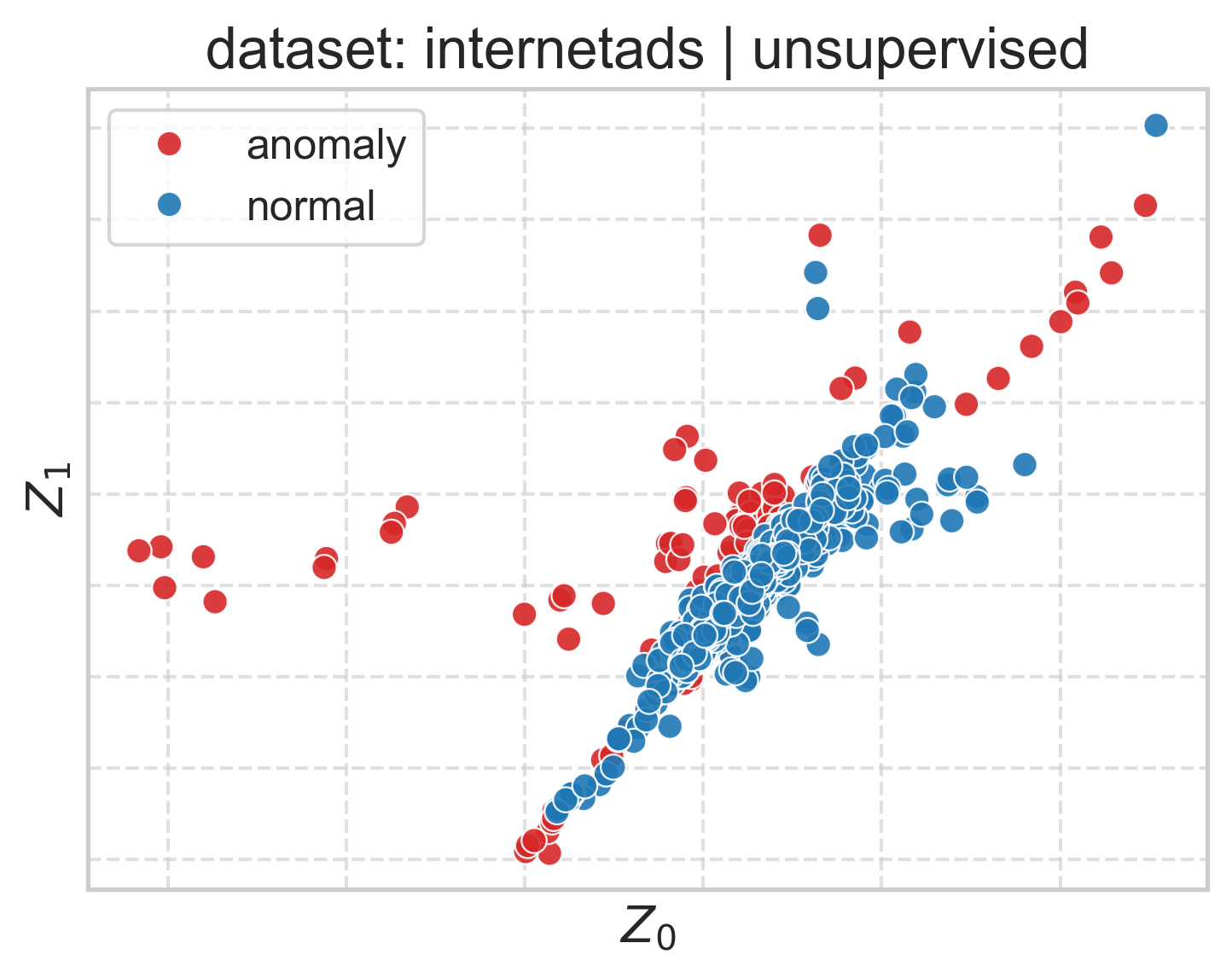}
  \end{subfigure}
   \hfill  
  \begin{subfigure}[b]{0.24\linewidth}
    \centering
    \includegraphics[width=\linewidth]{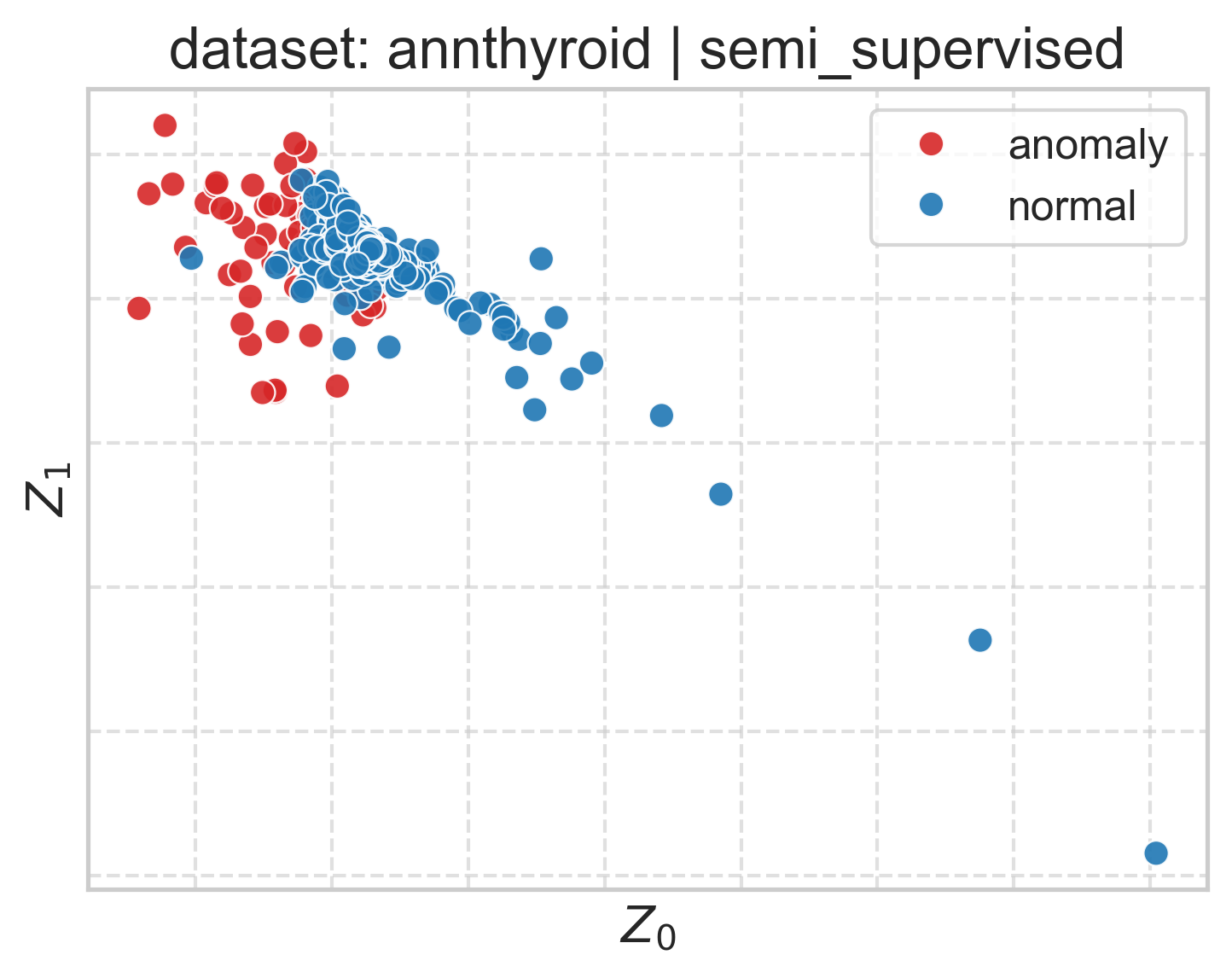}
  \end{subfigure}
   \hfill
  \begin{subfigure}[b]{0.24\linewidth}
    \centering
    \includegraphics[width=\linewidth]{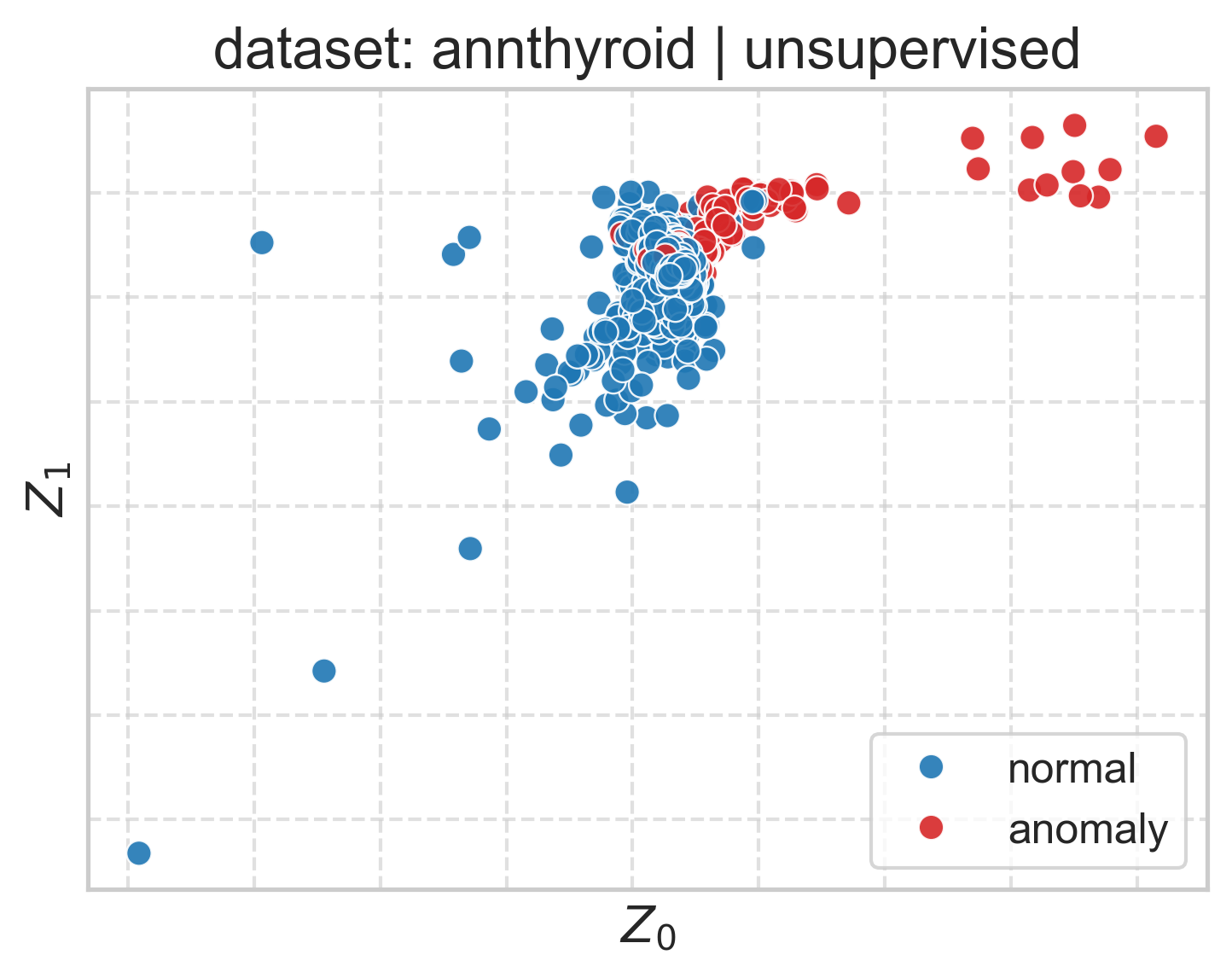}
  \end{subfigure}

    \begin{subfigure}[b]{0.24\linewidth}
    \centering
    \includegraphics[width=\linewidth]{figures/latent_semi/latent_emb_musk_epoch_100.png}
  \end{subfigure}
   \hfill
   \begin{subfigure}[b]{0.24\linewidth}
    \centering
    \includegraphics[width=\linewidth]{figures/latent_unsup/latent_emb_musk_epoch_100.png}
  \end{subfigure}
   \hfill  
  \begin{subfigure}[b]{0.24\linewidth}
    \centering
    \includegraphics[width=\linewidth]{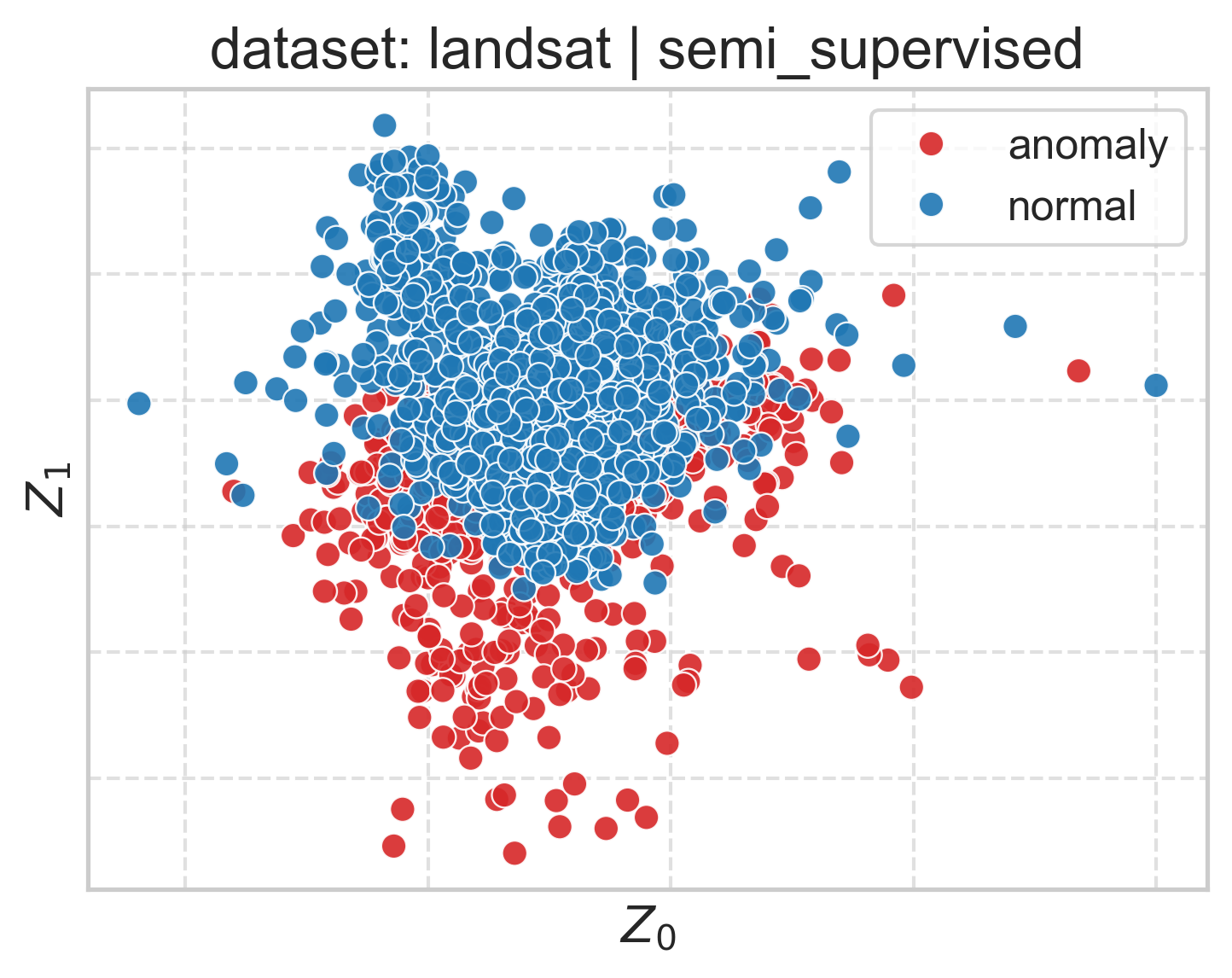}
  \end{subfigure}
   \hfill
   \begin{subfigure}[b]{0.24\linewidth}
    \centering
    \includegraphics[width=\linewidth]{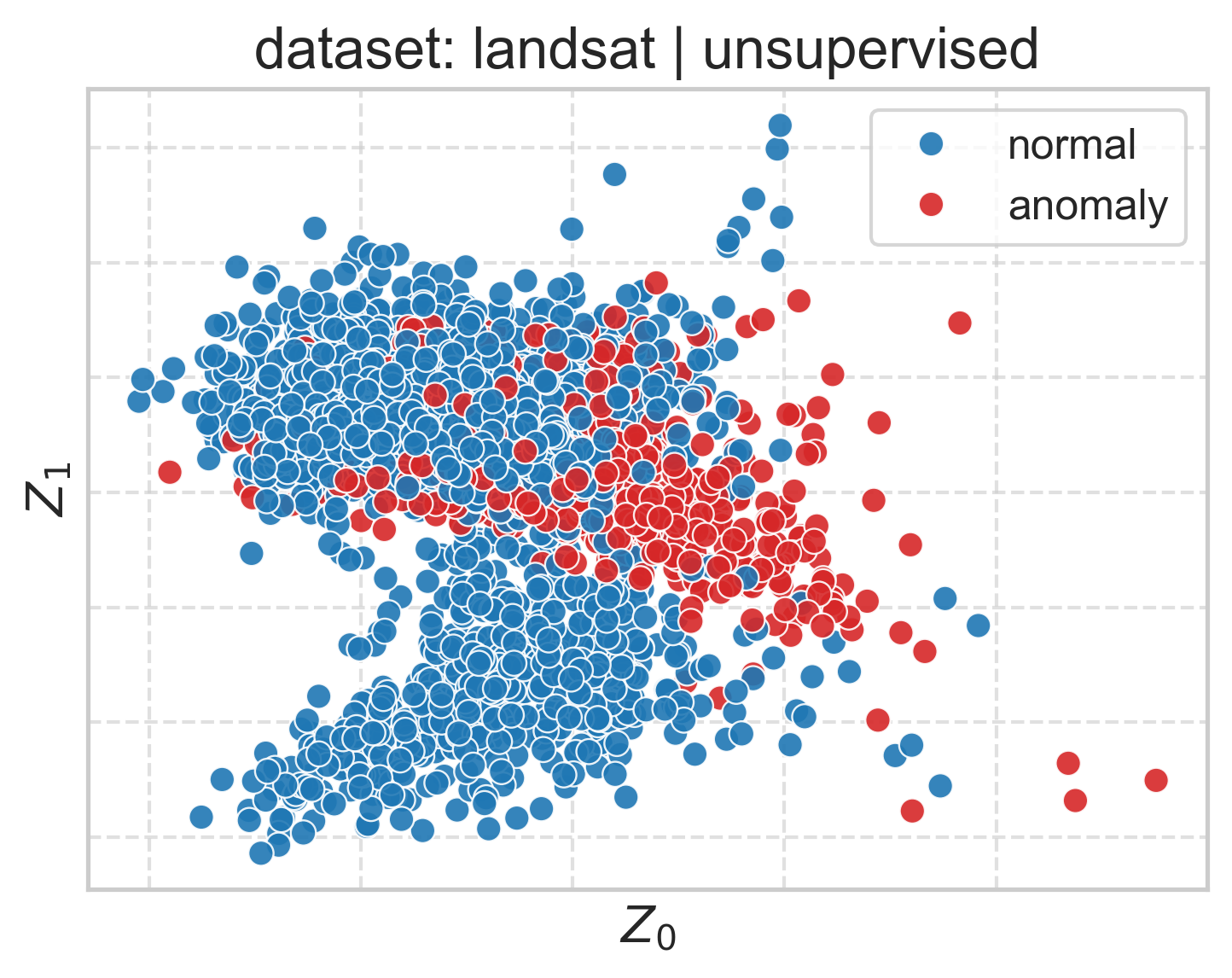}
  \end{subfigure}

    \begin{subfigure}[b]{0.24\linewidth}
    \centering
    \includegraphics[width=\linewidth]{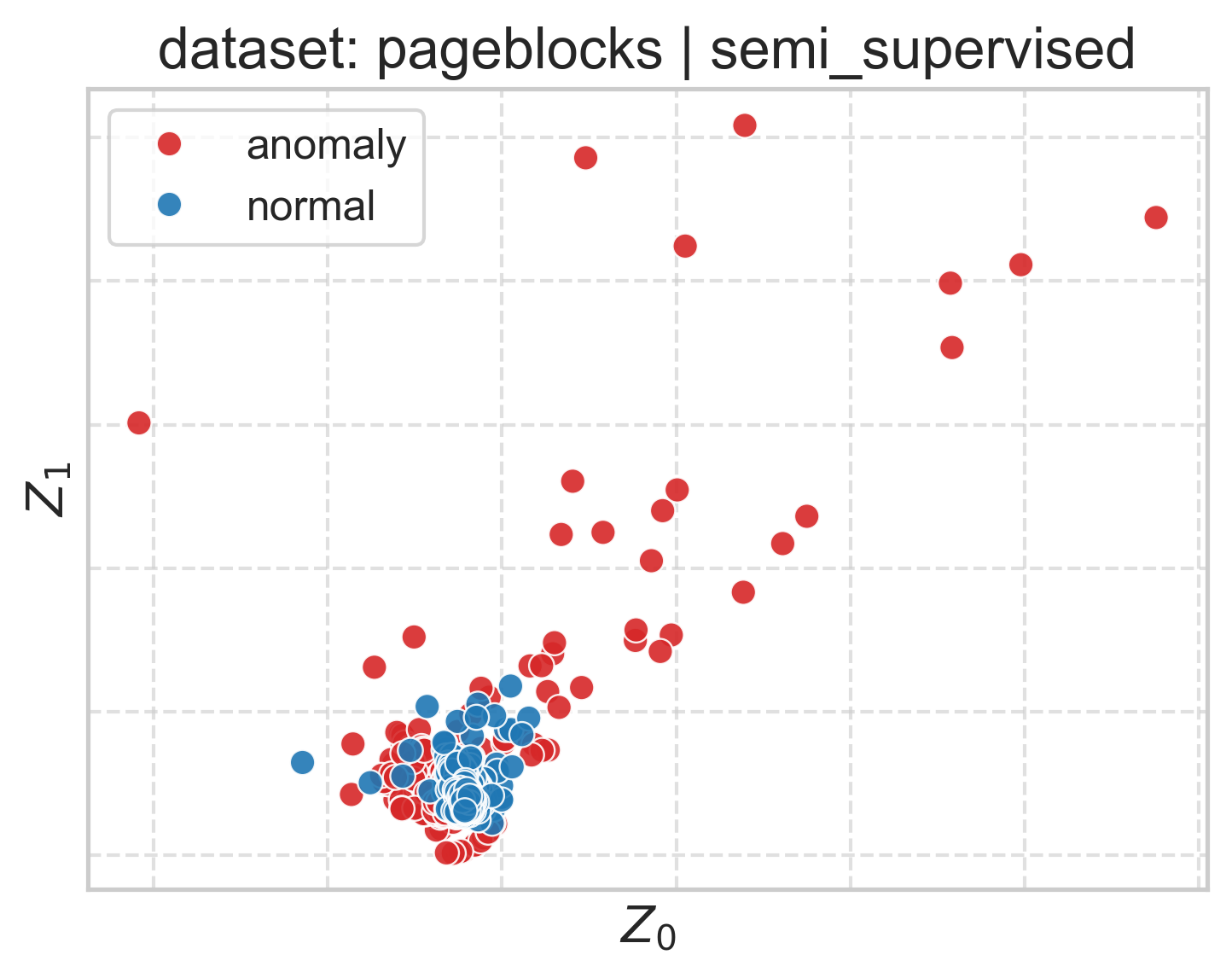}
  \end{subfigure}  
  \hfill
  \begin{subfigure}[b]{0.24\linewidth}
    \centering
    \includegraphics[width=\linewidth]{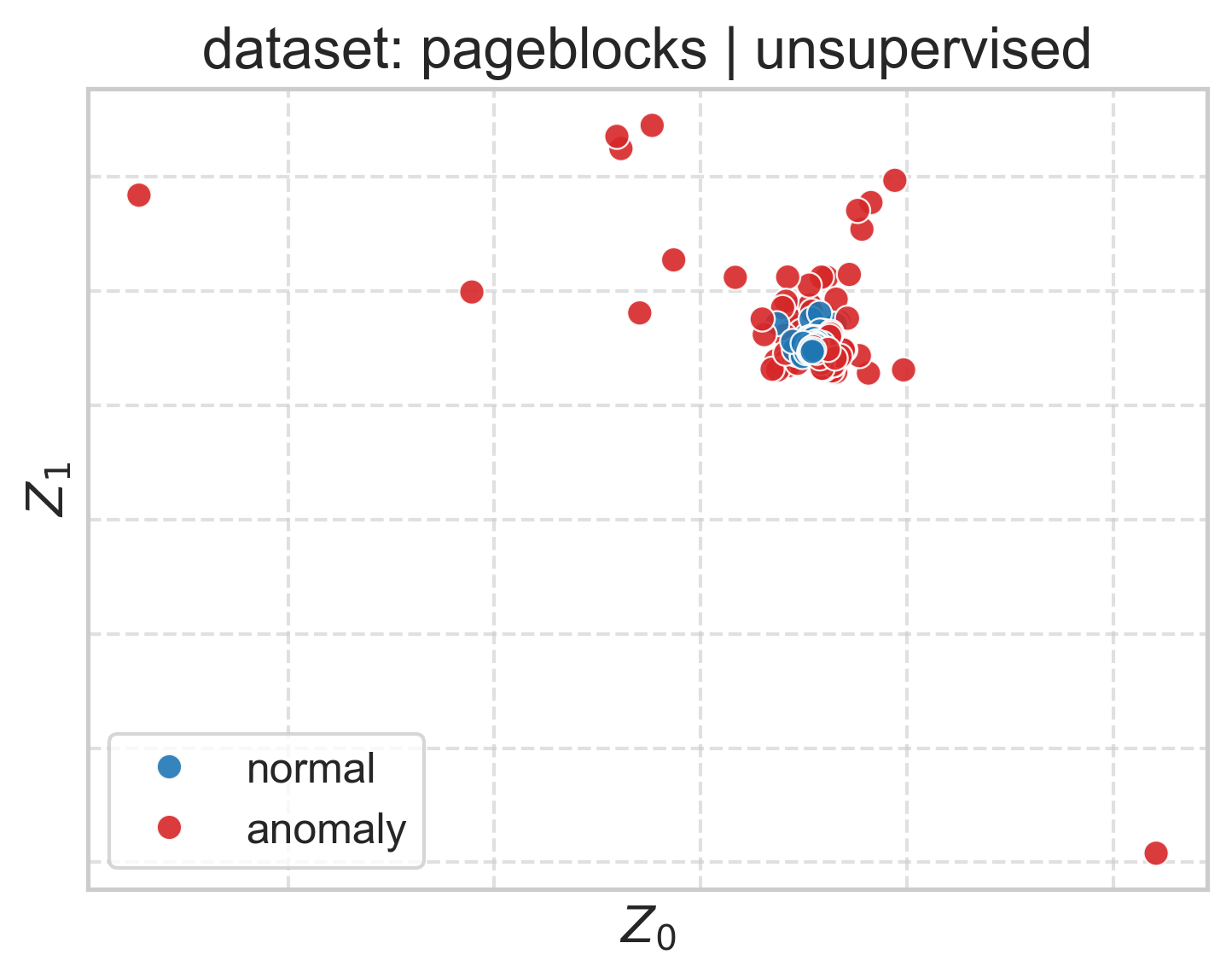}
  \end{subfigure}  
   \hfill  
  \begin{subfigure}[b]{0.24\linewidth}
    \centering
    \includegraphics[width=\linewidth]{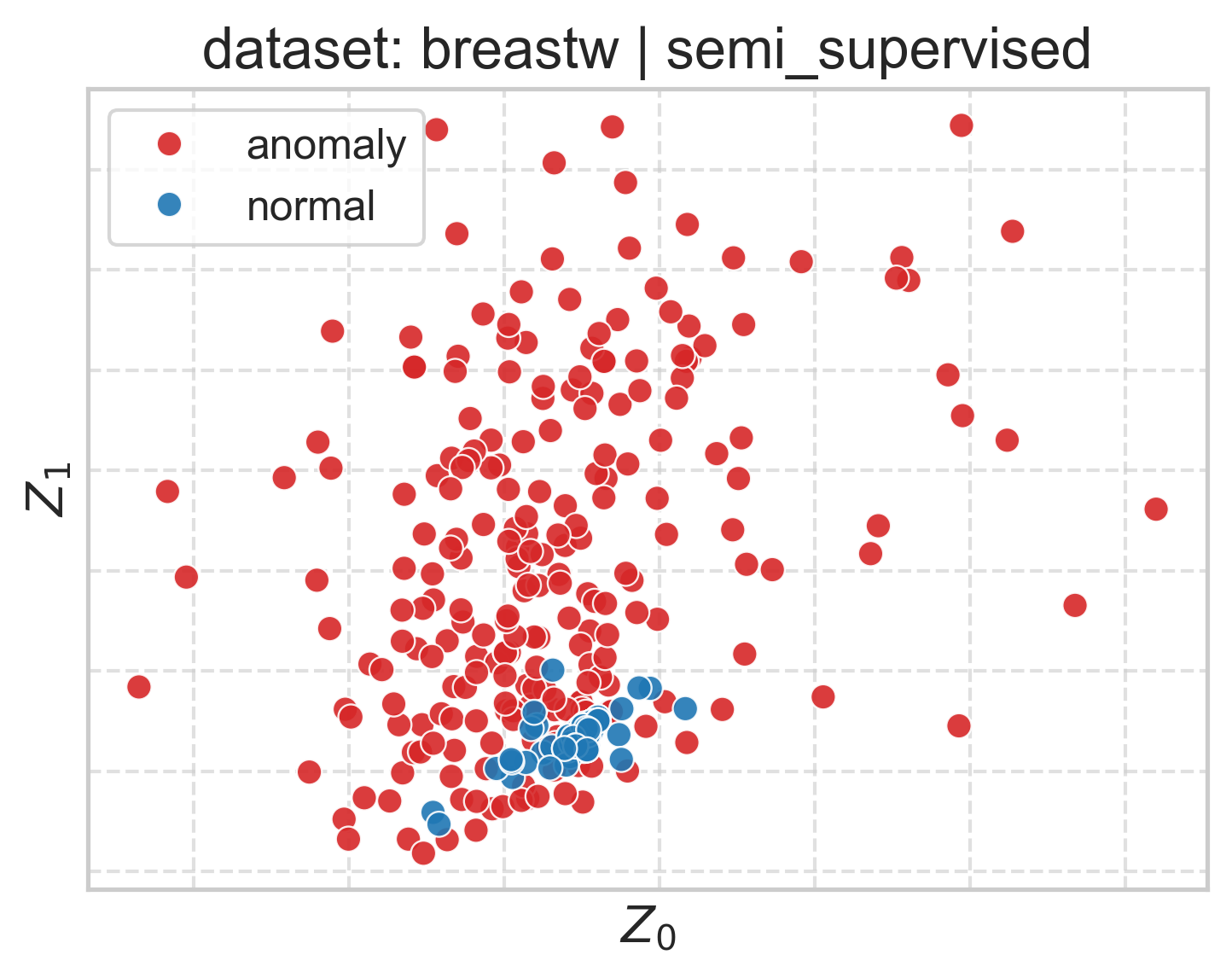}
  \end{subfigure}  
  \hfill
  \begin{subfigure}[b]{0.24\linewidth}
    \centering
    \includegraphics[width=\linewidth]{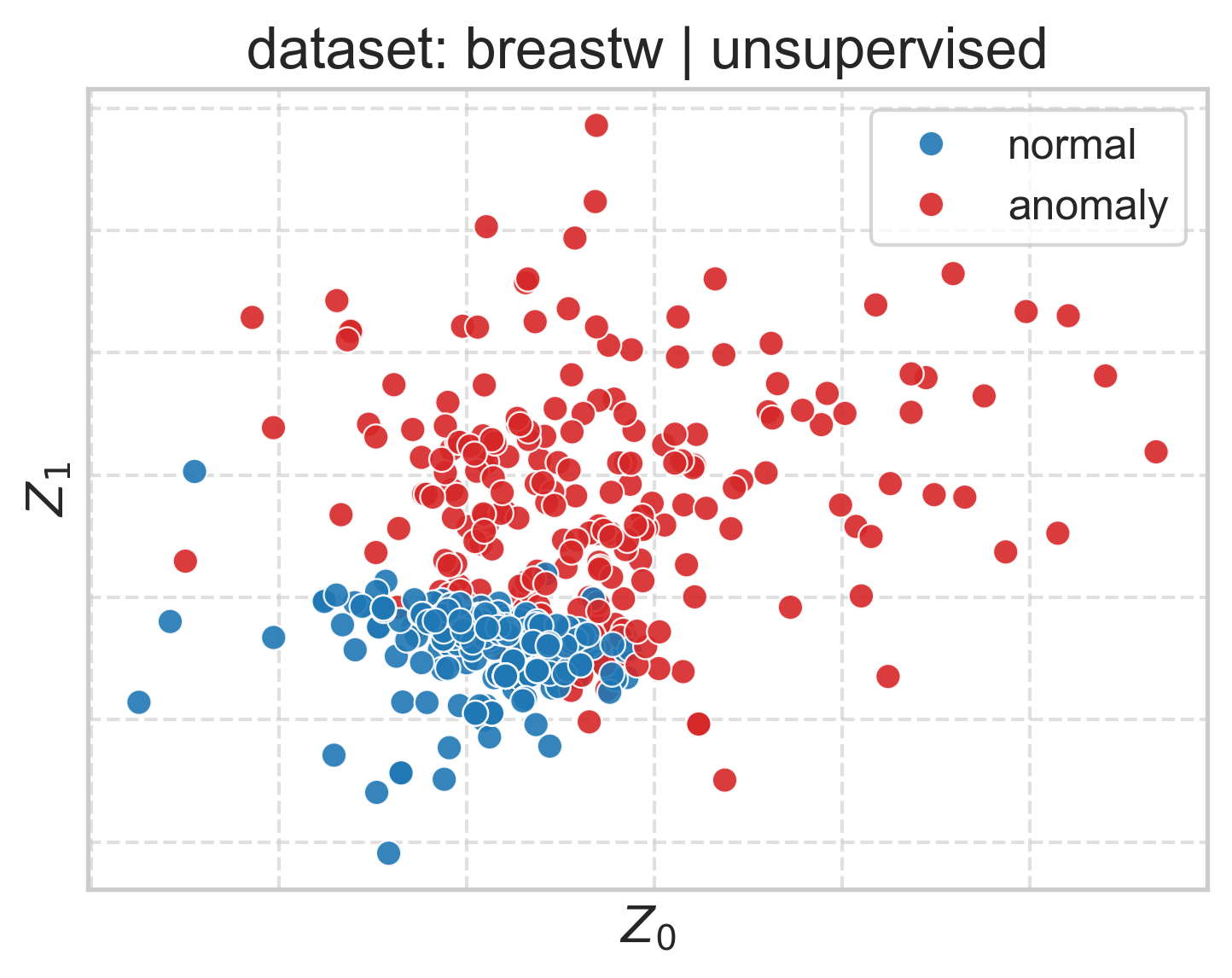}
  \end{subfigure}  

    \begin{subfigure}[b]{0.24\linewidth}
    \centering
    \includegraphics[width=\linewidth]{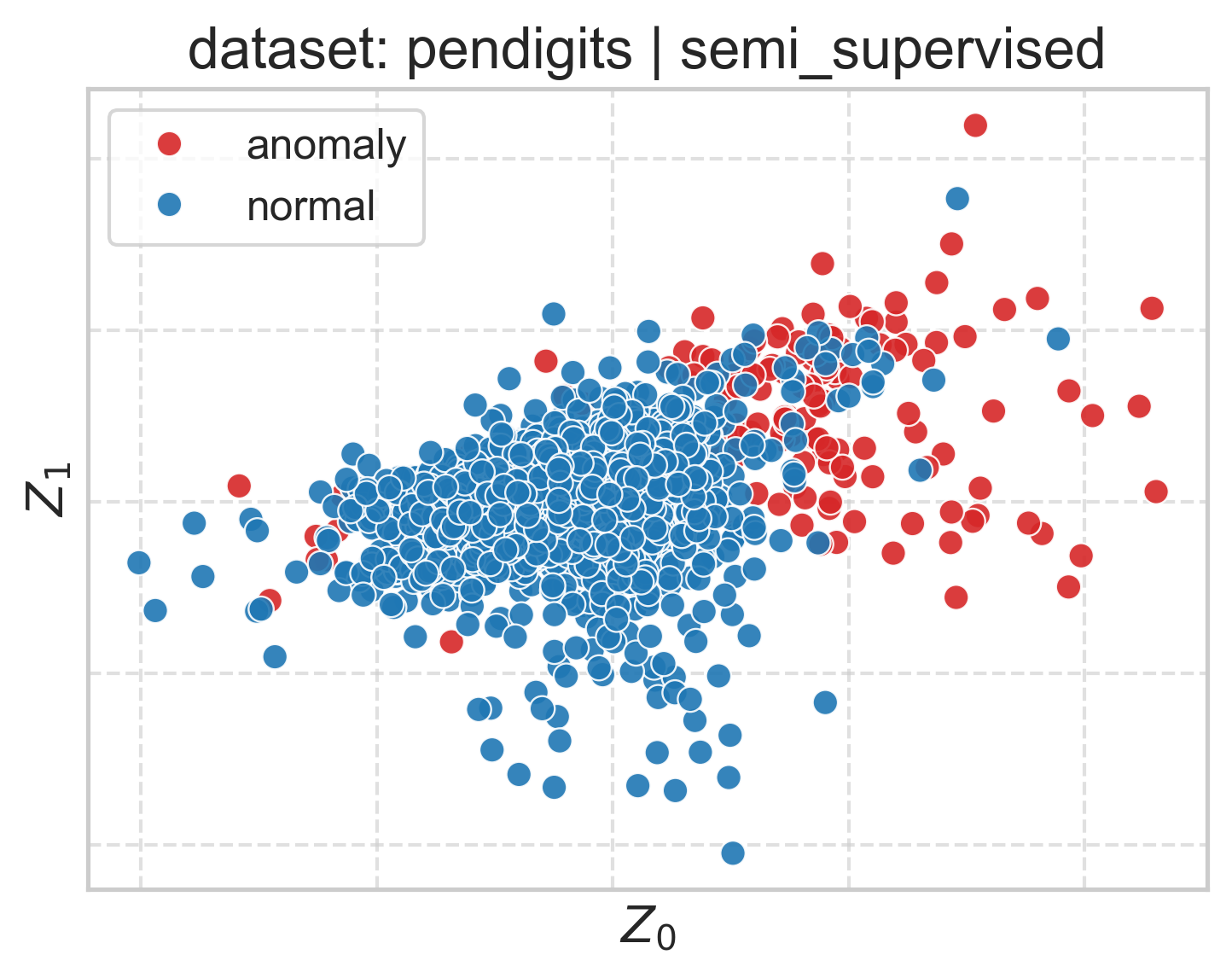}
  \end{subfigure}
   \hfill
  \begin{subfigure}[b]{0.24\linewidth}
    \centering
    \includegraphics[width=\linewidth]{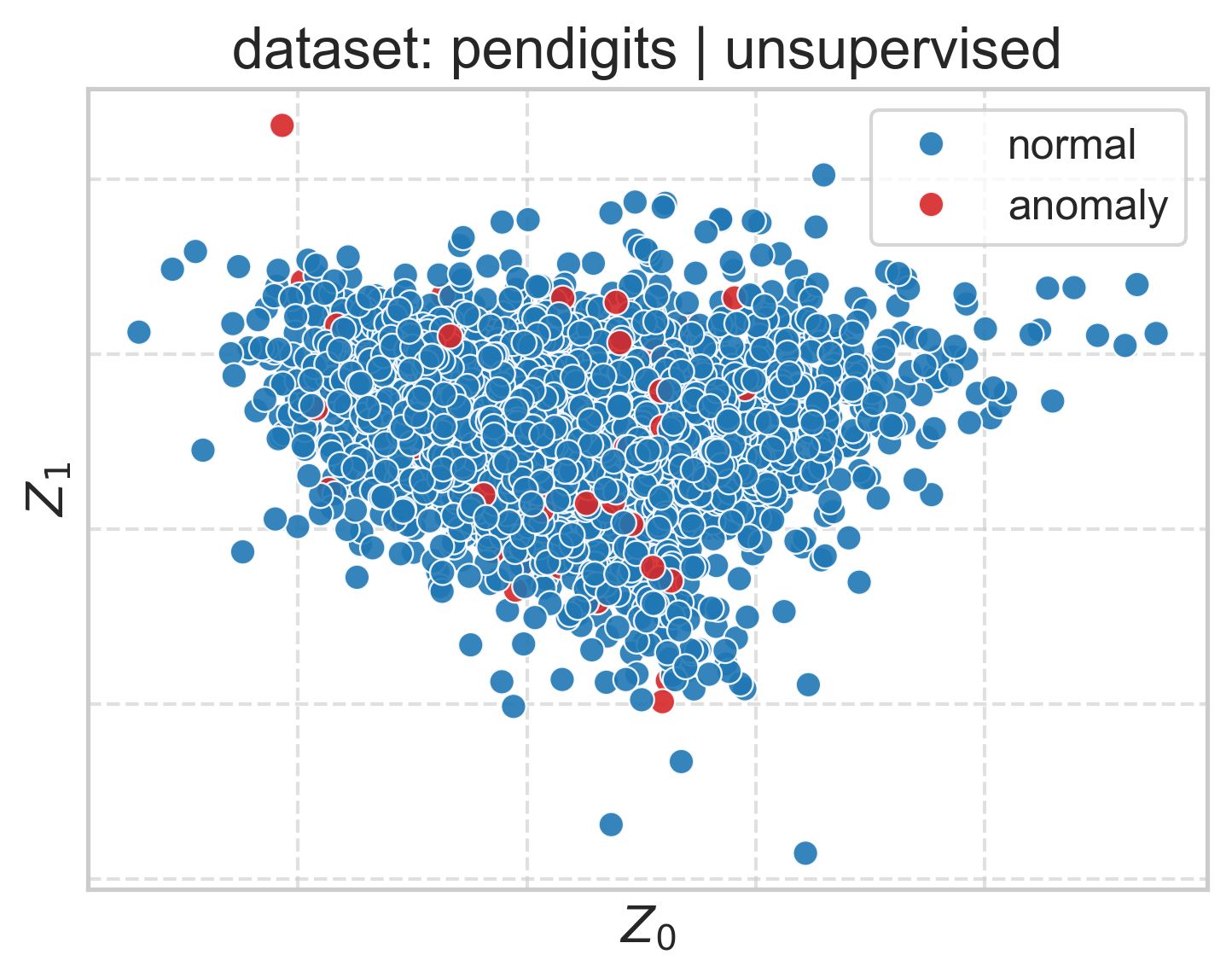}
  \end{subfigure}
   \hfill  
  \begin{subfigure}[b]{0.24\linewidth}
    \centering
    \includegraphics[width=\linewidth]{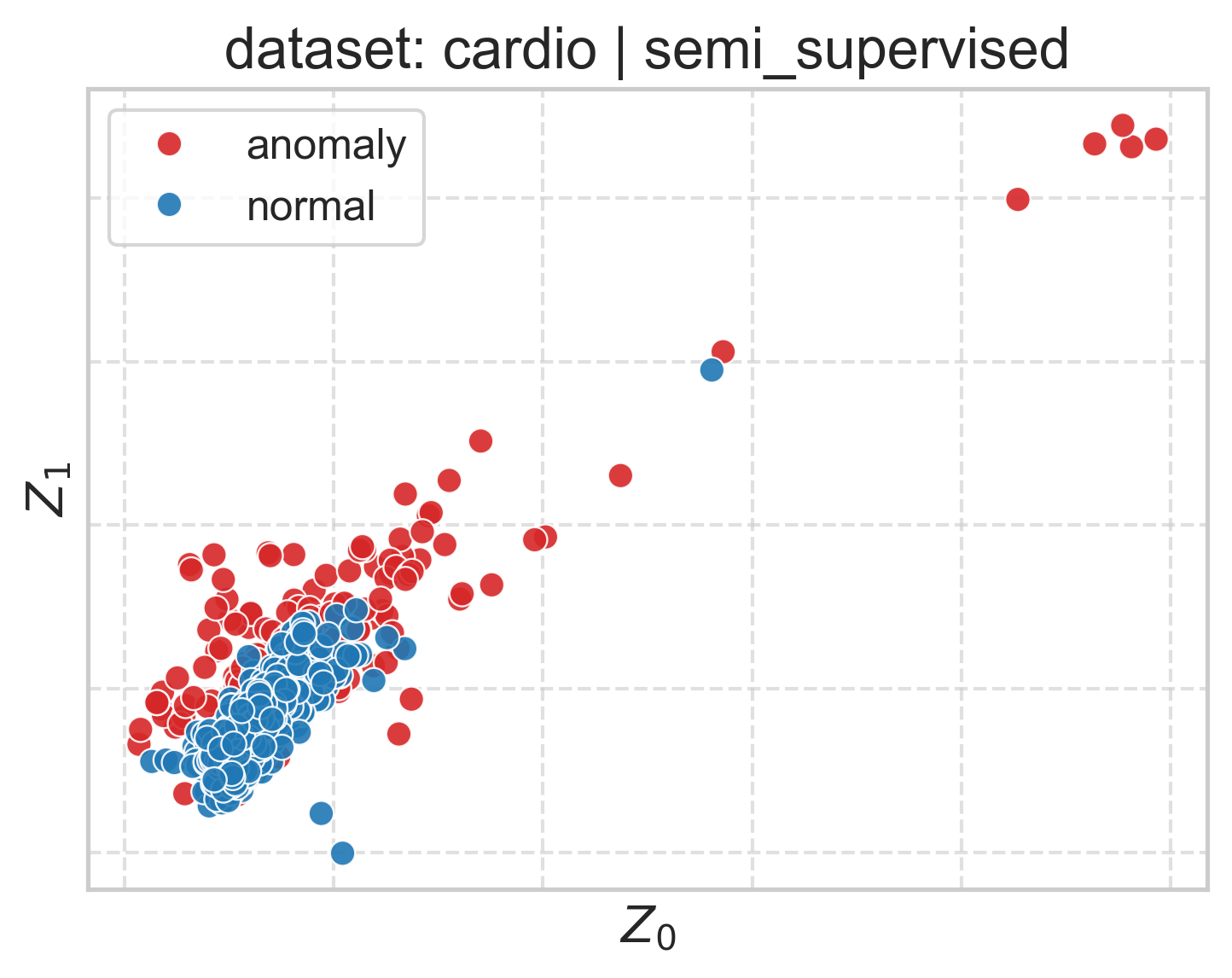}
  \end{subfigure}
   \hfill
  \begin{subfigure}[b]{0.24\linewidth}
    \centering
    \includegraphics[width=\linewidth]{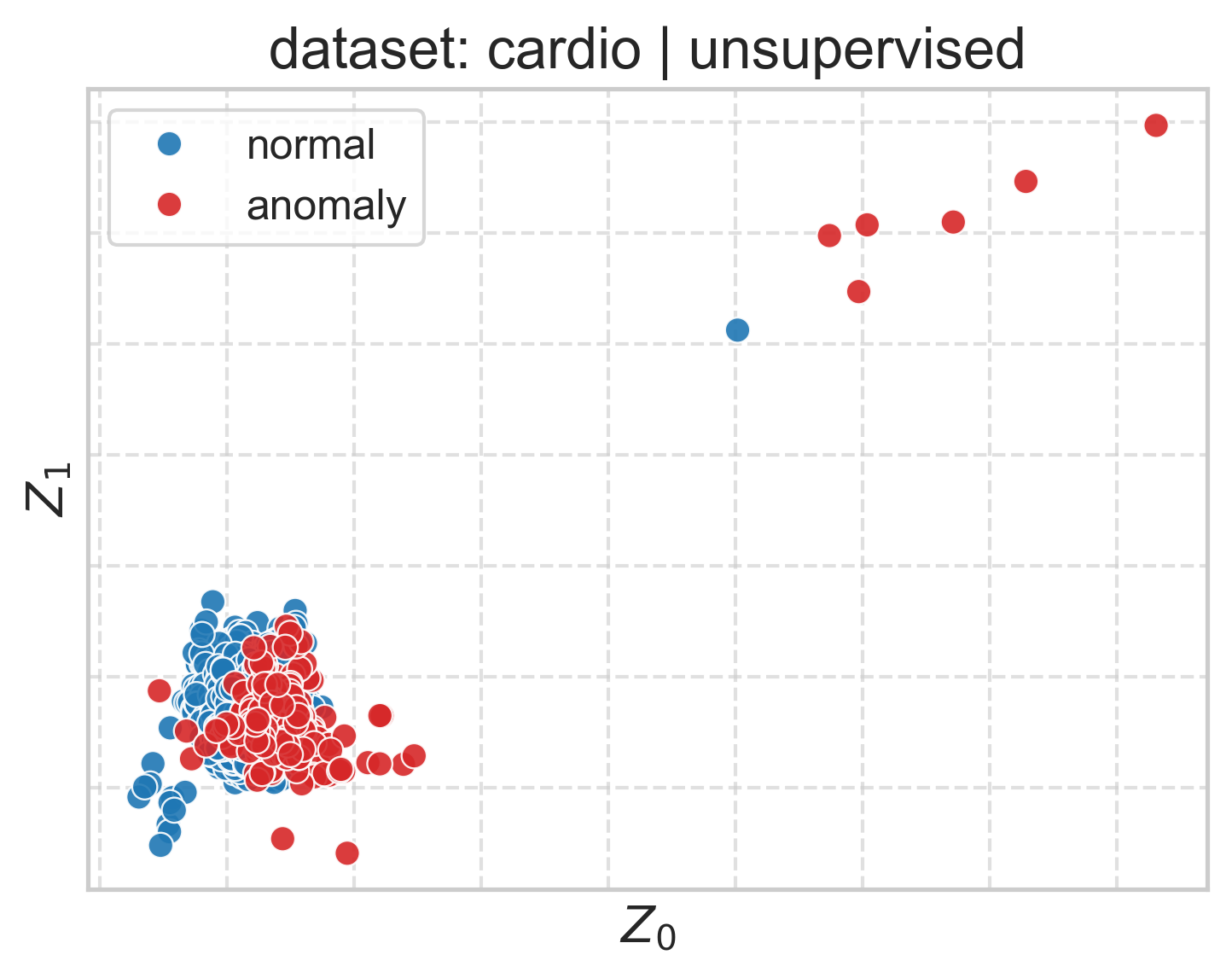}
  \end{subfigure}

    \begin{subfigure}[b]{0.24\linewidth}
    \centering
    \includegraphics[width=\linewidth]{figures/latent_semi/latent_emb_satimage-2_epoch_100.png}
  \end{subfigure}
   \hfill
  \begin{subfigure}[b]{0.24\linewidth}
    \centering
    \includegraphics[width=\linewidth]{figures/latent_unsup/latent_emb_satimage-2_epoch_100.png}
  \end{subfigure}
   \hfill  
  \begin{subfigure}[b]{0.24\linewidth}
    \centering
    \includegraphics[width=\linewidth]{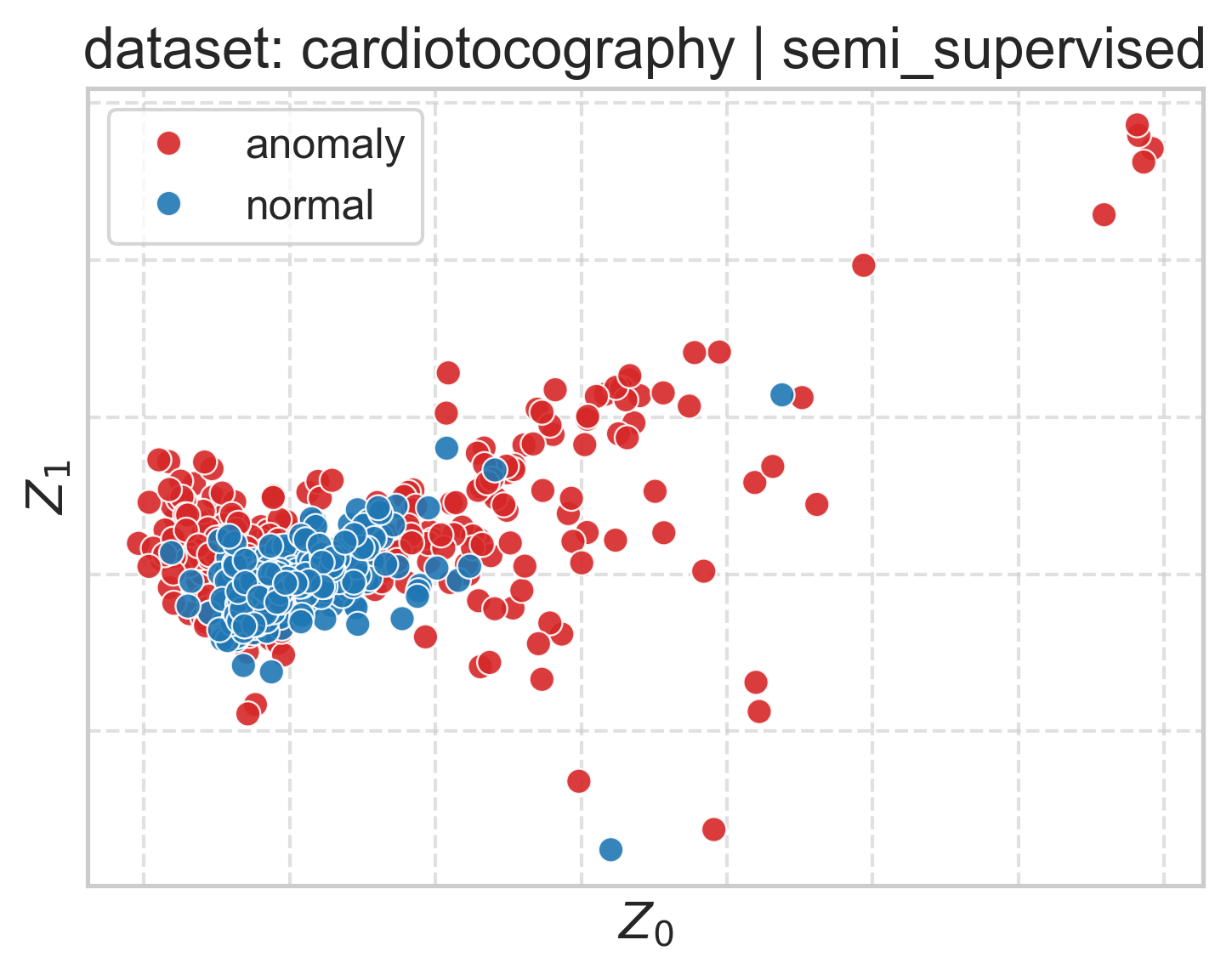}
  \end{subfigure}
   \hfill
  \begin{subfigure}[b]{0.24\linewidth}
    \centering
    \includegraphics[width=\linewidth]{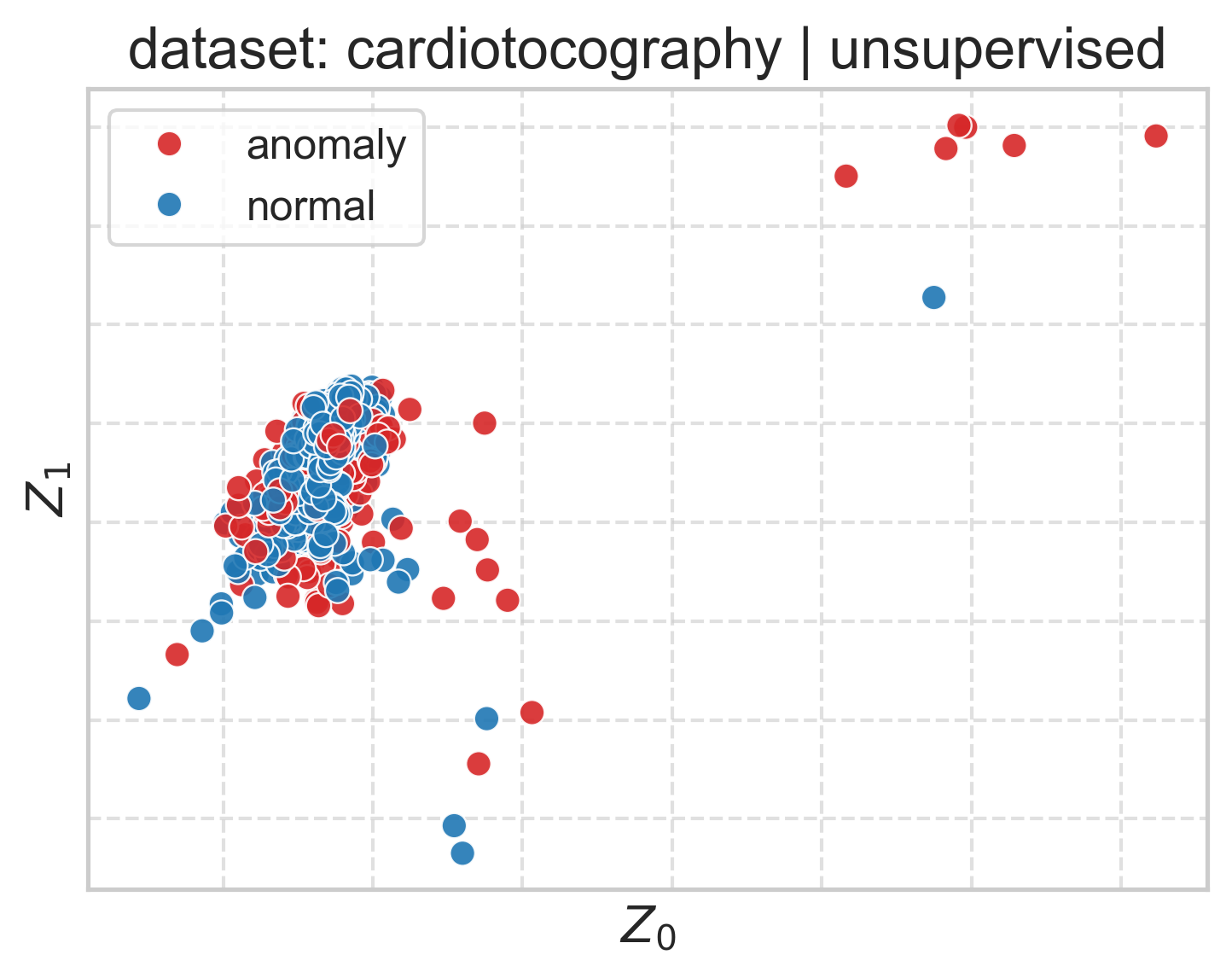}
  \end{subfigure}

    \begin{subfigure}[b]{0.24\linewidth}
    \centering
    \includegraphics[width=\linewidth]{figures/latent_semi/latent_emb_thyroid_epoch_100.png}
  \end{subfigure}  
   \hfill
    \begin{subfigure}[b]{0.24\linewidth}
    \centering
    \includegraphics[width=\linewidth]{figures/latent_unsup/latent_emb_thyroid_epoch_100.png}
  \end{subfigure}  
   \hfill  
  \begin{subfigure}[b]{0.24\linewidth}
    \centering
    \includegraphics[width=\linewidth]{figures/latent_semi/latent_emb_mammography_epoch_100.png}
  \end{subfigure}  
   \hfill
    \begin{subfigure}[b]{0.24\linewidth}
    \centering
    \includegraphics[width=\linewidth]{figures/latent_unsup/latent_emb_mammography_epoch_100.png}
  \end{subfigure}  

  \vspace{-0.3cm}
  \caption{The latent 2D representation of ADBench datasets learned by DDAE-C model in a semi-supervised (left) and unsupervised (right) settings.}
  \label{fig:latent_repr_all}
\end{figure*}

\autoref{fig:latent_repr_all} presents the 2D latent representations learned by the DDAE-C model in a semi-supervised (left) and unsupervised (right) setting across multiple datasets from ADBench. The results demonstrate varying degrees of anomaly separation, influenced by the underlying structure of each dataset. In some datasets (e.g., landsat, pendigits, and satimage-2), normal samples (blue) form dense clusters, while anomalies (red) are well-separated. In contrast, datasets like cardiotocography and mammography exhibit partial overlap, suggesting that anomalies share structural similarities with normal samples. Certain datasets, such as musk and breastw, show a more scattered representation, indicating potential challenges in anomaly discrimination. Compared to unsupervised learning, the semi-supervised representations exhibit better anomaly separability, with clearer decision boundaries and more structured clusters. This suggests that the availability of labeled anomalies helps the model refine the feature space, improving its ability to distinguish normal and anomalous samples more effectively.

\section{Evaluation of the Diffusion Noise Scheduler}
\label{sec:evaluation_diffusion_noise_scheduler}

\begin{figure*}
    \centering
    \includegraphics[width=0.75\linewidth]{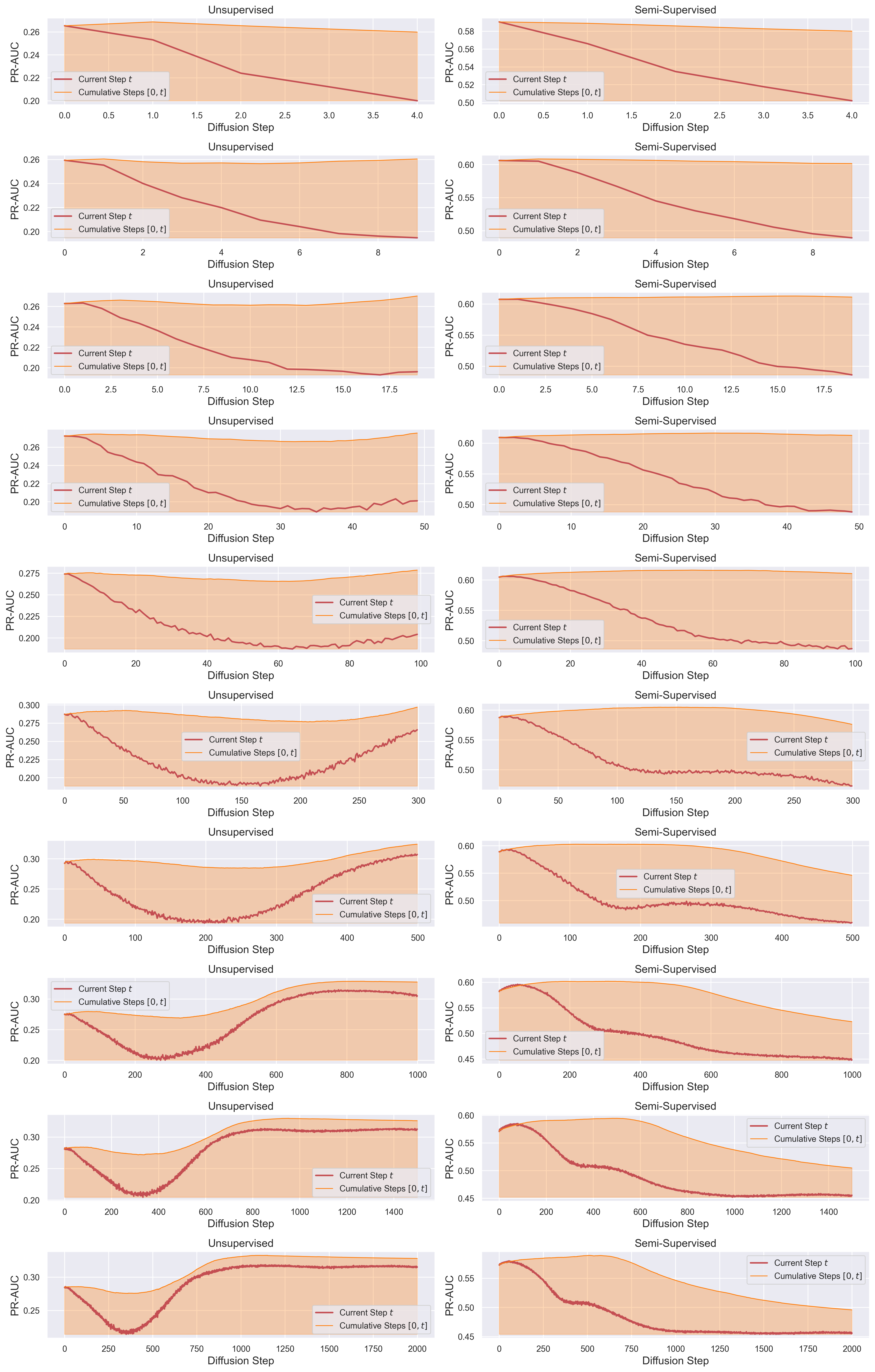}
    \caption{Anomaly detection performance evaluated on models trained with different diffusion noise levels $T\in [5, 10, 20, 50, 100, 300, 500, 1000, 1500, 2000]$}
    \label{fig:noise_analysis_all_steps}
\end{figure*}

\autoref{fig:noise_analysis_all_steps} presents the anomaly detection performance of models trained with different diffusion noise levels T$\in$ [5, 10, 20, 50, 100, 300, 500, 1000, 1500, 2000] across unsupervised and semi-supervised settings. The PR-AUC scores are tracked at each diffusion step, with per-step performance (red) and cumulative performance (orange) curves.

In the semi-supervised setting, earlier diffusion steps contribute the most to anomaly detection, and performance peaks at lower noise levels $T\approx50-100$ before declining. This suggests that smaller noise magnitudes help build a clearer decision boundary, improving generalization while preventing the model from overfitting to labeled anomalies. However, for models trained with $T>300$ (top plots), this trend is less evident, indicating that the noise level must be sufficiently large for structured learning effects to emerge.

In contrast, in the unsupervised setting, later diffusion steps play a more dominant role. These steps contain almost no signal (signal-to-noise ratio is near zero) since the input is almost completely corrupted. However, this extreme noise acts as a regularizer, preventing the model from memorizing anomalies and overfitting to spurious patterns. The bottom plots illustrate that in the unsupervised case, once $T>500$ is reached, performance stabilizes, meaning that increasing $T$ further does not yield improvements. This is expected, as beyond 500 steps, the input is effectively just Gaussian noise, providing no additional useful information to the model.

\begin{figure*}
    \centering
    \includegraphics[width=0.75\linewidth]{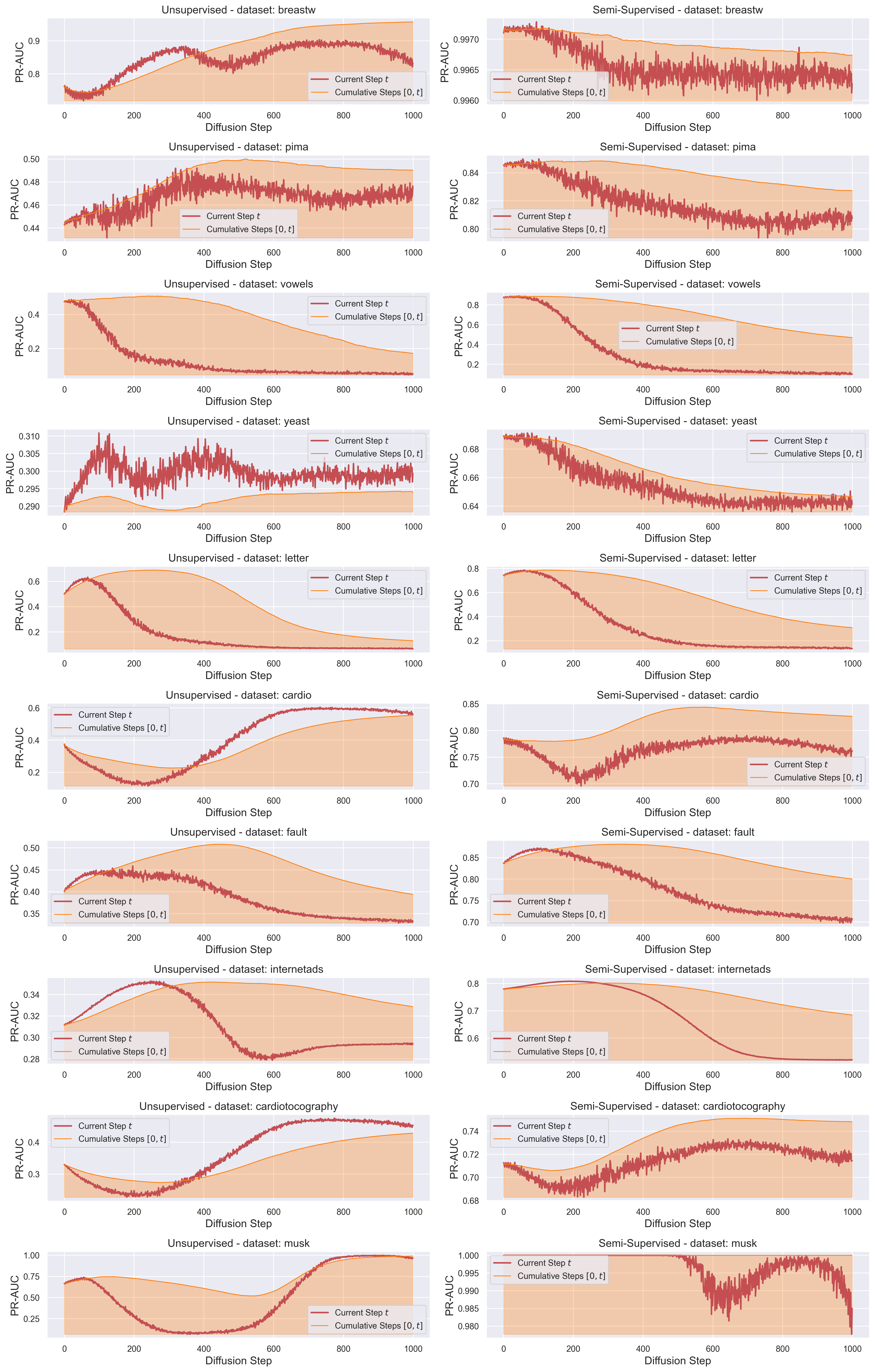}
    \caption{Anomaly detection performance of individual dataset evaluated at each diffusion noisy step out of total training $T=1000$ steps.}
    \label{fig:noise_analysis_all_steps_per_dataset}
\end{figure*}

Additionally, Figure~\ref{fig:noise_analysis_all_steps_per_dataset} provides the same metrics but on a per-dataset level, revealing that while the general trend holds, individual datasets exhibit variability. The optimal diffusion step range is highly dataset-dependent, influenced by the structure of normal and anomalous samples. These results emphasize the importance of tuning the noise schedule based on dataset characteristics, rather than relying on a single optimal setting for all cases.

\section{Experimental Results: Detailed}
\label{sec:experimental_results_detailed}

\begin{figure*}
    \centering
    \includegraphics[width=1\linewidth]{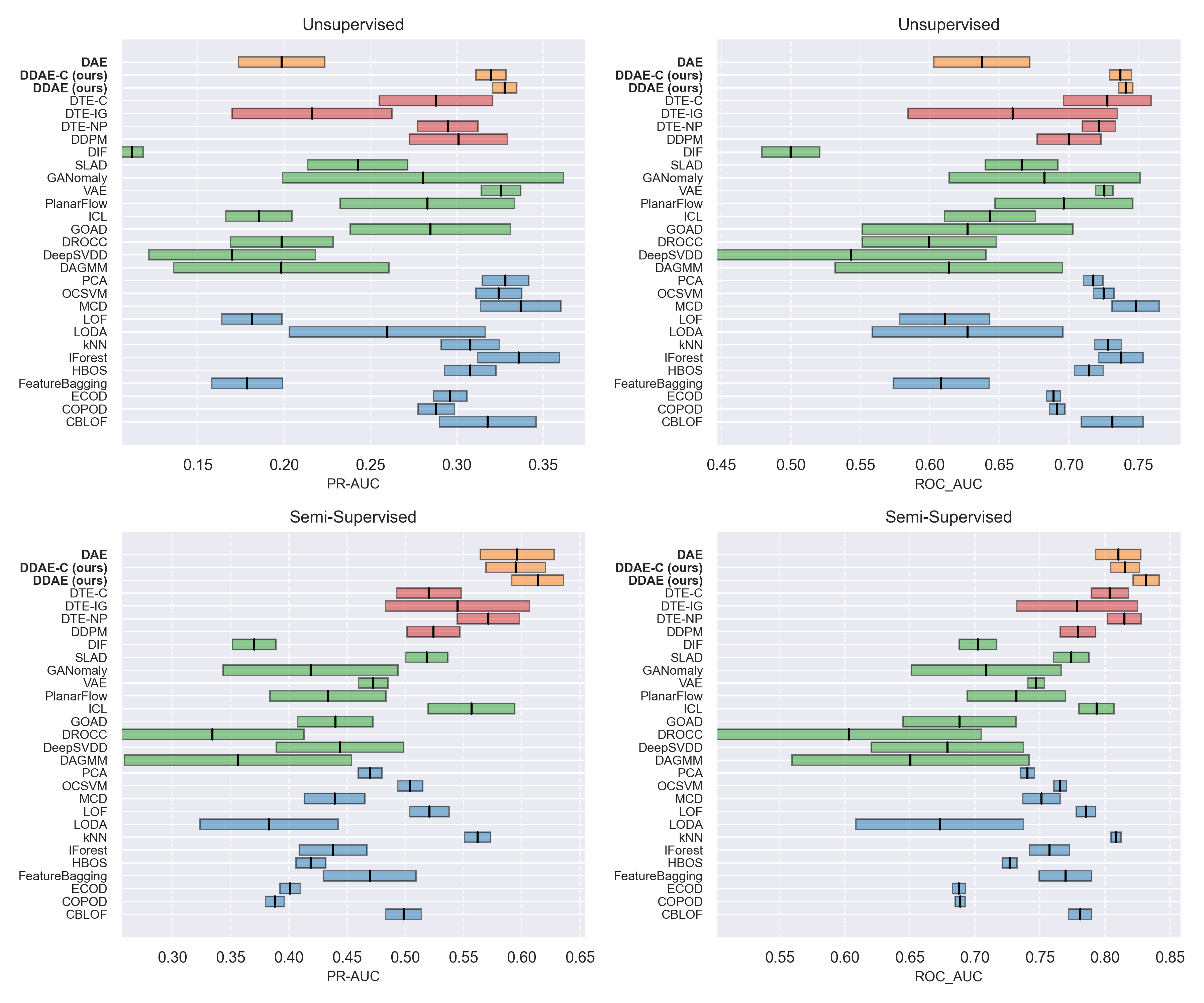}
\caption{Mean and standard deviation of PR-AUC and ROC-AUC scores across 57 datasets from ADBench, evaluated over five distinct seeds, in an unsupervised and semi-supervised settings. Color coding: red indicates diffusion-based approaches, green signifies deep learning techniques, blue represents classical methods, and orange reflects our methods. Results for models other than ours are sourced from \citep{livernoche2023diffusion}.}
\label{fig:baselines_full_grouped}
\end{figure*}

\begin{figure*}
    \centering
    \includegraphics[width=1\linewidth]{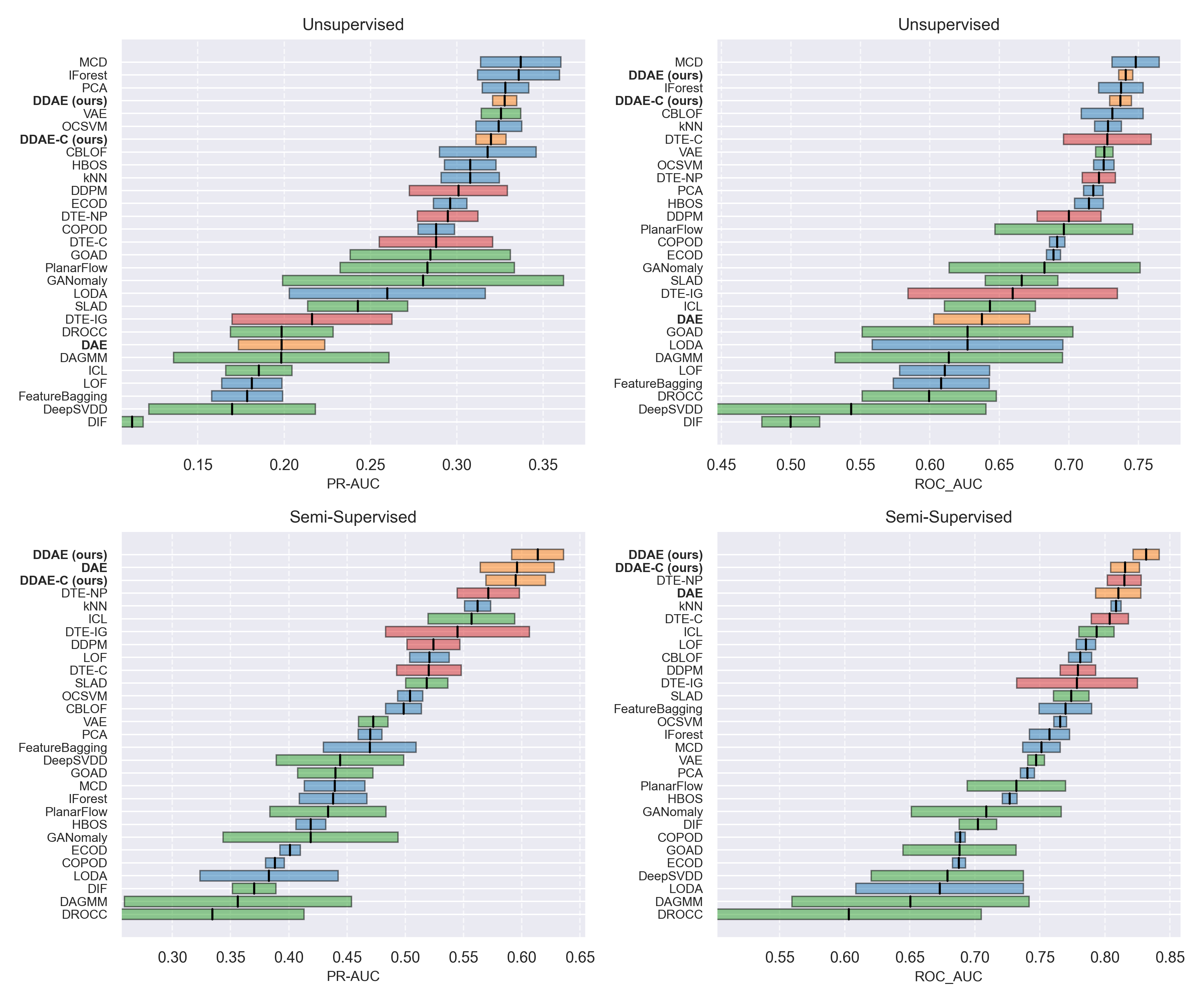}
\caption{Mean and standard deviation of PR-AUC and ROC-AUC scores across 57 datasets from ADBench, evaluated over five distinct seeds, in an unsupervised and semi-supervised settings. Color coding: red indicates diffusion-based approaches, green signifies deep learning techniques, and blue classical methods, and orange reflects our methods. Results for models other than ours are sourced from \citep{livernoche2023diffusion}.}
\label{fig:baselines_full_sorted}
\end{figure*}

\newpage

\begin{table*}[h]
\centering
\label{tab:pr_auc_semi_detailed}
\caption{Average PR-AUC scores over 5 seeds (semi-supervised). Results for models other than ours are sourced from \citep{livernoche2023diffusion}.}
\resizebox{!}{\textheight}{\rotatebox{90}{
\begin{tabular}{llllllllllllllllllllllllllllll}
\toprule
Dataset & CBLOF & COPOD & ECOD & FeatureBagging & HBOS & IForest & kNN & LODA & LOF & MCD & OCSVM & PCA & DAGMM & DeepSVDD & DROCC & GOAD & ICL & PlanarFlow & VAE & GANomaly & SLAD & DIF & DDPM & DTE-NP & DTE-IG & DTE-C & DAE & DDAE & DDAE-C \\
\midrule
aloi & .064\scriptsize{$\pm$0.02} & .0572\scriptsize{$\pm$0.0} & .0606\scriptsize{$\pm$0.0} & .068\scriptsize{$\pm$0.19} & .0642\scriptsize{$\pm$0.0} & .0582\scriptsize{$\pm$0.1} & .0602\scriptsize{$\pm$0.0} & .0593\scriptsize{$\pm$0.5} & .0654\scriptsize{$\pm$0.0} & .0555\scriptsize{$\pm$0.07} & .0652\scriptsize{$\pm$0.0} & .0654\scriptsize{$\pm$0.0} & .0607\scriptsize{$\pm$0.55} & .0623\scriptsize{$\pm$0.35} & .0591\scriptsize{$\pm$0.0} & .057\scriptsize{$\pm$0.24} & .055\scriptsize{$\pm$0.13} & .0548\scriptsize{$\pm$0.3} & .0654\scriptsize{$\pm$0.0} & .0809\scriptsize{$\pm$1.27} & .0599\scriptsize{$\pm$0.04} & .058\scriptsize{$\pm$0.0} & .0597\scriptsize{$\pm$0.06} & .0605\scriptsize{$\pm$0.06} & .0598\scriptsize{$\pm$0.14} & .0576\scriptsize{$\pm$0.04} & \textbf{.1155}\scriptsize{$\pm$\textbf{0.0}} & .0871\scriptsize{$\pm$0.0} & .0871\scriptsize{$\pm$0.0} \\
\bottomrule
\end{tabular}
}}
\end{table*}

\begin{table*}[htbp]
\centering
\label{tab:roc_auc_semi_detailed}
\caption{Average ROC-AUC scores over 5 seeds (semi-supervised). Results for models other than ours are sourced from \citep{livernoche2023diffusion}.}
\resizebox{!}{\textheight}{\rotatebox{90}{\begin{tabular}{llllllllllllllllllllllllllllll}
\toprule
Dataset & CBLOF & COPOD & ECOD & FeatureBagging & HBOS & IForest & kNN & LODA & LOF & MCD & OCSVM & PCA & DAGMM & DeepSVDD & DROCC & GOAD & ICL & PlanarFlow & VAE & GANomaly & SLAD & DIF & DDPM & DTE-NP & DTE-IG & DTE-C & DAE & DDAE & DDAE-C \\
\midrule
aloi & .5369\scriptsize{$\pm$0.15} & .4951\scriptsize{$\pm$0.0} & .5173\scriptsize{$\pm$0.0} & .4907\scriptsize{$\pm$0.53} & .5223\scriptsize{$\pm$0.0} & .5074\scriptsize{$\pm$0.68} & .5104\scriptsize{$\pm$0.0} & .4924\scriptsize{$\pm$2.75} & .4876\scriptsize{$\pm$0.0} & .4854\scriptsize{$\pm$0.35} & .5429\scriptsize{$\pm$0.0} & .5404\scriptsize{$\pm$0.0} & .5084\scriptsize{$\pm$2.97} & .5089\scriptsize{$\pm$2.05} & .5\scriptsize{$\pm$0.0} & .4801\scriptsize{$\pm$0.92} & .475\scriptsize{$\pm$0.98} & .4852\scriptsize{$\pm$2.34} & .5404\scriptsize{$\pm$0.0} & .5394\scriptsize{$\pm$1.65} & .5076\scriptsize{$\pm$0.29} & .511\scriptsize{$\pm$0.0} & .4991\scriptsize{$\pm$0.35} & .5119\scriptsize{$\pm$0.46} & .5087\scriptsize{$\pm$1.17} & .5044\scriptsize{$\pm$0.19} & \textbf{.5954}\scriptsize{$\pm$\textbf{0.0}} & .5617\scriptsize{$\pm$0.0} & .5594\scriptsize{$\pm$0.0} \\
\bottomrule
\end{tabular}
}}
\end{table*}

\begin{table*}[htbp]
\centering
\label{tab:pr_auc_unsup_detailed}
\caption{Average PR-AUC scores over 5 seeds (unsupervised). Results for models other than ours are sourced from \citep{livernoche2023diffusion}.}
\resizebox{!}{\textheight}{\rotatebox{90}{\begin{tabular}{llllllllllllllllllllllllllllll}
\toprule
Dataset & CBLOF & COPOD & ECOD & FeatureBagging & HBOS & IForest & kNN & LODA & LOF & MCD & OCSVM & PCA & DAGMM & DeepSVDD & DROCC & GOAD & ICL & PlanarFlow & VAE & GANomaly & SLAD & DIF & DDPM & DTE-NP & DTE-IG & DTE-C & DAE & DDAE & DDAE-C \\
\midrule
aloi & .0374\scriptsize{$\pm$0.07} & .0313\scriptsize{$\pm$0.0} & .0329\scriptsize{$\pm$0.0} & \textbf{.1036}\scriptsize{$\pm$\textbf{0.45}} & .0338\scriptsize{$\pm$0.03} & .0339\scriptsize{$\pm$0.03} & .0476\scriptsize{$\pm$0.02} & .0327\scriptsize{$\pm$0.29} & .0969\scriptsize{$\pm$0.28} & .0322\scriptsize{$\pm$0.05} & .0392\scriptsize{$\pm$0.14} & .0372\scriptsize{$\pm$0.03} & .0331\scriptsize{$\pm$0.26} & .0344\scriptsize{$\pm$0.27} & .0304\scriptsize{$\pm$0.0} & .0328\scriptsize{$\pm$0.23} & .046\scriptsize{$\pm$0.43} & .0323\scriptsize{$\pm$0.1} & .037\scriptsize{$\pm$0.0} & .0438\scriptsize{$\pm$0.29} & .0365\scriptsize{$\pm$0.1} & .0306\scriptsize{$\pm$0.15} & .0359\scriptsize{$\pm$0.02} & .0555\scriptsize{$\pm$0.02} & .0395\scriptsize{$\pm$0.11} & .0328\scriptsize{$\pm$0.07} & .0559\scriptsize{$\pm$0.0} & .038\scriptsize{$\pm$0.0} & .0382\scriptsize{$\pm$0.0} \\
\bottomrule
\end{tabular}
}}
\end{table*}

\begin{table*}[htbp]
\centering
\label{tab:roc_auc_unsup_detailed}
\caption{Average ROC-AUC scores over 5 seeds (unsupervised). Results for models other than ours are sourced from \citep{livernoche2023diffusion}.}
\resizebox{!}{\textheight}{\rotatebox{90}{\begin{tabular}{llllllllllllllllllllllllllllll}
\toprule
Dataset & CBLOF & COPOD & ECOD & FeatureBagging & HBOS & IForest & kNN & LODA & LOF & MCD & OCSVM & PCA & DAGMM & DeepSVDD & DROCC & GOAD & ICL & PlanarFlow & VAE & GANomaly & SLAD & DIF & DDPM & DTE-NP & DTE-IG & DTE-C & DAE & DDAE & DDAE-C \\
\midrule
aloi & .5558\scriptsize{$\pm$0.2} & .5153\scriptsize{$\pm$0.01} & .5306\scriptsize{$\pm$0.01} & \textbf{.7915}\scriptsize{$\pm$\textbf{0.56}} & .5311\scriptsize{$\pm$0.21} & .5422\scriptsize{$\pm$0.39} & .6132\scriptsize{$\pm$0.04} & .4952\scriptsize{$\pm$1.02} & .7666\scriptsize{$\pm$0.35} & .5204\scriptsize{$\pm$0.17} & .5486\scriptsize{$\pm$0.01} & .549\scriptsize{$\pm$0.08} & .5169\scriptsize{$\pm$2.13} & .5142\scriptsize{$\pm$2.92} & .5\scriptsize{$\pm$0.0} & .4969\scriptsize{$\pm$0.64} & .5484\scriptsize{$\pm$0.62} & .5201\scriptsize{$\pm$1.37} & .5485\scriptsize{$\pm$0.0} & .5475\scriptsize{$\pm$1.09} & .5424\scriptsize{$\pm$0.68} & .4977\scriptsize{$\pm$1.14} & .5322\scriptsize{$\pm$0.25} & .645\scriptsize{$\pm$0.09} & .541\scriptsize{$\pm$0.99} & .5248\scriptsize{$\pm$0.31} & .5992\scriptsize{$\pm$0.0} & .5553\scriptsize{$\pm$0.0} & .5555\scriptsize{$\pm$0.0} \\
\bottomrule
\end{tabular}
}}
\end{table*}

\end{document}